\PassOptionsToPackage{table,x11names,svgnames,dvipsnames}{xcolor}
\documentclass[sigconf]{acmart}

\copyrightyear{2025}
\acmYear{2025}

\acmConference[KDD '25] {Proceedings of the ACM SIGKDD Conference on Knowledge Discovery and Data Mining 2025} {August 3-7, 2025}{Toronto, Canada.}
\acmBooktitle{Proceedings of the ACM SIGKDD Conference on Knowledge Discovery and Data Mining 2025 (KDD '25), August 3-7, 2025, Toronto, Canada}

\usepackage{xspace}
\usepackage{enumerate}
\usepackage[ruled]{algorithm2e}
\usepackage{adjustbox}
\usepackage{multirow}
\usepackage{subfigure}
\usepackage{color}

\usepackage{soul}
\usepackage{makecell}
\definecolor{MyRed}{RGB}{151,0,0}
\definecolor{Yellow}{RGB}{243,140,10}

\newcommand{\model}{\texttt{DeSE}\xspace}

\newcommand{\nameHL}[1]{\underline{\textbf{#1}}}

\begin{document}
\title{Unsupervised Graph Clustering with Deep Structural Entropy}

\author{Jingyun Zhang}
\orcid{0009-0003-7215-0040}
\affiliation{%
  \institution{Beihang University}
  \state{Beijing}
  \country{China}
}
\email{zhangjingyun@buaa.edu.cn}

\author{Hao Peng}
\authornote{Corresponding author.}
\orcid{0000-0003-0458-5977}
\affiliation{%
  \institution{Beihang University}
  \state{Beijing}
  \country{China}
}
\email{penghao@buaa.edu.cn}

\author{Li Sun}
\orcid{0000-0003-4562-2279}
\affiliation{%
  \institution{NCEPU}
  \state{Beijing}
  \country{China}
}
\email{ccesunli@ncepu.edu.cn}

\author{Guanlin Wu}
\orcid{0000-0001-9968-2977}
\affiliation{%
  \institution{School of Systems Engineering, TUDT}
  \state{Changsha}
  \country{China}
}
\email{wuguanlin16@nudt.edu.cn}

\author{Chunyang Liu}
\orcid{0000-0001-7282-0284}
\affiliation{%
  \institution{Didi Chuxing Technology Co., Ltd.}
  \city{Beijing}
  \country{China}
}
\email{liuchunyang@didiglobal.com}

\author{Zhengtao Yu}
\orcid{0000-0003-1094-5668}
\affiliation{%
  \institution{KUST}
  \state{Yunnan}
  \country{China}
}
\email{yuzt@kust.edu.cn}


\begin{abstract}
Research on Graph Structure Learning (GSL) provides key insights for graph-based clustering, yet current methods like Graph Neural Networks (GNNs), Graph Attention Networks (GATs), and contrastive learning often rely heavily on the original graph structure. 
Their performance deteriorates when the original graph's adjacency matrix is too sparse or contains noisy edges unrelated to clustering. 
Moreover, these methods depend on learning node embeddings and using traditional techniques like k-means to form clusters, which may not fully capture the underlying graph structure between nodes.
To address these limitations, this paper introduces \textbf{~\model{}}, a novel unsupervised graph clustering framework incorporating \nameHL{De}ep \nameHL{S}tructural \nameHL{E}ntropy. 
It enhances the original graph with quantified structural information and deep neural networks to form clusters. 
Specifically, we first propose a method for calculating structural entropy with soft assignment, which quantifies structure in a differentiable form. 
Next, we design a Structural Learning layer (SLL) to generate an attributed graph from the original feature data, serving as a target to enhance and optimize the original structural graph, thereby mitigating the issue of sparse connections between graph nodes.
Finally, our clustering assignment method (ASS), based on GNNs, learns node embeddings and a soft assignment matrix to cluster on the enhanced graph. 
The ASS layer can be stacked to meet downstream task requirements, minimizing structural entropy for stable clustering and maximizing node consistency with edge-based cross-entropy loss.
Extensive comparative experiments are conducted on four benchmark datasets against eight representative unsupervised graph clustering baselines, demonstrating the superiority of the ~\model in both effectiveness and interpretability.
\end{abstract}





\keywords{Unsupervised clustering, Graph structure learning, Structural entropy}

\maketitle

\section{Introduction}
\label{sec: introduction}
Graph structure learning (GSL) has a wide range of applications in recommender systems~\cite{wei2022contrastive}, community detection~\cite{zhang2020commdgi}, interest discovery~\cite{zhou2023interest}, web topic mining~\cite{zhang2022cross}, graph clustering~\cite{kang2021structured, li2020deep}, dimensionality reduction~\cite{mao2015dimensionality}, etc. 
It integrates with downstream tasks to refine the graph topology and generate node classifications, enabling the learning of robust semantic embeddings and structural information, thereby improving the performance of various applications. 
Unsupervised graph clustering typically employs contrastive loss to learn an appropriate graph structure embedding, enhancing the representational similarity of nodes within clusters while avoiding dependence on labeled data. \par
\vspace{-3mm}
\begin{figure}[h]
    \centering
    \includegraphics[width=0.9\linewidth]{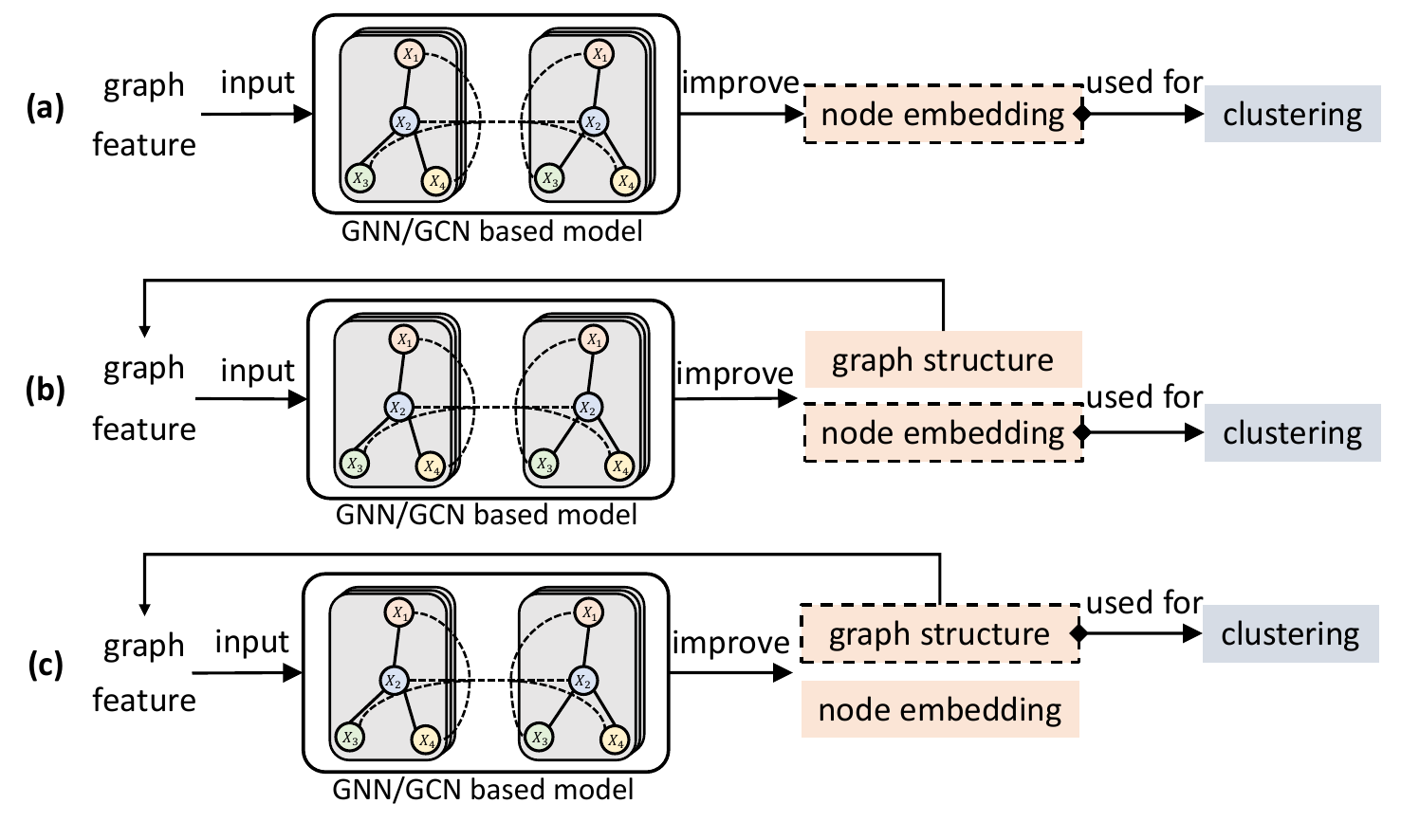}
    \vspace{-5mm}
    \caption{Concept maps of three type models. ((a) and (b) are existing models, (c) is our~\model{})}
    \vspace{-3mm}
    \label{fig: map}
\end{figure}

Early unsupervised graph clustering methods rely heavily on the original graph structure and focus solely on optimizing it within the model. 
Examples include hierarchical graph learning methods~\cite{ying2018hierarchical, wu2022learning, lin2023hyperbolic}, pooling techniques~\cite{bianchi2020hierarchical}, and structure-based embedding learning approaches~\cite{tsitsulin2023graph, lyu2017enhancing}.
The primary goal of these methods is to learn better node embeddings by minimizing the distance between neighboring or structurally similar nodes, as shown in Figure~\ref{fig: map}(a).
However, nodes with similar attributes may not always have direct connections in practice.
For instance, papers in the same category often lack direct or indirect citation links in citation networks. 
This dependence on the original graph, which is typically a sparse adjacency matrix, significantly limits the performance of the models. \par

To address these issues, existing methods optimize the original graph structure through approaches such as graph simplification~\cite{chen2018harp, gong2022attributed} and latent structure learning~\cite{wu2022nodeformer, fatemi2021slaps, liu2017structure}.
Specifically, these approaches use graph contrastive techniques to extract structural knowledge~\cite{liu2022towards, ding2023eliciting, yang2023convert}, or graph autoencoders to simultaneously learn representations and perform clustering tasks~\cite{zhang2022embedding, mrabah2022rethinking}, in order to mitigate feature drift. 
However, these methods still rely on the learned embeddings to form clusters, as shown in Figure~\ref{fig: map}(b). 
The most common approach is the classic K-Means algorithm, which requires prior knowledge of the number of clusters. 
We argue that such a model relies on both the quality of representation learning and the configuration of the clustering algorithm. 
Moreover, models that first learn embeddings and then perform clustering do not directly capture the essential relationship between node features and adaptive clusters during model convergence.
Additionally, although these methods focus on unsupervised graph clustering and produce structured clusters, few models quantitatively represent the graph structure, leading to poor interpretability. 
While ~\cite{tsitsulin2023graph} proposes optimizing the node assignment matrix using modularity, this measure captures the difference between actual intra-cluster edges and expected edges, where nodes with higher degrees are more likely to be connected. 
This approach is not applicable in all scenarios. 
In tasks like citation networks and social event networks, each cluster often contains a central node closely connected to other nodes. 
Still, these central nodes belong to different categories and have limited connections with one another. \par

In this work, we propose\textbf{~\model{}}, a novel unsupervised graph clustering framework with \nameHL{De}ep \nameHL{S}tructural \nameHL{E}ntropy, which enhances the original graph using quantifiable structural information and deep neural networks to improve clustering performance and interpretability. 
First, in terms of structural quantification, we introduce structural information theory and propose a new method for calculating soft assignment structural entropy in the context of the graph clustering task. 
We transform structural entropy into a continuous and differentiable form by utilizing a probability matrix that assigns cluster nodes. 
This allows for retaining more information about boundary nodes during the embedding aggregation process rather than discarding low-probability nodes outright.
Second, we design a Structure Learning Layer (SLL) to enhance the original graph structure. 
By constructing a K-nearest neighbor graph in the node feature mapping space, we create an attribute graph to address the sparsity and missing interactions between graph nodes. 
This attribute graph is continuously optimized and refined during the training process.
Third, we propose a cluster assignment method (ASS) based on Graph Neural Networks, directly in the enhanced graph rather than using embeddings, as shown in Figure~\ref{fig: map}(c).
ASS employs two convolutional layers: one for learning the embeddings of the current layer's nodes and the other for learning the soft assignment matrix. 
These are then aggregated to obtain embeddings for higher-level communities and update the graph structure at the upper levels. 
Finally, the model is optimized by minimizing the structural entropy of the assignments to stabilize the cluster structure and by using an edge-based cross-entropy loss to maximize the consistency between connected nodes, thereby achieving unsupervised graph clustering. \par

We conduct extensive experiments on four datasets, Cora, Citeseer, Computer, and Photo, to demonstrate the effectiveness of~\model{}.
First, the overall experimental results indicate that~\model{} demonstrates superior overall performance compared to the eight baseline models.
Second, a series of ablation experiments analyze the SLL, GNN layer for features, and structure entropy loss~\model{}.
Thirdly, the experiment for hyperparameters also illustrates the high performance and stability of~\model{}.
The main contributions of this work are summarized as follows.

$\bullet$
A novel unsupervised graph clustering framework with Deep Structural Entropy is proposed with high effectiveness.

$\bullet$
A structure learning layer (SLL) and a cluster assignment method (ASS) based on Graph Neural Networks are designed for enhancing structure and graph clustering.

$\bullet$
A new optimization method that minimizes the structural entropy of the assignments and maximizes the consistency between connected nodes.

$\bullet$
A series of comparative analysis experiments show that ~\model{} achieves higher graph clustering effectiveness and strong interpretability.

\vspace{-2mm}
\section{Related Work}
\label{sec: related_work}
\vspace{-2mm}
\subsection{Unsupervised Graph Clustering}
Unsupervised graph clustering has evolved significantly over the past decades, from traditional methods like spectral clustering~\cite{von2007tutorial} and modularity-based approaches~\cite{newman2006modularity} to the more sophisticated deep learning models used today. 
Early methods primarily focused on leveraging graph structure alone, which limited their performance in dealing with complex, feature-rich data.
Spectral clustering and its variants, which use eigenvalue decomposition of graph Laplacians, have been widely used due to their theoretical simplicity and effectiveness~\cite{white2005spectral}. 
However, these approaches scale poorly to large graphs and are noise-sensitive, especially when the graph structure is sparse or incomplete. 
With the advent of deep learning, there has been a shift towards models that integrate node features and graph structure. 
Techniques such as Graph Neural Networks (GNNs), particularly Graph Convolutional Networks (GCNs)~\cite{kipf2016semi}, and Graph Autoencoders (GAEs)~\cite{kipf2016variational}, have gained traction for unsupervised graph clustering tasks. 
These models learn node embeddings that preserve local and global structures, facilitating better clustering results. 
Furthermore, some approaches have started to explore the role of structure in enhancing clustering~\cite{wu2022learning, lin2023hyperbolic}. 
These methods aim to quantify and improve the quality of graph structures, particularly in the presence of noise, leading to more stable clustering results.\par

\vspace{-4mm}
\subsection{Graph Structure Learning}
Graph Structure Learning (GSL) has gained increasing attention in recent years as researchers seek to optimize graph structures for downstream tasks like clustering and node classification. 
Unlike traditional methods that rely on predefined graphs, GSL techniques dynamically learn or refine the graph structure based on node features and interactions. 
This approach has shown to be particularly effective in handling noisy, incomplete, or poorly defined graphs. 
Adaptive graph learning methods~\cite{jin2020graph, chen2018harp} focus on pruning noisy edges or adding informative ones to improve graph quality. 
These models typically apply sparsity constraints or similarity metrics to update the graph structure during training, resulting in a cleaner and more informative graph representation.
Joint learning approaches~\cite{yu2021graph, jin2020graph} simultaneously learn the graph structure and node embeddings in an end-to-end manner. 
These methods are particularly powerful because they allow structure optimization based on the specific task.
A key challenge in GSL is balancing structural refinement with maintaining meaningful graph relationships. 
These techniques show promise in handling noisy graphs and enhancing overall task performance, especially in unsupervised or semi-supervised settings.\par

\begin{figure*}[t]
    \centering
    \includegraphics[width=0.75\linewidth]{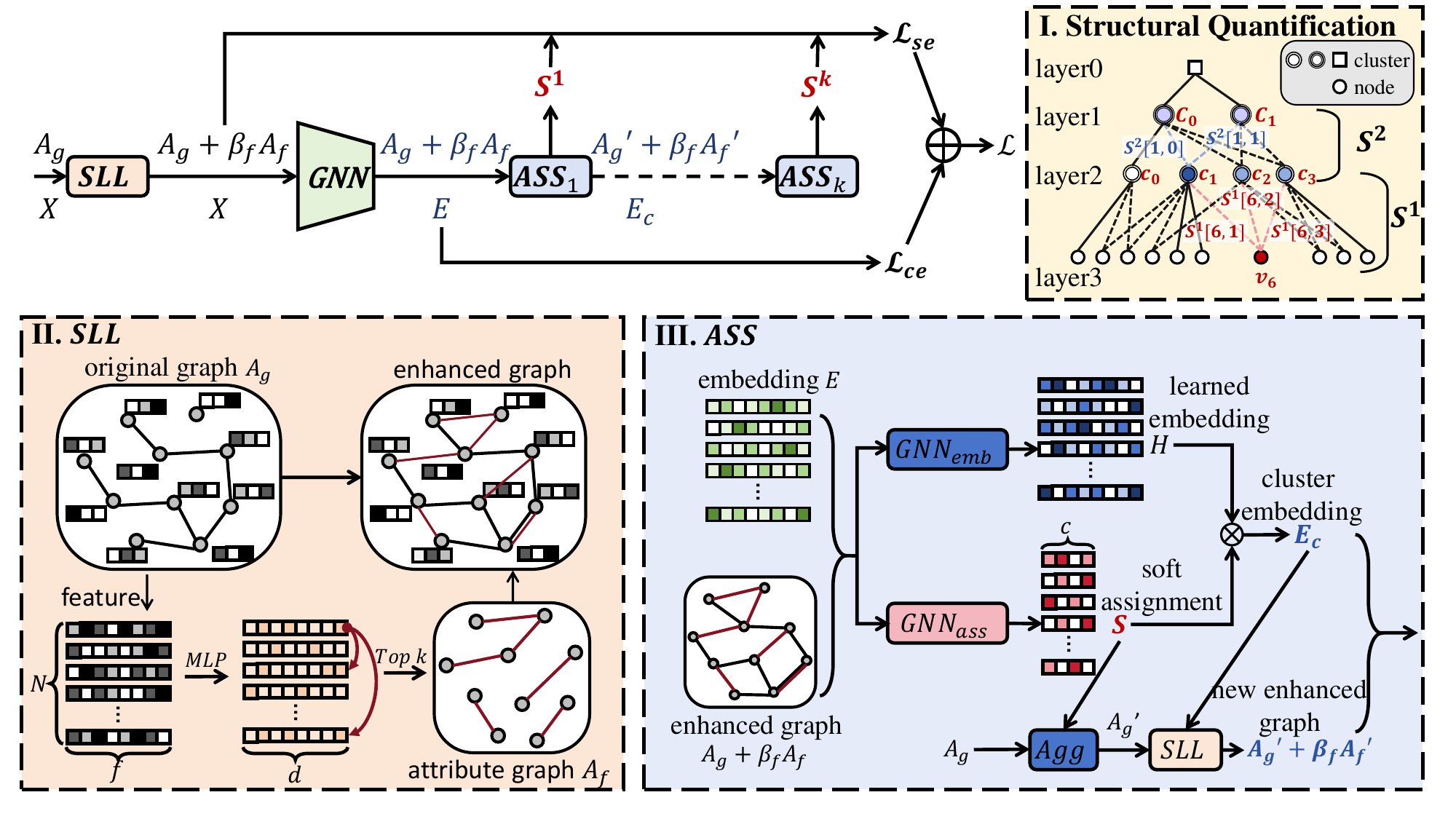}
    \vspace{-6mm}
    \caption{The overall framework of~\model{}.}
    \label{fig: DeSE}
    \vspace{-5mm}
\end{figure*}

\vspace{-4mm}
\subsection{Structural Information Theory}
The Structural Information Theory decoding network's ability to capture the structure's essence has been validated in many applications. 
Introducing structural entropy in neural networks captures the underlying connectivity graph and reduces random interference~\cite{wang2023user}. 
The hierarchical nature of the structure entropy encoding tree provides new methods for hierarchical structure pooling in graph neural network~\cite{wu2022structural}, unsupervised image segmentation~\cite{zeng2023unsupervised}, dimension estimation~\cite{yang2023minimum}, state abstraction~\cite{zeng2023hierarchical} in reinforcement learning, social bot detection~\cite{yang2024sebot,peng2024unsupervised}, and unsupervised social event detection~\cite{cao2024hierarchical}. 
Additionally, reconstructing the graph structure on the hierarchical encoding tree suppresses edge noise and enhances the learning ability of the graph structure~\cite{zou2023se, zou2024multispans}. 
Furthermore, modifying the network structure based on minimizing structural entropy achieves maximum deception of community structure~\cite{liu2019rem}. 
Similarly, the anchor view, guided by the principle of minimizing structural entropy, improves the performance of graph contrastive learning~\cite{wu2023sega}.
Based on the homogeneous graph structure entropy, the study of multi-relational graph structure entropy~\cite{cao2024multi} further extends the structural information theory, making it suitable for more complex scenarios.\par

\vspace{-2mm}
\section{Preliminary}
\label{sec: preliminary}
\begin{definition}[\textbf{Unsupervised Graph Clustering}] 
The unsupervised graph clustering task aims to cluster nodes based solely on the provided interaction and feature information. 
From a data perspective, the input consists of an undirected homogeneous graph with node features, represented as $G=(V, \mathcal{E}, X)$, where $V$ is a set of $N$ nodes, $\mathcal{E}$ is a set of $M$ edges, and $X \in R^{N \times f}$ is the node feature matrix with dimension $f$. 
The edge relationships between nodes in $G$ are represented by a symmetric adjacency matrix $A_g \in {\{0,1\}}^{N \times N}$, where each element denotes the edge weight between nodes. 
The objective of the unsupervised graph clustering task is to learn a node assignment matrix $S \in {\{0,1\}} ^{N \times c}$, which reflects the community memberships of nodes, where $c$ is the number of clusters. 
In our model, $c$ does not need to be specified in advance.
\end{definition}

\vspace{-2mm}
\begin{definition}[\textbf{Structural Entropy}]
Structural information theory~\cite{li2016structural} is originally proposed for measuring the structural information contained within a graph. 
Specifically, this theory aims to calculate the structural entropy of the homogeneous graph $G=(V, \mathcal{E})$, which reflects its uncertainty when undergoing hierarchical division. 
The structural information of the homogeneous graph $G$ determined by the encoding tree $\mathcal{T}$ is defined as:
\begin{eqnarray} 
    H^\mathcal{T}(G)=-\sum_{\alpha \in \mathcal{T},\alpha \neq \lambda}\frac{g_\alpha}{vol(G)}log\frac{vol(\alpha)}{vol(\alpha^-)},
\label{Eq: H_T}
\end{eqnarray}
where $vol(G)$ is the sum of the degrees of all nodes in the graph $G$. 
Each vertice in the encoding tree $\mathcal{T}$ corresponds to a node subset $T_{\alpha}$  in the graph $G$.
$vol(\alpha)$ is the volume of $T_{\alpha}$ and is the sum of the degrees of all nodes in the subset $T_{\alpha}$. 
$\alpha^-$ is the parent vertice of vertice $\alpha$ in the encoding tree.
$g_{\alpha}$ is the sum of weights of all edges from node subset $T_{\alpha}$ to node subset $V/ T_{\alpha}$, which can be understood as the total weight of the edges from the nodes outside the node subset $T_{\alpha}$ to the nodes inside $T_{\alpha}$, or the total weight of the cut edges.
$\frac{g_\alpha}{vol(G)}$ represents the probability that the random walk enters $T_{\alpha}$.
The structural entropy $H(G)$ of graph $G$ is the minimum $H^{\mathcal{T}}(G)$. 
Let $\mathcal{T}_k$ be encoding trees whose height is not greater than $k$, then the $k$-dimensional structural entropy of $G$ is defined as $H_k(G)=min \{ H^{\mathcal{T}_k}(G) \}$.
\end{definition}

\vspace{-2mm}
\section{Methodology}
\label{sec: methodology}
This section elaborates on the unsupervised graph clustering framework~\model{} with deep structural entropy. 
As shown in Figure~\ref{fig: DeSE}, \model{} consists of three key modules: 
Structural Quantification, Structural Learning Layer, and
Clustering Assignment Layer.
Specifically, \textbf{Structural Quantification} (Section~\ref{subsec: structural quantification}) introduces a soft assignment structural entropy, quantifying the structural information and transforming discrete clustering into a continuous and differentiable objective. 
\textbf{Structural Learning Layer (SLL)} (Section~\ref{subsec: SLL}) learns a K-nearest neighbor attribute graph in the feature mapping space to enhance the original graph structure. 
\textbf{Clustering Assignment Layer (ASS)} (Section~\ref{subsec: ASS}), based on GNN, simultaneously learns node embeddings and a soft assignment matrix in the enhanced graph, updating new cluster embeddings and the cluster-enhanced graph. 
\textbf{Optimization} (Section~\ref{subsec: optimization}) integrates all modules, optimizing the learning process using structural entropy loss between nodes and clusters and cross-entropy loss between node embeddings.

\vspace{-3mm}
\subsection{Structural Quantification}
\label{subsec: structural quantification}
Structural information theory offers significant advantages in learning hierarchical structures and clusters of graph nodes. 
It quantifies the uncertainty in graph structures and represents them in a mathematically computable form. 
Although prior research has extended structural entropy~\cite{peng2024unsupervised, cao2024multi} and its optimization methods from simple homogeneous graphs to multi-relational graphs and hypergraphs, structural entropy is still calculated in a discrete manner. 
This limitation restricts current optimization methods to operations like the merge operator, where nodes are greedily merged in pairs. 
However, in unsupervised graph clustering tasks, we aim for structural entropy to not only divide nodes into clusters but also provide trainable feedback to enhance the graph structure. 
This highlights the limitations of traditional structural entropy. \par

To address this issue, we transform the original binary "belong/not belong" relationship between nodes and clusters (represented as discrete values of 0 or 1 in the assignment matrix) into a probabilistic relationship. 
A node is no longer exclusively assigned to a single cluster; instead, it can belong to multiple clusters with varying probabilities. 
This approach aligns with real-world scenarios, such as interdisciplinary papers in citation networks. 
Although these papers have a primary category, they also contribute to and are associated with other relevant categories, which is valuable information in embedding and structure learning. 
This probabilistic cluster assignment is also known as a "soft assignment." \par

\textbf{Soft Assignment SE}.
The traditional definition of structural entropy is presented in Section~\ref{sec: preliminary}. 
Structural information theory quantifies the uncertainty in a graph's structure based on the random walk of nodes through edges. 
When the lower-level vertices belong to the parent vertex according to the assignment matrix $S^k$ at layer $k$, we first express Eq.~\ref{Eq: H_T} as the sum of node entropies at each layer and introduce the concept of a direct assignment matrix:
\begin{equation}
    H^{\mathcal{T}}_{sa}(G) =\sum_{k=1}^{h}{H_{sa}(G;k)},
    \label{Eq: H_soft}
\end{equation}
\begin{equation}
    C^k = S^{h} \cdot S^{h-1} \cdot ... \cdot S^{k+1},
    \label{Eq: C}
\end{equation}
where $H_{sa}(G;k)$ in Eq.~\ref{Eq: H_soft} denotes the structural entropy at layer $k$ with $N_{k}$ vertices, while the total height of the encoding tree is $h$.
The $S^k \in R^{N_k \times N_{k-1}}$ represents the assignment matrix between the vertices of layer $k$ and those of layer $k-1$.
As computed in Eq.~\ref{Eq: C}, $C^k \in R^{N \times N_k}$ is the direct assignment matrix between the leaf vertices (i.e., the nodes in the graph) and the vertices of layer $k$, representing the probability that each node belongs to a cluster at layer $k$.
Additionally, we redefine the calculation and representation of cut edges and volume as follows:
\begin{equation}
    {vol}^k[i]=D (C^k)_i,
    \label{Eq: vol}
\end{equation}
\begin{equation}
    g^k_i = {vol}^k[i]-{vol}^k_{in}[i] = {vol}^k[i]-({{C^k})_i}^{\intercal} W ({C^k})_i,
    \label{Eq: g}
\end{equation}
where the volume ${vol}^k[i]$ of vertex $i$ at layer $k$ is the sum of the assignment probabilities over all node degrees, as expressed in Eq.~\ref{Eq: vol}.
And $D \in R^{N}$ is the degree vector for all nodes, which can be obtained through the weight matrix of edges $W \in R^{N \times N} $ that $D=\mathbf{1_N}^\intercal \cdot W$ ($\mathbf{1_N}^\intercal$ is a length-N vector consisting entirely of ones and the calculation of weight matrix $W$ will be detailed in Section~\ref{subsec: SLL}).
The subscript $i$ on $C^k$ refers to the $i$-th column vector of the matrix, which represents the direct clustering probability of $N$ nodes in the graph to the $i$-th cluster at layer $k$.
And the $i$-th vertex at layer k represents the $i$-th cluster at layer $k$.
The term $g^k_i$ represents the cut value of the $i$-th vertex at layer $k$, which is calculated as the difference between the volume ${vol}^k[i]$ of vertex $i$ and its internal volume ${vol}^k_{in}[i]$. 
The internal volume ${vol}^k_{in}[i]$ is expressed as the sum of the weighted probabilities of all edges, where the probability refers to the likelihood that the two nodes connected by an edge belong to the same cluster $i$ at layer $k$.
Thus, the computation of structural entropy is modified as follows:
\begin{equation}
\begin{split}
    H_{sa}(G;k) &=- \sum_{i=1}^{N_k}{ \frac{g^k_i}{vol^0} log \frac{{vol}^k[i]}{\sum_{j=1}^{N_{k-1}} {vol}^{k-1}[j] \cdot S^k[i,j] } }\\
    &= - \sum_{i=1}^{N_k}{ \frac{{vol}^k[i]-({{C^k})_i}^{\intercal} W ({C^k})_i }{vol^0} log \frac{{vol}^k[i]}{ {vol}^{k-1} {{(S^k)}^{\intercal}}_i} }\\
    &= - \sum_{i=1}^{N_k}{ \frac{D (C^k)_i -({{C^k})_i}^{\intercal} W ({C^k})_i }{D C^0} log \frac{D (C^k)_i}{ D C^{k-1} {{(S^k)}^{\intercal}}_i} },
\end{split}
\label{Eq: H_soft_k}
\end{equation}
where the original volume associated with a single parent vertex is replaced by the probabilistic sum of the volumes of all parent vertices ${vol}^{k-1} {{(S^k)}^{\intercal}}_i$ in the soft assignment approach.
And ${vol}^{k-1}=[{vol}^{k-1}[1],...,{vol}^{k-1}[N_{k-1}]]$ is the vector representation of the volume of all $N_k$ vertices at layer $k$, which can be further simplified to $D C^{k-1}$ by Eq.~\ref{Eq: vol}

\vspace{-3mm}
\subsection{Structural Learning Layer (SLL)}
\label{subsec: SLL}
In graph clustering tasks, the input typically consists of node features $X$ and an adjacency matrix $A_g$. 
The general approach is to train the node features based on the original structure to generate embeddings. 
However, the connections in $A_g$ do not always align perfectly with clustering objectives. 
For instance, in citation networks, papers on similar topics may not be directly linked (e.g., papers on neural networks but focusing on different application areas), or papers that are linked may not belong to the same topic (e.g., interdisciplinary or multi-methodology papers). 
Similarly, in product co-purchase networks, items bought together may be complementary rather than similar (e.g., desktop computers and monitors or cameras and lenses). 
Such mismatches in the original graph structure $A_g$ will lead to information loss during the aggregation process, which can hinder the effectiveness of graph clustering. \par

The SLL aims to enhance the original graph structure by leveraging the available feature information $X$ and dynamically optimizing and updating the graph during training. 
Since the original features are often high-dimensional and sparse binary vectors, we first map the node features into a lower-dimensional dense space using a multilayer perceptron (MLP), as shown in Figure~\ref{fig: DeSE}~\uppercase\expandafter{\romannumeral2}.
Based on these embeddings, we apply the K-nearest neighbors algorithm (KNN) to select the top $K$ neighbors for each node and create edges with a weight of $1$. 
This process constructs an attribute graph, mathematically represented as:
\begin{equation}
    A_f = KNN (MLP (X;\Theta_f);K),
    \label{Eq: A_f_1}
\end{equation}
where $MLP$ is a multilayer perceptron with an input size of $f$ and both hidden and output layers of size $d$, while $\Theta_f \in R^{f \times d}$ represents the parameters of the $MLP$.
$KNN$ is the K-nearest neighbors operation, and each row of the obtained adjacency matrix $A_f \in {\{0,1\}}^{N \times N}$ indicates the neighbor selection for each node.
Since $KNN$ selects neighbors by ranking the distances between nodes, the neighbors may be unidirectional. 
That is if the node $v_i$ is among the top $K$ neighbors of node $v_j$, node $v_j$ may not necessarily be among the top $K$ neighbors of node $v_i$. 
Given the different implications of unidirectional versus bidirectional neighbor selection, we adjust the adjacency matrix of the attribute graph as follows:
\begin{equation}
    A_f = (A_f +{A_f}^\intercal )/2.
    \label{Eq: A_f_2}
\end{equation}
This ensures that the attribute graph's adjacency matrix becomes symmetric while still partially retaining the unidirectional and bidirectional neighbor selections.
Finally, we combine the original graph adjacency matrix with the attribute graph adjacency matrix to obtain an enhanced graph:
\begin{equation}
    W=A_g+\beta_f A_f,
    \label{Eq: W}
\end{equation}
where $\beta_f$ is a hyperparameter controlling the weight of the attribute graph in the fusion, and $W \in R^{N \times N}, $ is the new adjacency matrix with edge weights used for soft assignment SE in Section~\ref{subsec: structural quantification} and subsequent calculations.

\vspace{-3mm}
\subsection{Clustering Assignment Layer (ASS)}
\label{subsec: ASS}
The clustering assignment layer utilizes the initial embeddings and the adjacency matrix to learn the soft assignments and embeddings of nodes while updating the graph structure and cluster embeddings after aggregation. 
It consists of three components: Embedding Learner ${GNN}_{emb}$, Soft Assignment Learner ${GNN}_{ass}$, and Aggregator $Agg$, as shown in Figure~\ref{fig: DeSE}~\uppercase\expandafter{\romannumeral3}.\par

\textbf{Embedding Learner}. 
The embedding learner is based on a GNN architecture, mathematically represented as follows:
\begin{equation}
    H = {GNN}_{emb}(E,W;\Theta_1)=ReLU(mean(WE\Theta_1)),
    \label{Eq: GNN_emb}
\end{equation}
where ${GNN}_{emb}$ applies a linear transformation to the initial embedding $E$, mapping it to the embedding space.
It then aggregates the average embeddings of connected nodes to generate new node embeddings, followed by an activation function. 
The learnable parameters of the linear transformation are denoted by $\Theta_1 \in R^{d \times d}$. \par

\textbf{Soft Assignment Learner}.
The soft assignment learner extends the GNN architecture with an attention mechanism, mathematically represented as follows:
\begin{equation}
    \begin{split}
    & S = {GNN}_{ass}(E,W;\Theta_2)=ReLU((\Gamma \circ W)E\Theta_2),\\
    & \Gamma_{i,j} = \frac{LeakyReLU((e_i || e_j)\Theta_3))}{\sum_{j=1}^{N}{LeakyReLU((e_i || e_j)\Theta_3))}},    
    \end{split}
    \label{Eq: GNN_ass}
\end{equation}
where ${GNN}_{ass}$ performs a linear transformation on the initial embeddings $E$ into the cluster space (i.e., with dimensions equal to the number of clusters). 
It then computes the attention matrix $\Gamma$ for each edge to serve as aggregation weights, performing non-averaged embedding aggregation to obtain cluster embeddings. 
The learnable parameters of the linear transformation in ${GNN}_{ass}$ are denoted by $\Theta_2 \in R^{d \times c}$.
The computation of attention involves linearly transforming the concatenated embeddings of the nodes at each end of an edge into a 1-dimensional space (i.e., the weight space), followed by activation and normalization. 
$e_i$ and $e_j$ represent the embedding of node $v_i$ and node $v_j$.
The learnable parameters for the linear transformation in the attention mechanism are denoted by $\Theta_3 \in R^{2d \times 1}$. \par

\textbf{Aggregator}.
The aim of $Agg$ is to update the embedding and adjacency matrix of clusters.
We use the probability sum of node embeddings for cluster embeddings, denoted as $E_c=S^\intercal H \in R^{c \times d}$. 
The new adjacency matrix is combined from the attribute graph adjacency matrix and the structural graph adjacency matrix, similar to the structural learning method described in Section~\ref{subsec: SLL} Eq.~\ref{Eq: W}. 
The detailed computation process is as follows:
\begin{equation}
    {A_g}^{'} = S^\intercal A_g S, \quad {A_f}^{'} = KNN(MLP(E_c; \Theta_c); K),
    \label{Eq: update A}
\end{equation}
where $\Theta_c \in R^{d \times d}$ represents the learnable parameters of the $MLP$ for clusters and the new weighted adjacency matrix is $W^{'}={A_g}^{'} + \beta_f {A_f}^{'}$.

\begin{table*}[t]
    \centering
    \caption{Comparison of the NMI, ARI, ACC, and F1 across different methods on four datasets. The best results are bolded, and the second-best results are underlined.}
    \vspace{-4mm}
    \label{tab: results}
    \renewcommand\arraystretch{1.0}
    \setlength{\tabcolsep}{1.1mm}
    \begin{tabular}{c|cccc|cccc|cccc|cccc}
        \toprule
        \multirow{2}*{Method} & \multicolumn{4}{c|}{\multirow{1}*{\textbf{Cora}}} & \multicolumn{4}{c|}{\multirow{1}*{\textbf{Citeseer}}} & \multicolumn{4}{c|}{\multirow{1}*{\textbf{Computer}}} & \multicolumn{4}{c}{\multirow{1}*{\textbf{Photo}}}\\
        
        \cline{2-17}
        
        & {\textbf{NMI}} & {\textbf{ARI}} & {\textbf{ACC}} & \multicolumn{1}{c|}{\textbf{F1}} 
        & {\textbf{NMI}} & {\textbf{ARI}} & {\textbf{ACC}} & \multicolumn{1}{c|}{\textbf{F1}} 
        & {\textbf{NMI}} & {\textbf{ARI}} & {\textbf{ACC}} & \multicolumn{1}{c|}{\textbf{F1}} 
        & {\textbf{NMI}} & {\textbf{ARI}} & {\textbf{ACC}} & {\textbf{F1}} \\

        \hline

        DMoN
        & 48.98 & 40.55 & 63.55 & 58.91
        & 33.70 & 30.27 & 59.12 & 56.68 
        & 49.30 & 31.82 & 49.13 & 39.29
        & \underline{63.38} & 52.41 & 73.83 & \underline{71.05}\\

        MinCut
        & 40.40 & 29.44 & 54.69 & 53.13
        & 28.70 & 23.60 & 49.02 & 46.96
        & 48.37 & 26.07 & 38.92 & 34.73
        & 57.47 & 43.88 & 66.26 & 64.84\\

        DGI
        & 53.56 & 49.71 & 70.35 & 67.56
        & 38.36 & 31.53 & 53.14 & 42.25
        & 22.68 & 12.76 & 29.21 & 24.52
        & 30.70 & 14.63 & 39.59 & 37.74 \\

        \hline
        
        SUBLIME
        & 54.20 & 50.30 & 71.30 & 63.50
        & \underline{44.10} & 43.90 & 68.50 & 63.20 
        & 45.55 & 24.89 & 40.94 & 34.64 
        & 56.91 & 45.11 & 64.95 & 61.24\\

        EGAE
        & 54.00 & 47.20 & 72.40 & 50.94
        & 41.20 & 43.20 & 67.40 & 42.85
        & 45.98 & 34.80 & 49.59 & 40.28 
        & 59.99 & 50.14 & 70.33 & 67.78\\

        CONVERT
        & \underline{55.57} & 50.58 & \underline{74.07} & \textbf{72.92}
        & 41.62 & 42.77 & 68.43 & 62.39
        & 48.44 & 34.25 & 51.30 & 42.17
        & 63.04 & \underline{55.20} & \underline{74.24} & 69.13\\

        AGC-DRR
        & 18.74 & 14.80 & 40.62 & 31.23
        & 40.28 & \underline{45.34} & 68.32 & \textbf{64.82}
        & \underline{49.31} & 35.93 & 55.72 & \underline{42.61}
        & 60.43 & 48.20 & 68.44 & 62.63\\

        RDGAE
        & 55.30 & \underline{53.00} & 73.10 & 48.32
        & 43.70 & \textbf{45.70} & \textbf{69.50} & 43.54
        & 42.64 & \underline{37.34} & \underline{56.72} & 35.77
        & 48.12 & 33.14 & 51.92 & 42.39\\
        
        \hline

        \rowcolor{gray!20}
        \textbf{\model{}} 
        & \textbf{57.96} & \textbf{55.47} & \textbf{75.22} & \underline{72.63} 
        & \textbf{44.34} & 45.01 & \underline{69.34} & \underline{64.29}
        & \textbf{52.10} & \textbf{45.64} & \textbf{58.78} & \textbf{43.17}
        & \textbf{70.13} & \textbf{62.50} & \textbf{80.55} & \textbf{76.55} \\

        \hline

        \textbf{Improv.(\%)}
        & $\uparrow$\textbf{4.3}\% & $\uparrow$\textbf{4.7}\% & $\uparrow$\textbf{1.6}\% & $\downarrow$0.4\%
        & $\uparrow$\textbf{0.5}\% & $\downarrow$1.5\% & $\downarrow$0.2\% & $\downarrow$0.8\%
        & $\uparrow$\textbf{5.7}\% & $\uparrow$\textbf{22.2}\% & $\uparrow$\textbf{3.6}\% & $\uparrow$\textbf{1.3}\%
        & $\uparrow$\textbf{10.7}\% & $\uparrow$\textbf{13.2}\% & $\uparrow$\textbf{8.5}\% & $\uparrow$\textbf{7.7}\%\\

        \bottomrule
    \end{tabular}
    \vspace{-4mm}
\end{table*}

\vspace{-3mm}
\subsection{Optimization}
\label{subsec: optimization}
The entire process of graph clustering in ~\model{} is illustrated in Appendix~\ref{ap: algo} Algorithm~\ref{algo: DeSE}. 
Initially, the original graph is enhanced as detailed in Section~\ref{subsec: SLL}, followed by a round of GNN propagation on the new weighted adjacency matrix $W$, transforming the sparse and high-dimensional feature vectors $X$ into the initial node embeddings $E$. 
Next, several ASS (Section~\ref{subsec: ASS}) modules are used to learn soft assignment matrices $\{S^k\}$ at different layers. 
We employ soft assignment structural entropy (SE loss) and negative sampling cross-entropy loss (CE loss) to optimize the graph clustering task.
Let the set of positive and negative edges be denoted as $\mathcal{E}^{'}$, with an equal number of positive and negative edges. 
The CE loss is then calculated as follows:
\begin{equation}
\begin{split}
    & {p}_{i,j} = Sigmoid(2-{||e_i-e_j||}_2), \quad {l}_{i,j} = \left\{
    \begin{array}{ll}
        1, \quad W_{i,j} \neq 0 \\
        0, \quad else
    \end{array}
    \right. ,\\
    & \mathcal{L}_{ce} = -\frac{1}{|W|}\sum_{(i,j) \in \mathcal{E}^{'}}{{l}_{i,j} log({p}_{i,j}) + (1-{l}_{i,j})log(1-{p}_{i,j})},
    \label{Eq: loss_ce}
\end{split}  
\end{equation}
where $p_{i,j}$ represents the probability of an edge existing between node $v_i$ and node $v_j$, calculated based on the distance between their embeddings. 
Let $l_{i,j}$ denote the ground truth label indicating whether an edge actually exists between them.
The final loss is composed of SE loss, as calculated in Section~\ref{subsec: structural quantification}, and CE loss:
\begin{equation}
    \mathcal{L} = \lambda_{se} H^\mathcal{T}_{sa} + \lambda_{ce} \mathcal{L}_{ce},
    \label{Eq: loss}
\end{equation}
where $\lambda_{se}$ and $\lambda_{ce}$ are hyperparameters of coefficients for SE loss and CE loss, respectively.

\vspace{-3mm}
\subsection{Complexity Analysis}
We analyze the time complexity of each component of~\model{}.
For SLL, the time complexity is mainly contributed by MLP and KNN, which are $O(Nfd)$ and $O(NlogN)$, respectively.
In the ASS, the complexities of embedding learner and soft assignment learner are $O(Nd^2)$ and $O(Nd^2+Nd)$.
For loss computation, the complexity is $O(Nc)$ for SE loss and $O(2Md)$ for CE loss.
Since the number of clusters $c$ is usually small, the total time complexity of the model can be summarized as $O((d+logN+f)Nd+Md)$, which can be simplified to $O(dNlogN)$.

\vspace{-3mm}

\section{Experiments}
\label{sec: experiments}
In this section, we conduct empirical experiments to demonstrate the effectiveness of the proposed framework~\model{}. 
We aim to answer five research questions as follows:
Q1: How effective the ~\model{} is for unsupervised graph clustering, and what kind of clusters are learned compared with baselines (Section~\ref{sec: main result})?
Q2: How do the Structure Learning Layer and SE loss influence the performance of~\model{} (Section~\ref{sec: ablation})?
Q3: How do key hyperparameters impact the performance of~\model{} (Section~\ref{sec: hyperparameter})?
Q4: How robust is~\model{} to the number of clusters, and what kind of graph structure is learned (Section~\ref{sec: cluster num})?

\vspace{-4mm}
\subsection{Experiment Setup}
\textbf{Datasets.}
\begin{table}[t]
    \centering
    \caption{Statistics of four datasets.}
    \label{tab: dataset}
    \vspace{-4mm}
    \renewcommand\arraystretch{0.9}
    \setlength{\tabcolsep}{1.2mm}
    \begin{tabular}{c|cccccc}
        \toprule
        Dataset & \#Node & \#Edge & \#Feature & \#Cluster & Sparsity & Iso.\\
        \hline
        Cora & 2,708 & 5,278 & 1,433 & 7 & 3.9 & 0\\
        Citeseer & 3,327 & 4,552 & 3,703 & 6 & 2.7 & 48\\
        Computer & 13,752 & 245,861 & 767 & 10 & 35.7 & 281\\
        Photo & 7,650 & 119,081 & 745 & 8 & 31.1 & 115\\
        \bottomrule
    \end{tabular}
    \vspace{-6mm}
\end{table}
We conduct experiments on four benchmark datasets: Cora, Citeseer, Computer, and Photo.
Details of datasets are summarized in Table~\ref{tab: dataset} of Appendix~\ref{ap: dataset}, supplemented with the number of independent nodes "Iso." and the average number of edges per node "Sparsity". \par

\noindent \textbf{Baselines.}
For graph clustering, we mainly compare ~\model{} with two categories of methods, including three structure-driven models (i.e., DMoN~\cite{tsitsulin2023graph}, MinCut~\cite{bianchi2020spectral}, and DGI~\cite{velivckovic2018deep}) and five unsupervised GSL models (i.e., SUBLIME~\cite{liu2022towards}, EGAE~\cite{zhang2022embedding}, CONVERT~\cite{yang2023convert}, AGC\-DRR~\cite{gong2022attributed}, and RDGAE~\cite{mrabah2022rethinking}).
Details of baselines are summarized in Appendix~\ref{ap: baseline}.
The codes for all baseline models and~\model{}, along with all datasets, are publicly accessible on GitHub\footnote{\url{https://github.com/SELGroup/DeSE}}. \par

\noindent \textbf{Evaluation Metrics.}
We evaluate the accuracy and consistency of graph clustering with four metrics.
NMI (Normalized Mutual Information) evaluates how well the predicted clusters match the true clusters in terms of information shared.
ARI (Adjusted Rand Index) assesses the similarity between the predicted and true cluster assignments, adjusting for random chance.
ACC (Accuracy) measures the proportion of nodes correctly assigned to their true clusters.
F1 Score evaluates the balance between precision and recall in cluster assignments.

\vspace{-4mm}
\subsection{Graph Clustering Performance}
\label{sec: main result}
Table~\ref{tab: results} reports the graph clustering results of our method ~\model{} in comparison with eight baseline models across four datasets. 
The evaluation metrics include NMI, ARI, ACC, and F1.
For all methods,  both the original graph structure and node features from the datasets are used as input.
The baseline models are drawn from open-source implementations.
As can be observed, despite the absence of labeled data, our proposed~\model{} model outperforms all baselines on 12 out of the 16 evaluated metrics across the four datasets and ranks second on three of the remaining metrics. 
Notably, the~\model{} model achieves the best performance on the NMI metric in all benchmarks. 
This strong performance is attributed to the novel approach of leveraging deep structural entropy to enhance graph structure learning, guiding adaptive clustering.\par

\begin{figure*}[thp]
	\centering
	\subfigure[\model{}.]{
		\includegraphics[width=0.24\linewidth]{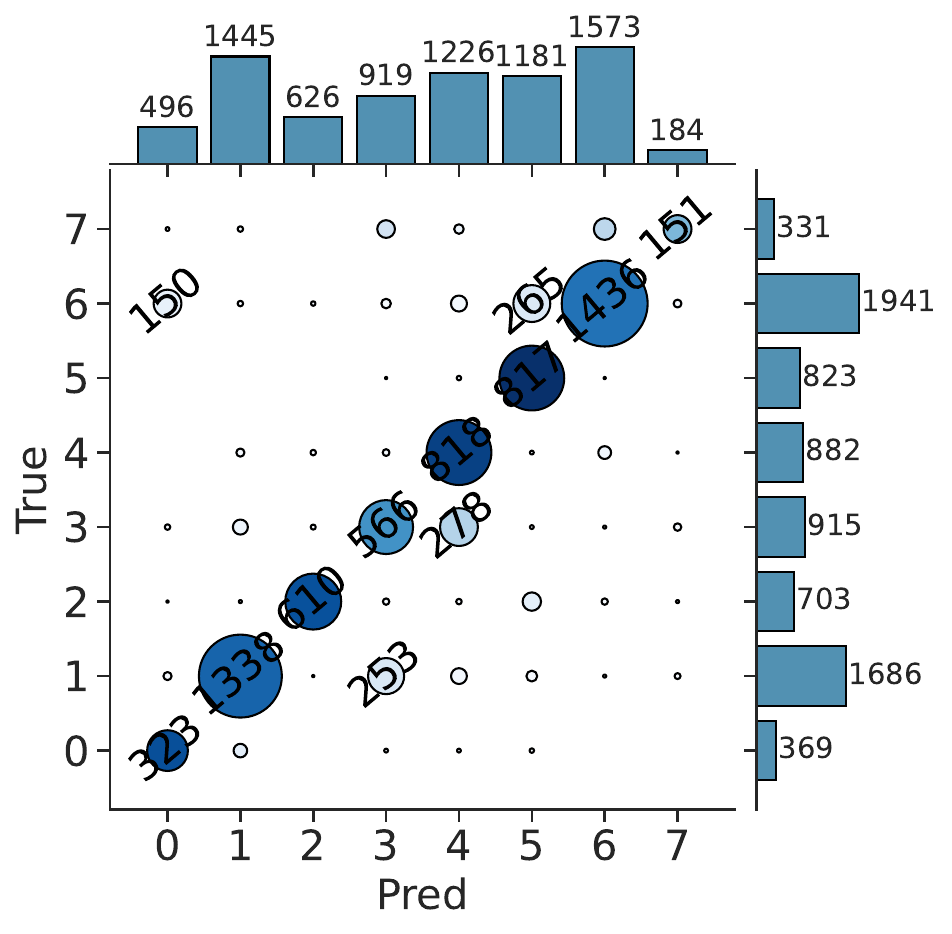}
		\label{fig: cluster_DeSE}
	}\hfill
    \subfigure[MinCut.]{
		\includegraphics[width=0.24\linewidth]{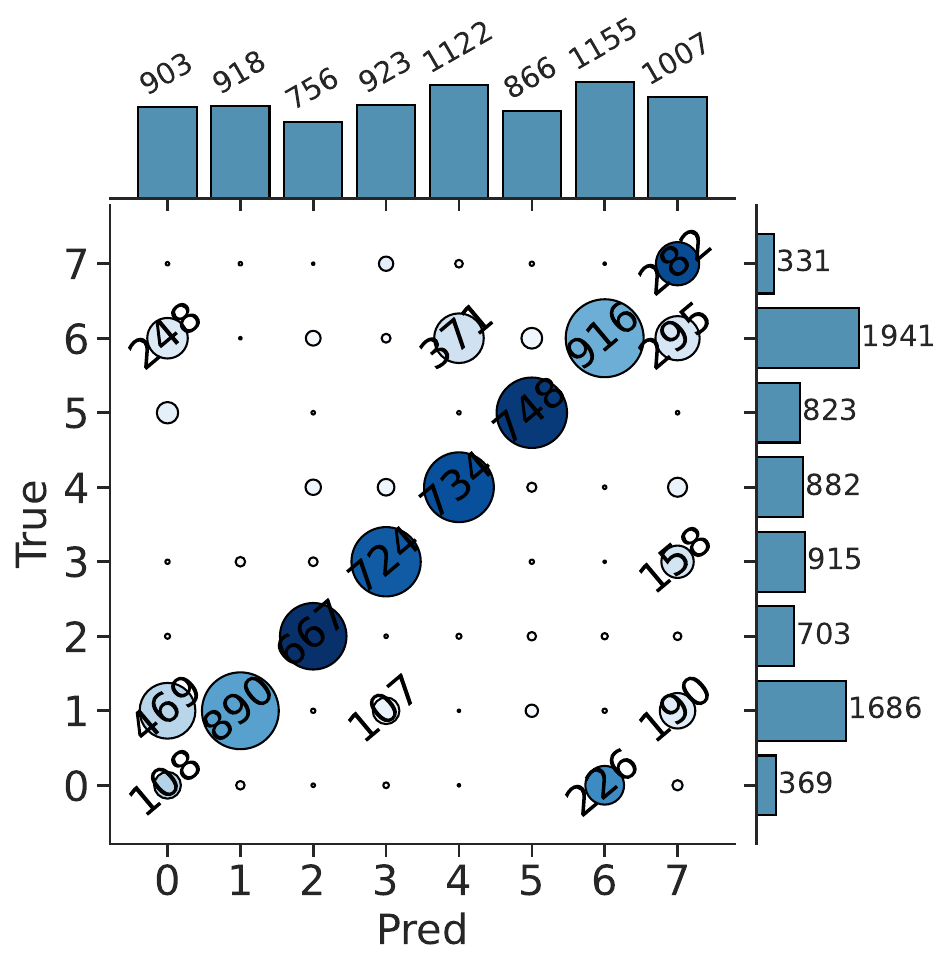}
		\label{fig: cluster_MinCut}
	}\hfill
	\subfigure[EGAE.]{
		\includegraphics[width=0.24\linewidth]{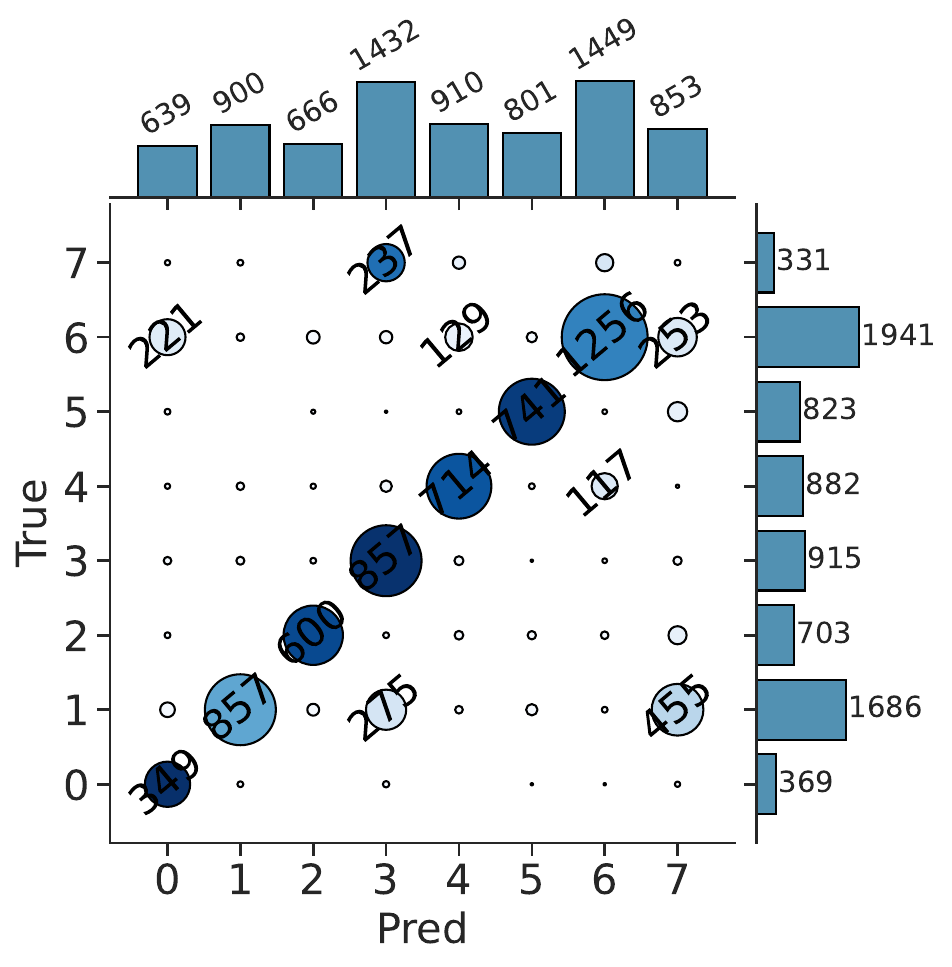}
		\label{fig: cluster_EGAE}
	}\hfill
	\subfigure[RDGAE.]{
		\includegraphics[width=0.24\linewidth]{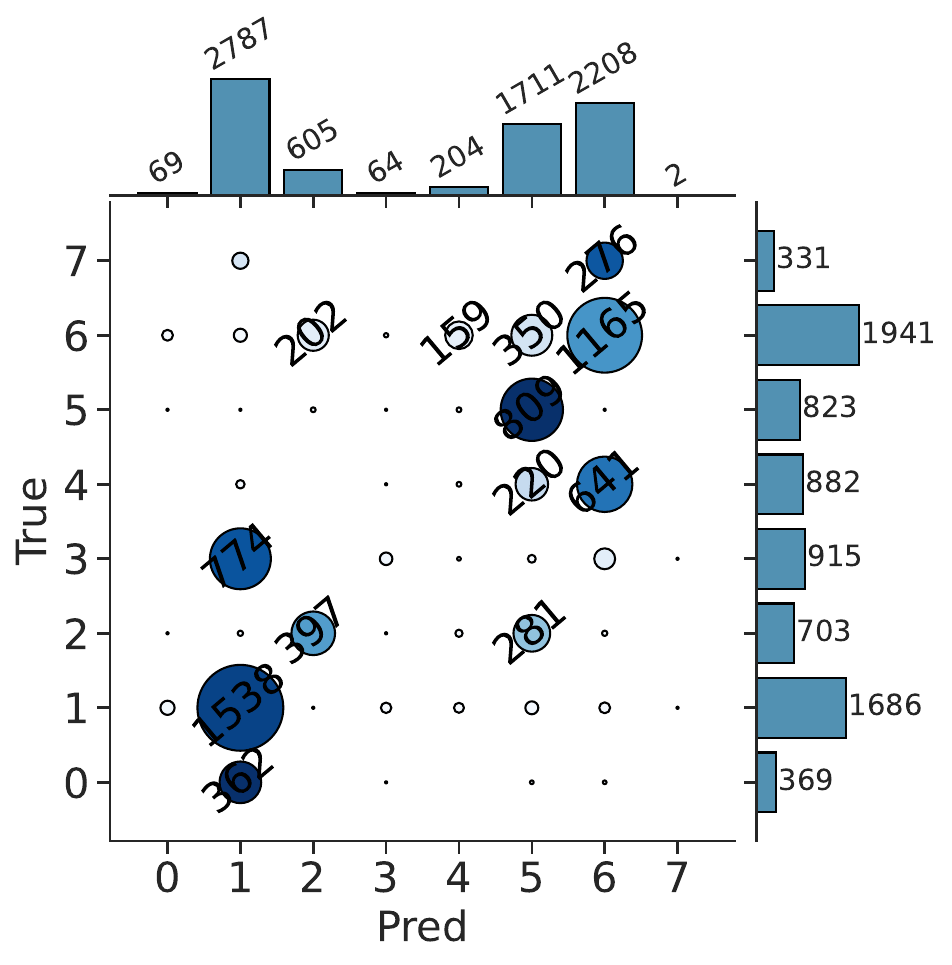}
		\label{fig: cluster_RDGAE}
	}
	\vspace{-5mm}
	\caption{Clusters of~\model{}, EGAE, MinCut, and RDGAE on the Photo dataset. (The vertical axis represents the number of nodes contained in the true clusters, while the horizontal axis represents the number of nodes predicted by the model. Each circle in the heatmap shows the number of nodes from true cluster $i$ predicted to belong to cluster $j$. The circle's size represents the node count, and its color intensity indicates the proportion of these nodes within true cluster $i$, with darker colors showing a higher proportion.) }
	\label{fig: cluster}
	\vspace{-2mm}
\end{figure*}

Additionally, we make other observations:
First, the GSL (Graph Structure Learning) baselines outperform methods that directly use the raw graph structure for clustering on most datasets and metrics, demonstrating the importance of structure learning and structure enhancement. 
Second, while most baselines perform relatively well on NMI and ACC, they struggle to balance ARI and F1. 
For instance, DMoN and MinCut exhibit particularly poor ARI, while EGAE and RDGAE have notably low F1 scores.  
This imbalance can largely be attributed to differences in the prediction of majority and minority classes. 
ACC and NMI tend to focus on overall alignment, whereas F1 and ARI place greater emphasis on local precision, particularly in the handling of imbalanced classes.
In contrast, our~\model{} model does not exhibit such conflicts and performs well across all four metrics.
Figure~\ref{fig: cluster} presents the detailed correspondence between the node numbers of the predicted clusters and the true clusters on~\model{} and three baselines.
It can be observed that RDGAE predicts large clusters with relatively high accuracy but shows noticeable errors for smaller categories, such as the misprediction of the majority of nodes from cluster 0 and cluster 3 into cluster 1. 
In contrast, EGAE and MinCut do not provide comprehensive predictions for the larger clusters, with a significant number of nodes from cluster 1 and cluster 6 being distributed across other clusters.
Our~\model{} model maintains relatively focused and accurate clustering for both large and small clusters.
The clusters of~\model{} plots for the remaining three datasets are discussed in Appendix~\ref{ap: graph clustering performance}.

\vspace{-3mm}
\subsection{Ablation Study}
\label{sec: ablation}
\begin{table}[t]
    \centering
    \caption{Test results corresponding without SLL module and SE loss on four datasets.}
    \vspace{-2mm}
    \label{tab: ablation}
    \renewcommand\arraystretch{1.0}
    \setlength{\tabcolsep}{0.8mm}
    \begin{tabular}{c|cccc|cccc}
        \toprule
        \multirow{2}*{Variation} & \multicolumn{4}{c|}{\multirow{1}*{\textbf{Cora}}} & \multicolumn{4}{c}{\multirow{1}*{\textbf{Citeseer}}} \\
        
        \cline{2-9}

        & NMI & ARI & ACC & \multicolumn{1}{c|}{F1}   
        & NMI & ARI & ACC & F1 \\

        \hline
        
        w/o SLL 
        & 54.21 & 47.72 & - & -
        & 43.00 & 43.56 & 52.63 & 45.69 \\
        
        w/o SE 
        & 35.19 & 24.87 & 47.27 & 39.43
        & 24.65 & 19.25 & 42.14 & 34.13 \\

        \rowcolor{gray!20}
        \model{} 
        & \textbf{57.96} & \textbf{55.47} & \textbf{75.22} & \textbf{72.63}
        & \textbf{44.34} & \textbf{45.01} & \textbf{69.34} & \textbf{64.29} \\

        \hline

        \multirow{2}*{Variation} & \multicolumn{4}{c|}{\multirow{1}*{\textbf{Computer}}} & \multicolumn{4}{c}{\multirow{1}*{\textbf{Photo}}} \\
        
        \cline{2-9}

        & NMI & ARI & ACC & \multicolumn{1}{c|}{F1}   
        & NMI & ARI & ACC & F1 \\

        \hline

        w/o SLL 
        & 51.18 & 33.13 & 55.65 & 43.18 
        & 70.09 & 62.29 & 79.67 & 73.23 \\

        w/o SE  
        & 39.19 & 34.21 & - & - 
        & 53.05 & 37.50 & - & - \\

        \rowcolor{gray!20}
        \model{}  
        & \textbf{52.10} & \textbf{45.64} & \textbf{58.79} & \textbf{43.17}
        & \textbf{70.13} & \textbf{62.50} & \textbf{80.55} & \textbf{76.55} \\
        
        \bottomrule
    \end{tabular}
    \vspace{-6mm}
\end{table}
In our proposed~\model{}, the Structural Learning Layer refines the graph structure, and SE loss with soft assignment is introduced for optimization.
To evaluate the effectiveness of these two components, we independently disabled the SLL and SE loss (i.e., set $\beta_f=0$ and $\lambda_{se}=0$), resulting in the variations referred to as "w/o SLL" and "w/o SE". 
The corresponding results are presented in Table~\ref{tab: ablation}.
Without the SLL, the clustering performance deteriorates more significantly on datasets with relatively sparse connections, such as Cora and Citeseer. 
This is especially evident in the ACC and F1 scores on Cora, where they become undefined, indicating a mismatch between the predicted and actual number of clusters. 
This suggests that the SLL improves the quality of the graph structure, particularly for sparse graphs.
Similarly, in the absence of SE loss, clustering performance declines sharply across all datasets, with large and relatively dense datasets such as Computer and Photo showing undefined ACC and F1 scores due to the mismatch between the predicted and actual number of clusters. 
This highlights the importance of optimizing the model using quantified structural information to improve clustering performance.
\begin{figure}[t]
	\centering
    \includegraphics[width=1.0\linewidth]{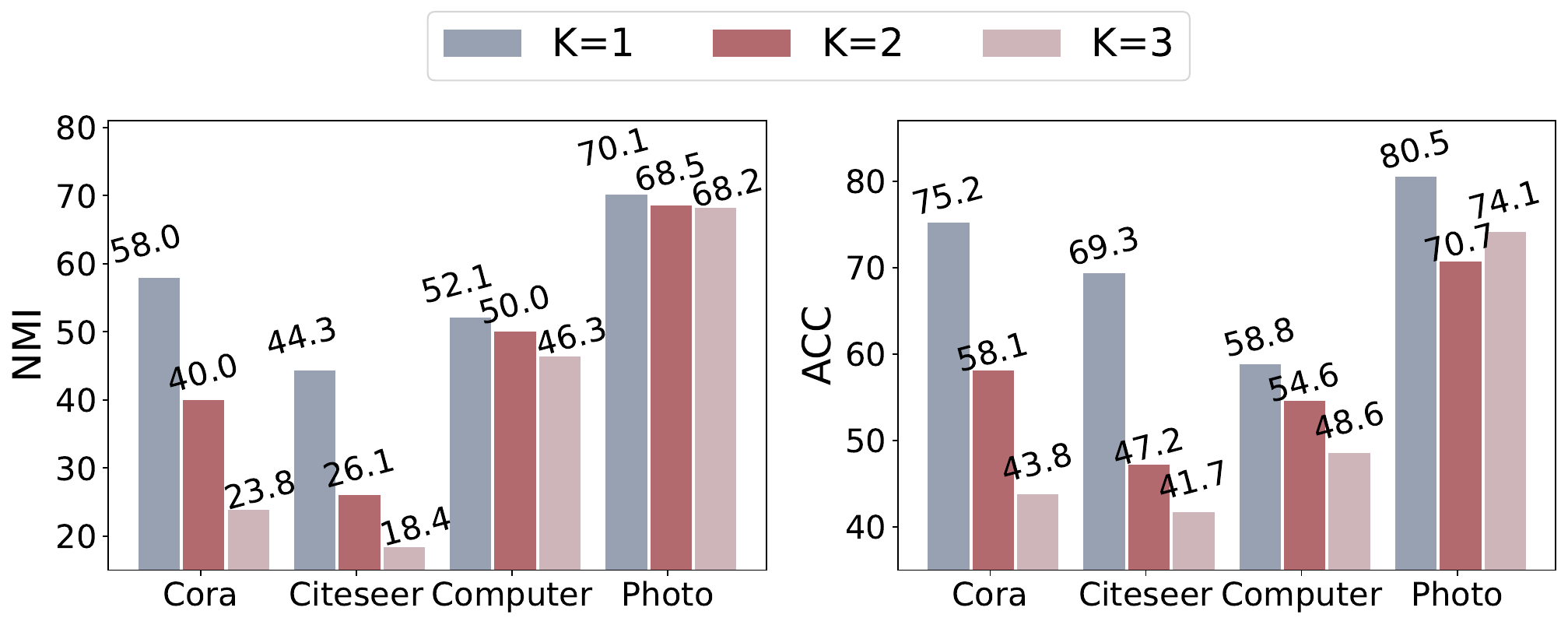}
	\vspace{-7mm}
	\caption{Sensitivity of hyperparameter $K$ with NMI and ACC. }
	\label{fig: hyper_k}
	\vspace{-5mm}
\end{figure}

\vspace{-3mm}
\subsection{Sensitivity Analysis}
\label{sec: hyperparameter}
In this section, we investigate the sensitivity of critical hyperparameter in ~\model{}, including the number of neighbors in KNN $K$, the weight of attribute graph $\beta_f$, and the coefficients of SE loss and CE loss $\lambda_{se}$ and $\lambda_{ce}$. \par

\begin{table*}[t]
    \centering
    \caption{Performance of ~\model{} on different degree-based distributions for $K$.}
    \vspace{-4mm}
    \label{tab: distribution k}
    \renewcommand\arraystretch{1.0}
    \setlength{\tabcolsep}{1.4mm}
    \begin{tabular}{c|cccc|cccc|cccc|cccc}
        \toprule
        \multirow{2}*{Method} & \multicolumn{4}{c|}{\multirow{1}*{\textbf{Cora}}} & \multicolumn{4}{c|}{\multirow{1}*{\textbf{Citeseer}}} & \multicolumn{4}{c|}{\multirow{1}*{\textbf{Computer}}} & \multicolumn{4}{c}{\multirow{1}*{\textbf{Photo}}}\\
        
        \cline{2-17}
        
        & {\textbf{NMI}} & {\textbf{ARI}} & {\textbf{ACC}} & \multicolumn{1}{c|}{\textbf{F1}} 
        & {\textbf{NMI}} & {\textbf{ARI}} & {\textbf{ACC}} & \multicolumn{1}{c|}{\textbf{F1}} 
        & {\textbf{NMI}} & {\textbf{ARI}} & {\textbf{ACC}} & \multicolumn{1}{c|}{\textbf{F1}} 
        & {\textbf{NMI}} & {\textbf{ARI}} & {\textbf{ACC}} & {\textbf{F1}} \\

        \hline
        
        \rowcolor{gray!20}
        $K=1$
        & \textbf{57.96} & \textbf{55.47} & \textbf{75.22} & \textbf{72.63}
        & \textbf{44.34} & \textbf{45.01} & \textbf{69.34} & \textbf{64.29}
        & \textbf{52.10} & \textbf{45.64} & \textbf{58.78} & \textbf{43.17}
        & \textbf{70.13} & 62.50 & \textbf{80.55} & \textbf{76.55}\\

        $K \sim /5$
        & 54.72 & 49.57 & 73.33 & 68.70
        & 39.98 & 40.38 & 66.39 & 61.62 
        & 39.89 & 26.54 & 46.16 & 29.51
        & 65.79 & 61.42 & - & -\\

        $K \sim /10$
        & 56.57 & 51.44 & 74.29 & 70.21
        & 43.08 & 43.56 & 68.19 & 63.17
        & 45.08 & 32.93 & 48.56 & 34.05
        & 62.31 & 58.91 & 65.47 & 52.67 \\
        
        $K \sim sqrt$
        & 36.20 & 33.74 & 62.81 & 57.55
        & 21.47 & 20.32 & 45.77 & 36.76
        & 35.27 & 23.40 & 43.20 & 23.12
        & 58.76 & 49.59 & 69.68 & 57.92 \\

        $K \sim log$
        & 30.04 & 25.37 & 55.31 & 52.10
        & 20.73 & 19.90 & 44.72 & 35.93
        & 42.96 & 27.47 & - & -
        & 54.49 & 43.75 & 63.07 & 45.62 \\

        $K \sim \hat{}$
        & 57.88 & 52.67 & \textbf{75.22} & 71.07
        & 42.79 & 43.30 & 68.17 & 63.28
        & 45.79 & 32.32 & 56.43 & 30.07
        & 65.06 & \textbf{64.99} & 73.79 & 64.30 \\

        \hline
        $K \sim random$
        & 39.17 & 32.21 & 56.28 & 54.39
        & 22.25 & 20.90 & 45.00 & 39.45
        & 47.86 & 30.00 & 52.07 & 41.47
        & 65.75 & 58.68 & 58.31 & 43.25 \\

        \bottomrule
    \end{tabular}
    \vspace{-6mm}
\end{table*}

\begin{figure*}[t]
	\centering
	\subfigure[Cora.]{
		\includegraphics[width=0.24\linewidth]{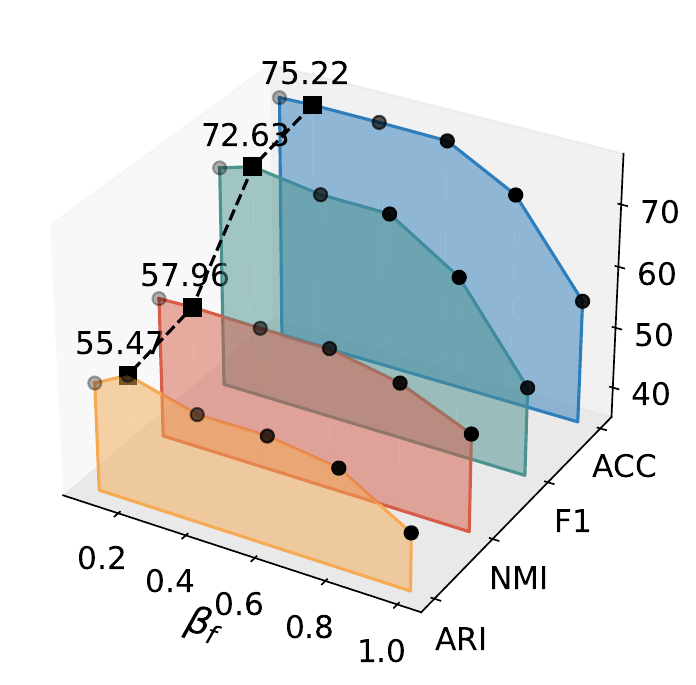}
		\label{fig: hyper_cora}
	}\hfill
	\subfigure[Citeseer.]{
		\includegraphics[width=0.24\linewidth]{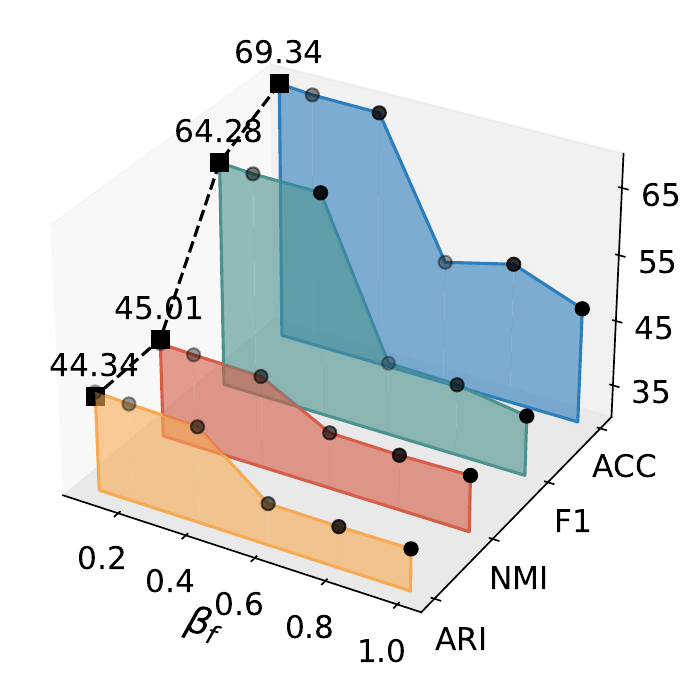}
		\label{fig: hyper_citeseer}
	}\hfill
    \subfigure[Computer.]{
		\includegraphics[width=0.24\linewidth]{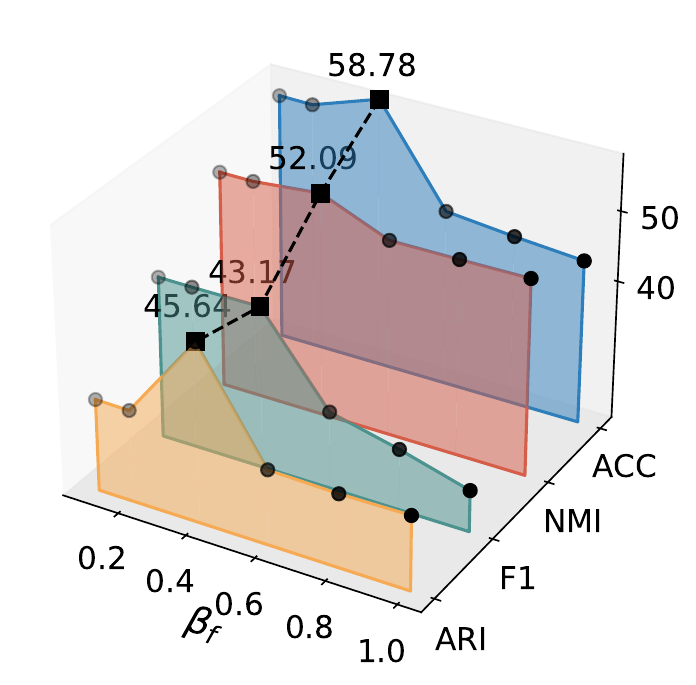}
		\label{fig: hyper_computer}
	}\hfill
    \subfigure[Photo.]{
		\includegraphics[width=0.24\linewidth]{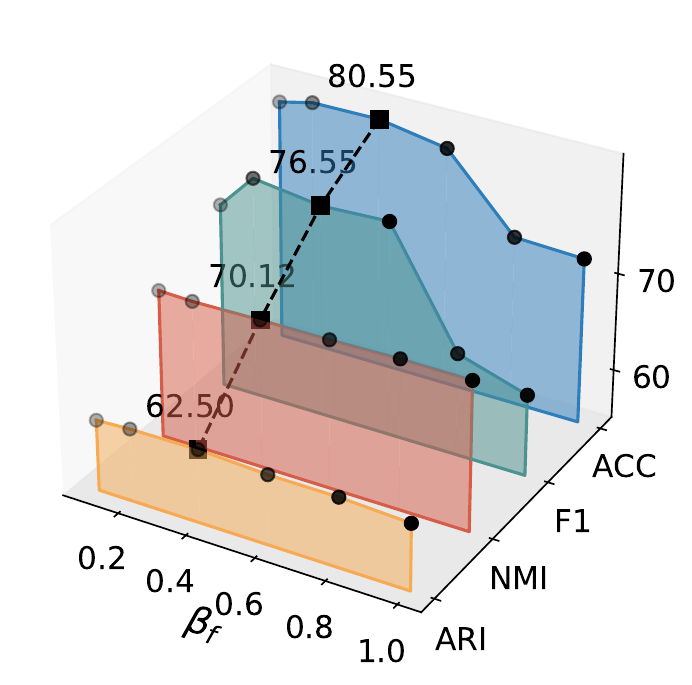}
		\label{fig: hyper_photo}
	}
	\vspace{-5mm}
	\caption{Sensitivity of hyperparameter $\beta_f$ on four datasets with four metrics. }
        \vspace{-4mm}
	\label{fig: hyper_SSL}
\end{figure*}

\textbf{Number of neighbors $K$.}
Figure~\ref{fig: hyper_k} shows the NMI and ACC performance under different numbers of neighbors across four datasets.
It can be observed that the value of $K=1$ yields high NMI and ACC scores.
As the value of $K$ increases, both NMI and ACC show a significant decline in the Cora and Citeseer datasets, which have relatively sparse graph structures. 
However, on the Computer and Photo datasets, which have a higher average number of edges per node, the decrease is less pronounced. 
This suggests that the selection of neighbors has a significant impact on clustering results. 
While improving the graph structure using neighbors offers advantages for clustering, introducing too many neighbors can be detrimental, as it may introduce noise. 
Nodes with similar features in the embedding space do not necessarily belong to the same cluster.
Next, we provide a more detailed explanation of the degree-based distributions used for selecting $K$ and present the complete experimental results in Table~\ref{tab: distribution k}. 
The first row represents the results for $K=1$, which are reported in our paper. 
The subsequent five sets of comparative experiments consider ($EPS=1e-6$):
\begin{itemize}
    \item $K \sim /5$: $K=ceil(\frac{degree}{5} + EPS)$,
    
    \item $K \sim /10$: $K=ceil(\frac{degree}{10} + EPS)$,

    \item $K \sim sqrt$: $K=ceil(\sqrt{degree} + EPS)$,

    \item $K \sim log$: $K=ceil(log_2(degree+1)+EPS)$,

    \item $K \sim \hat{}$: $K=floor(degree^{\frac{1}{degree+1}})$.
    
\end{itemize}

In Table~\ref{tab: distribution k}, a '-' indicates that the number of clusters produced by~\model{} under this configuration does not match the actual number, making it impossible to compute ACC and F1.
We can see that the best performance of~\model{} is still at $K=1$, as too many neighbors distort the graph while both enhancing and preserving it.
Therefore, we chose $K=1$ in our experiments.
Then, to understand whether the introduced edge ($K=1$) is noise or if it represents the missing edge in the original graph, we add a variation "$K \sim random$" where, instead of using KNN, a random edge is added and check its performance.
Results in the last line of Table~\ref{tab: distribution k} show a significant performance drop with random edges compared with $K=1$.
Thus, while~\model{} performs best when $K=1$; we argue that this setting effectively recovers missing but meaningful edges in the original graph.\par

\textbf{Weight of attribute graph $\beta_f$.}
To analyze its sensitivity, we search the value of $\beta_f$ in the range of $\{0.1, 0.2, 0.4, 0.6, 0.8, 1.0\}$.
As is demonstrated in Figure~\ref{fig: hyper_SSL}, the optimal choice of $\beta_f$ varies across datasets. 
For example, the best performance is achieved at $0.4$ for Computer and Photo, $0.2$ for Cora, and $0.1$ for Citeseer. 
However, a common trend is that an excessively large $\beta_f$ leads to poor performance, particularly with F1 and ACC scores, which drop sharply when $\beta_f$ exceeds $0.6$. 
We hypothesize that this occurs because assigning too much weight to the attribute graph may interfere with the effectiveness of the original graph structure.

\begin{table}[t]
    \centering
    \caption{Sensitivity of hyperparameters $\lambda_{se}$ and $\lambda_{ce}$ with NMI.}
    \label{tab: hyper_loss}
    \vspace{-4mm}
    \renewcommand\arraystretch{1.0}
    \setlength{\tabcolsep}{0.8mm}
    \begin{tabular}{c|cccc|cccc}
        \toprule
        \multirow{2}*{Variation} & \multicolumn{4}{c|}{\multirow{1}*{\textbf{SE loss $\lambda_{se}$}}} & \multicolumn{4}{c}{\multirow{1}*{\textbf{CE loss $\lambda_{ce}$}}} \\
        
        \cline{2-9}

        & 0.01 & 0.05 & 0.2 & \multicolumn{1}{c|}{0.5}   
        & 0.1 & 0.5 & 1 & 5 \\

        \hline
        
        Cora 
        & \textbf{57.96} & 56.94 & 55.02 & 54.19
        & 54.24 & 56.82 & 57.02 & \textbf{57.96}\\
        
        Citeseer 
        & \textbf{44.34} & 42.30 & 41.41 & 40.88
        & 44.01 & \textbf{44.34} & 44.13 & 44.26 \\
        
        Computer 
        & 40.74 & 44.45 & \textbf{52.10} & 42.26
        & 46.05 & \textbf{52.10} & 47.31 & 43.10 \\

        Photo 
        & \textbf{70.13} & 69.95 & 63.90 & 66.06
        & 66.08 & 68.67 & 70.13 & \textbf{70.13} \\
        
        \bottomrule
    \end{tabular}
    \vspace{-6mm}
\end{table}
\textbf{Coefficients of SE loss and CE loss $\lambda_{se}$ and $\lambda_{ce}$.}
The NMI results of~\model{} with $\lambda_{se}$ in the range of $\{0.01, 0.05, 0.2, 0.5\}$ and $\lambda_{ce}$ in the range of $\{0.1, 0.5, 1, 5\}$ are presented in Table~\ref{tab: hyper_loss}. 
The optimal choice varies across different datasets, but generally, the SE loss tends to favor smaller coefficients, while the CE loss tends to favor larger coefficients. 
However, compared to the variant w/o SE ($\lambda_{se} = 0$) in section~\ref{sec: ablation}, it is evident that despite the small coefficient of the SE loss, it plays a significant role in improving performance.

\vspace{-3mm}
\subsection{Robustness on Clusters}
\label{sec: cluster num}
\begin{table*}[thp]
    \centering
    \caption{Robustness of cluster numbers on baselines.}
    \label{tab: cluster baseline}
    \vspace{-4mm}
    \renewcommand\arraystretch{1.0}
    \setlength{\tabcolsep}{1.4mm}
    \begin{tabular}{c|c|cccc|cccc|cccc|cccc}
        \toprule
        \multirow{2}*{Method} & \multirow{2}*{Cluster} & \multicolumn{4}{c|}{\multirow{1}*{\textbf{Cora}}} & \multicolumn{4}{c|}{\multirow{1}*{\textbf{Citeseer}}} & \multicolumn{4}{c|}{\multirow{1}*{\textbf{Computer}}} & \multicolumn{4}{c}{\multirow{1}*{\textbf{Photo}}}\\
        
        \cline{3-18}
        
        && {\textbf{NMI}} & {\textbf{ARI}} & {\textbf{ACC}} & \multicolumn{1}{c|}{\textbf{F1}} 
        & {\textbf{NMI}} & {\textbf{ARI}} & {\textbf{ACC}} & \multicolumn{1}{c|}{\textbf{F1}} 
        & {\textbf{NMI}} & {\textbf{ARI}} & {\textbf{ACC}} & \multicolumn{1}{c|}{\textbf{F1}} 
        & {\textbf{NMI}} & {\textbf{ARI}} & {\textbf{ACC}} & {\textbf{F1}} \\

        \hline

        \multirow{2}*{\model{}}
        & +1 
        & 57.40 & 53.50 & 73.52 & 71.20
        & 44.34 & 44.84 & 68.86 & 63.95
        & 51.22 & 33.29 & 55.86 & 43.17
        & 70.13 & 62.50 & 80.55 & 76.55 \\

        & +2 
        & 57.96 & 52.68 & 75.22 & 71.08
        & 41.98 & 43.42 & 68.68 & 63.57
        & 49.20 & 31.15 & 56.25 & 29.58
        & 68.77 & 59.57 & 75.76 & 65.94 \\

        \hline

        \multirow{2}*{DMoN}
        & +1 
        & 35.18 & 24.49 & - & -
        & 15.73 & 10.65 & - & -
        & 43.93 & 23.22 & - & -
        & 55.03 & 38.34 & - & - \\

        & +2 
        & 42.72 & 29.20 & - & -
        & 20.84 & 18.39 & - & -
        & 44.61 & 22.11 & - & -
        & 56.67 & 40.21 & - & - \\

        \hline

        \multirow{2}*{MinCut}
        & +1 
        & 38.80 & 26.92 & - & -
        & 20.79 & 15.98 & - & -
        & 33.13 & 27.50 & - & -
        & 59.27 & 44.09 & - & - \\

        & +2 
        & 33.27 & 20.04 & - & -
        & 22.57 & 18.26 & - & -
        & 28.46 & 25.34 & - & -
        & 56.79 & 41.80 & - & - \\

        \hline
        
        \multirow{2}*{DGI}
        & +1 
        & 53.62 & 44.70 & - & -
        & 38.87 & 38.62 & - & -
        & 25.98 & 15.10 & - & -
        & 30.60 & 17.71 & - & - \\

        & +2 
        & 56.05 & 47.72 & - & -
        & 38.12 & 37.53 & - & -
        & 32.42 & 18.22 & - & -
        & 23.55 & 11.36 & - & - \\

        \bottomrule
    \end{tabular}
    \vspace{-4mm}
\end{table*}

\begin{figure}[t]
	\centering
	\begin{minipage}[b]{0.3\linewidth}
		\centering
		\includegraphics[width=\linewidth]{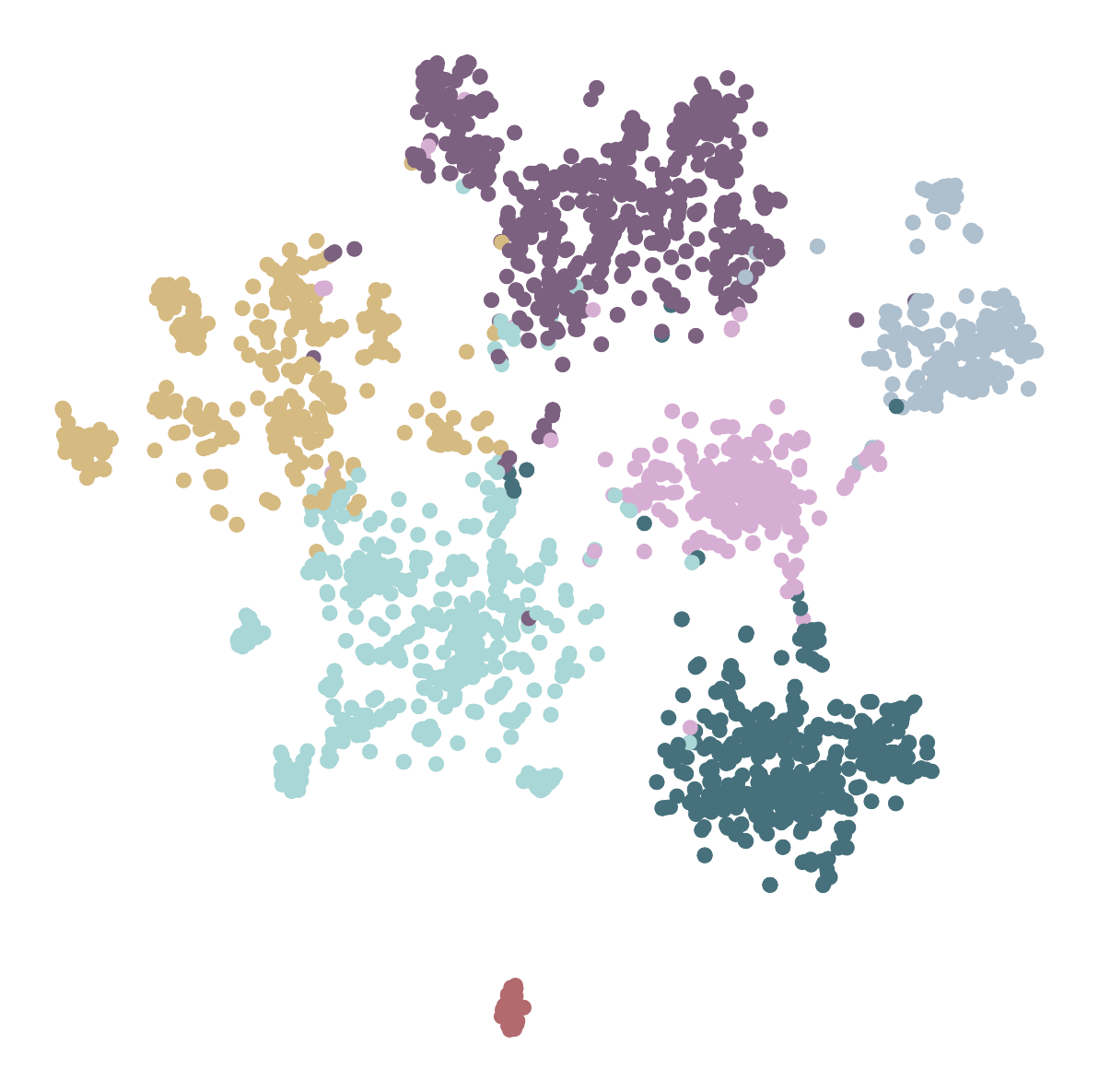}
        \vspace{-8mm}
		\caption*{\footnotesize\mdseries (a) Cora 7 (NMI=56.26).}
	\end{minipage}\hfill
	\begin{minipage}[b]{0.3\linewidth}
		\centering
		\includegraphics[width=\linewidth]{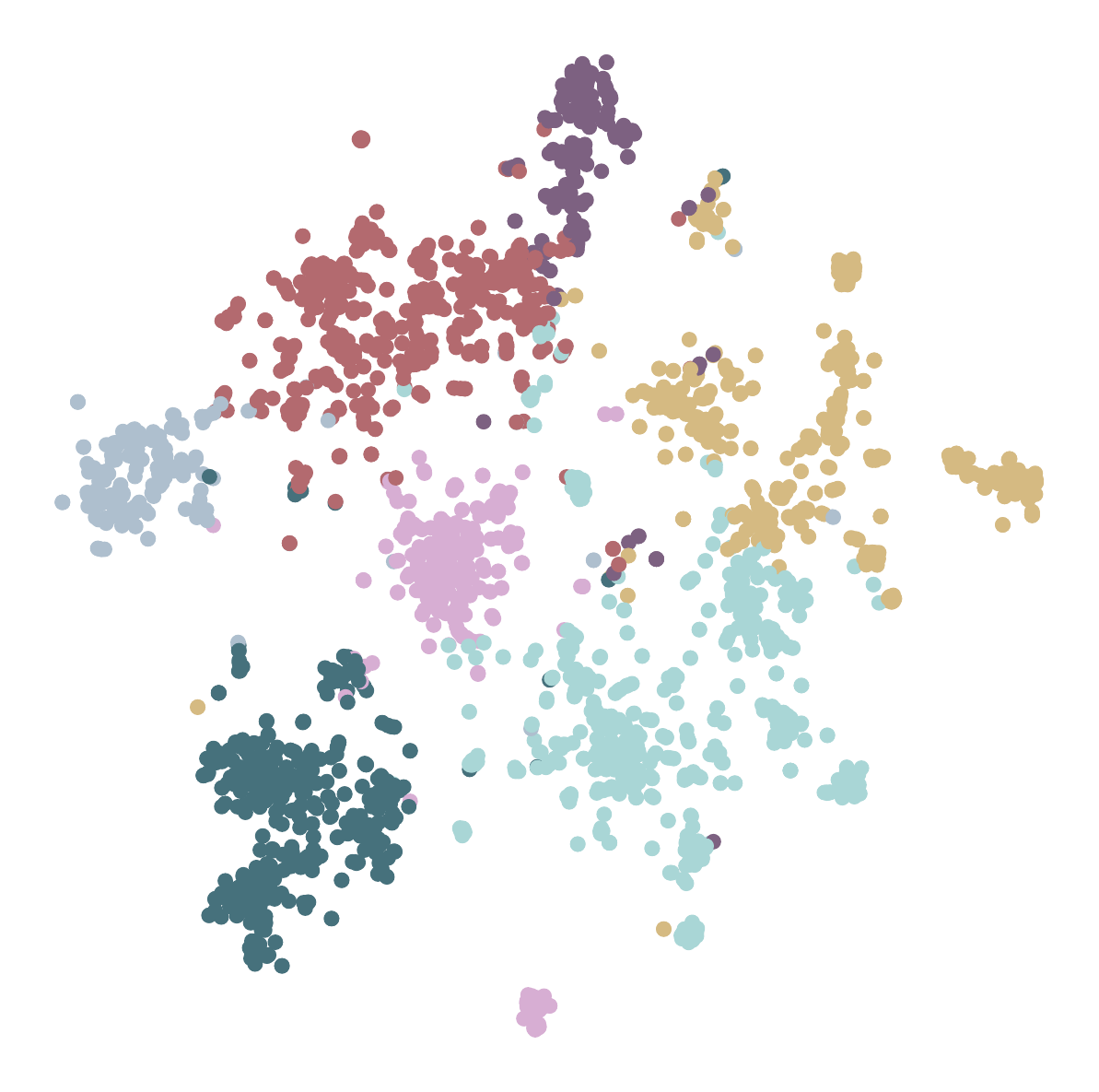}
        \vspace{-8mm}
		\caption*{\footnotesize\mdseries (c) Cora 8 (NMI=57.40).}
	\end{minipage}\hfill
	\begin{minipage}[b]{0.3\linewidth}
		\centering
		\includegraphics[width=\linewidth]{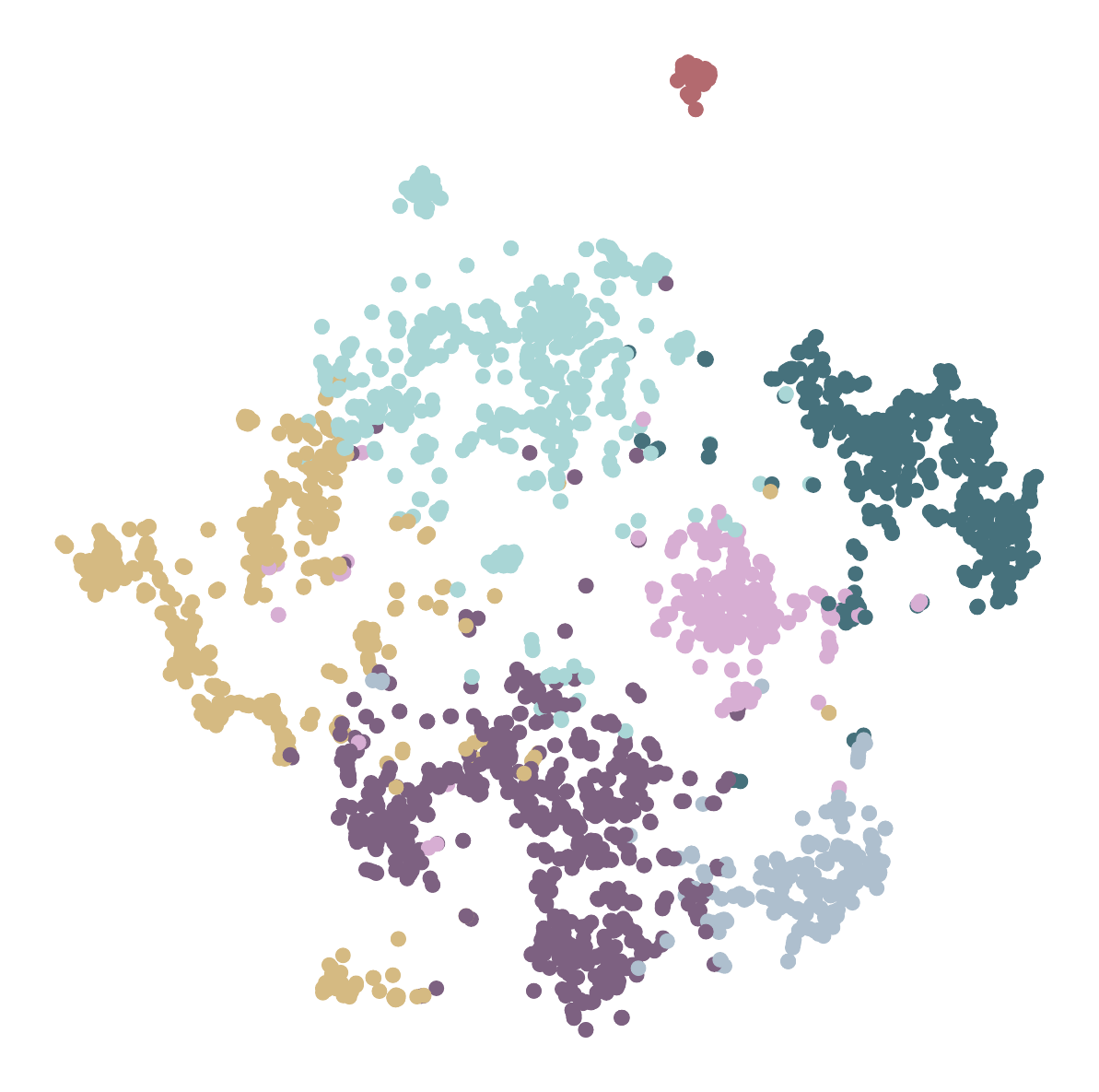}
        \vspace{-8mm}
		\caption*{\footnotesize\mdseries (e) Cora 9 (NMI=57.96).}
	\end{minipage}\hfill
 
	\begin{minipage}[b]{0.3\linewidth}
		\centering
		\includegraphics[width=\linewidth]{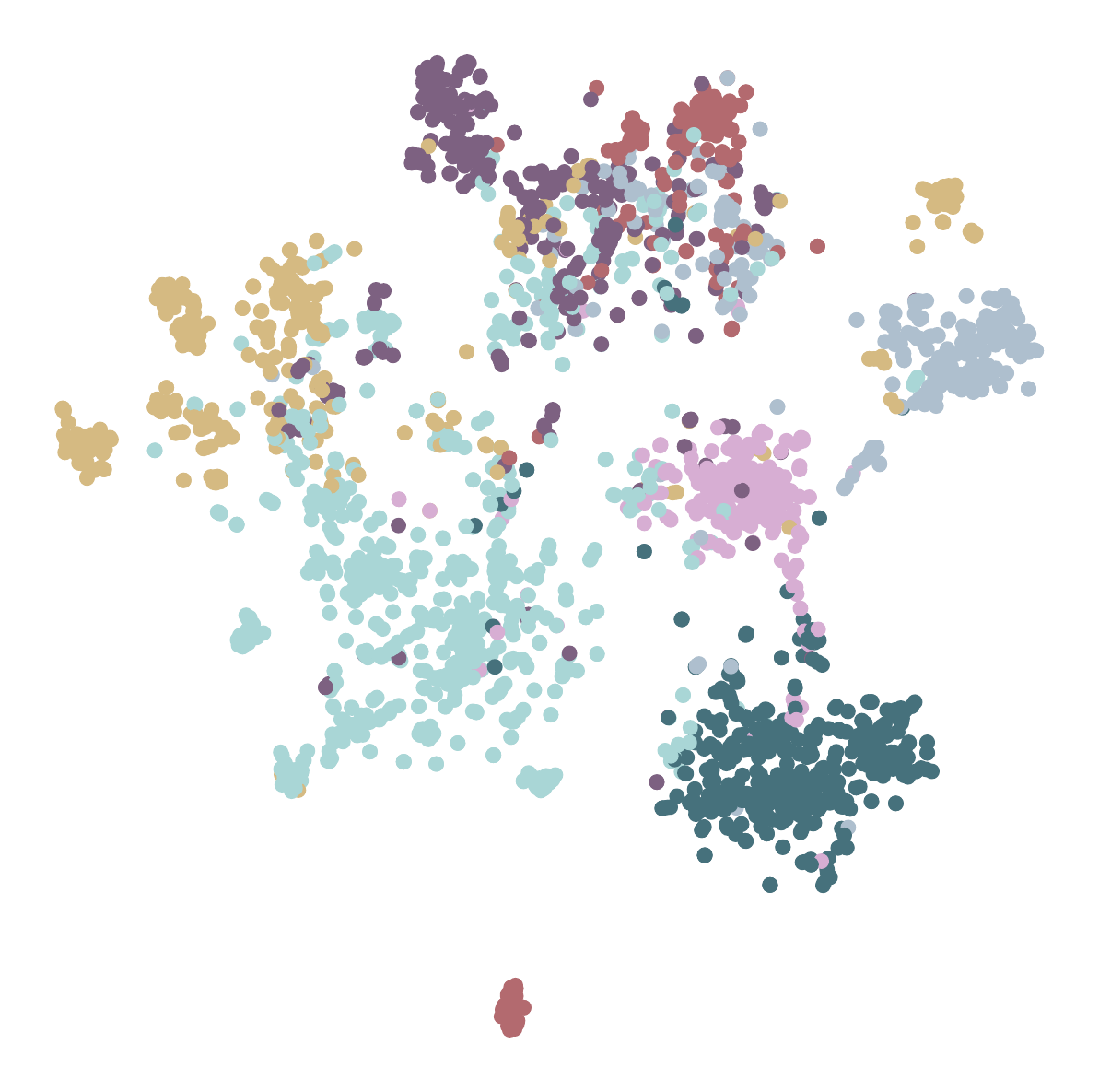}
        \vspace{-8mm}		
        \caption*{\footnotesize\mdseries (b) Cora 7 true.}
	\end{minipage}\hfill
	\begin{minipage}[b]{0.3\linewidth}
		\centering
		\includegraphics[width=\linewidth]{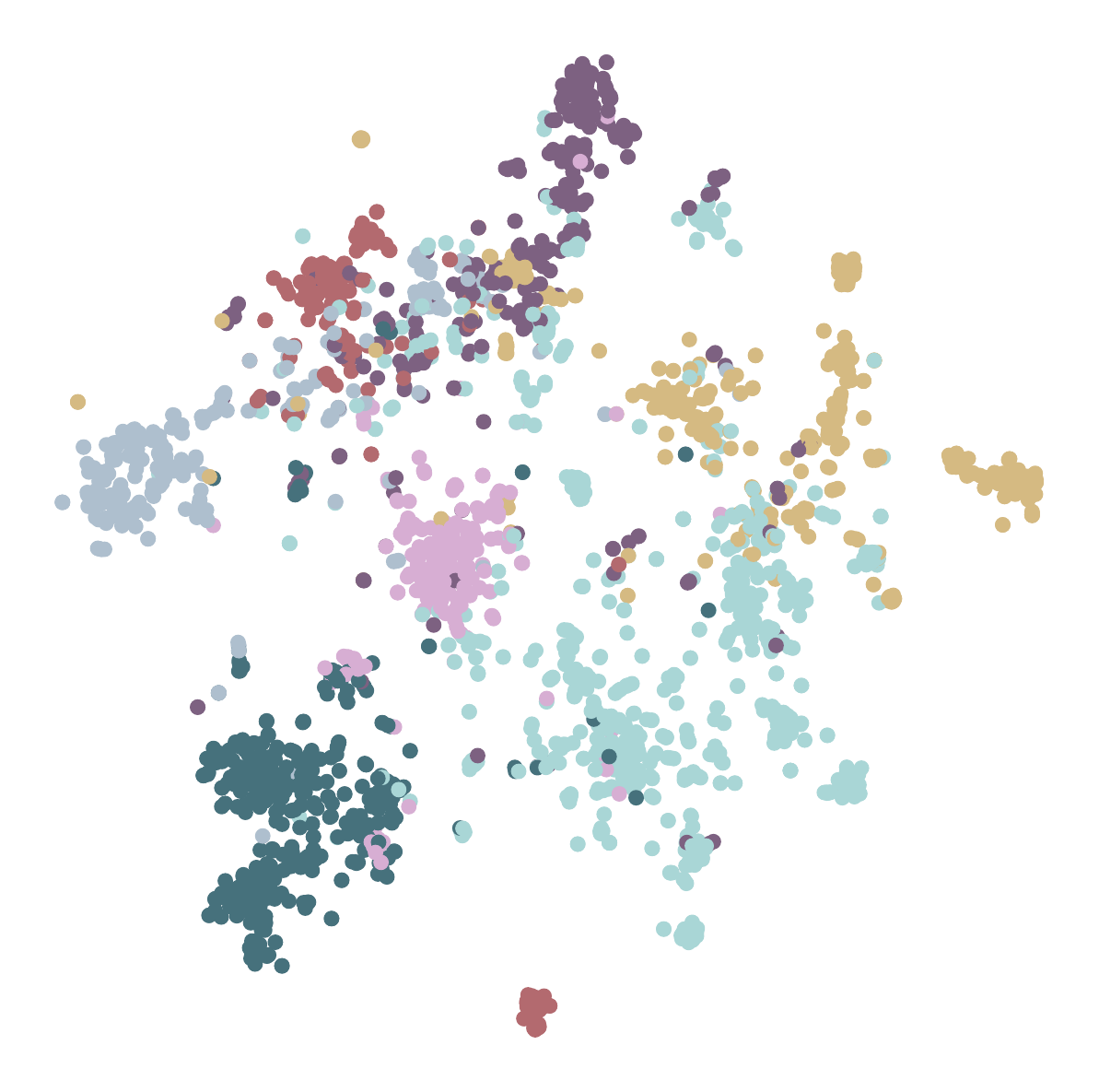}
        \vspace{-8mm}
		\caption*{\footnotesize\mdseries (d) Cora 8 true.}
	\end{minipage}\hfill
	\begin{minipage}[b]{0.3\linewidth}
		\centering
		\includegraphics[width=\linewidth]{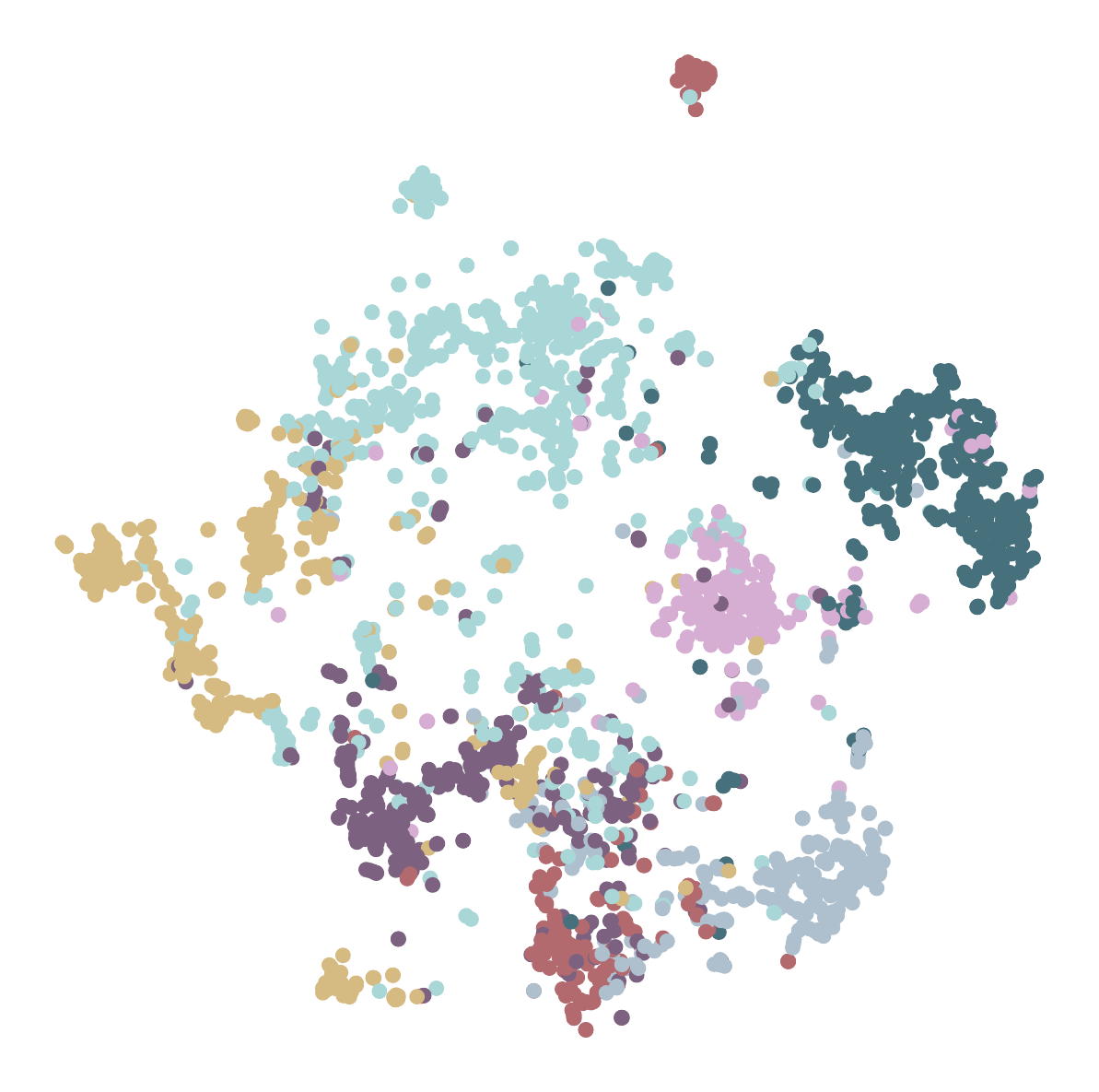}
        \vspace{-8mm}
		\caption*{\footnotesize\mdseries (f) Cora 9 true.}
	\end{minipage}\hfill
	\vspace{-4mm}
	\caption{Robustness of cluster numbers on Cora.}
        \vspace{-6mm}
	\label{fig: num_cluster_cora}
\end{figure}

To evaluate the robustness of ~\model{} under different cluster number settings, we set the number of clusters in ${GNN}_{ass}$ to the original cluster number $c$, $c+1$, and $c+2$, respectively, and visualize the results using t-SNE. 
As shown in Figure~\ref{fig: num_cluster_cora}, the first row presents the predictions of ~\model{} on the Cora dataset, while the second row displays the results under the ground-truth labels.
It can be observed that regardless of the set number of clusters, \model{} consistently forms seven clusters, matching the ground truth, and achieves a relatively high NMI. 
This demonstrates the robustness of our model with respect to cluster numbers. 
Additionally, cluster construction is not entirely dependent on embedding learning. 
The ground-truth distribution shows that distant nodes can belong to the same cluster. 
\model{} is able to capture such nodes to some extent. 
For instance, in Figure~\ref{fig: num_cluster_cora}(e), a small group of yellow nodes in the lower left is closer to the purple cluster in terms of embedding distribution, but~\model{} is still able to identify them correctly.
We attribute this to the use of SE loss and structural learning, which reduces the embedding distance between connected nodes and approaches cluster division from the perspective of overall structural stability.
More analysis on the robustness of the cluster for the remaining three datasets is discussed in Appendix~\ref{ap: robustness on clusters}.\par

We conduct experiments on three baselines for comparison. 
As shown in Table~\ref{tab: cluster baseline}, we report the NMI, ARI, ACC, and F1 for ~\model{}, DMoN, MinCut, and DGI. 
ACC and F1 cannot be tested due to mismatched cluster numbers of baselines.
It can be observed that when the number of clusters generated by the baselines slightly deviates from the ground truth, the performance of NMI and ARI declines sharply and exhibits instability. 
Moreover, due to the mismatch in the number of clusters with the ground truth, the baselines fail to provide ACC and F1 scores, which apply to all the baselines presented.
This highlights their dependency on a predefined number of clusters. 
In contrast, our model~\model{} does not suffer from this limitation. 
It not only maintains strong NMI and ARI performance even when the specified number of clusters deviates from the ground truth but also converges the number of clusters to the appropriate value, thereby achieving high ACC and F1 performance.\par

When the approximate range of the number of clusters is known in advance, as described above, ~\model{} achieves good clustering performance, and the expected number of clusters aligns well with the actual number. 
When the number of clusters is unknown, we can iteratively run ~\model{} to approximate convergence as shown in Table~\ref{tab: case}. 
Using the Cora dataset as a case study, we perform the following steps:
We set the number of clusters to $c=N=2708$, obtaining an NMI of 39.40 with 372 clusters.
We set $c=372$ and repeat the experiment, yielding an NMI of 45.00 with 36 clusters.
We set $c=36$ and repeat the experiment, obtaining an NMI of 51.17 with 14 clusters.
We then set $c=14$ and repeat the experiment, resulting in an NMI of 50.08 with 7 clusters.
We set $c=7$ and repeat the experiment, achieving an NMI of 57.96 with 7 clusters.
At this point, the preset number of clusters matches the output number, and we stop the testing process. \par
\begin{table}[t]
    \centering
    \caption{Case study on Cora when the number of clusters is unknown.}
    \label{tab: case}
    \vspace{-4mm}
    \renewcommand\arraystretch{0.9}
    \setlength{\tabcolsep}{4mm}
    \begin{tabular}{c|cc}
        \toprule
        
        Round & Input: $c$ & Output: NMI, clusters\\
        
        \hline
        
        1 & $c=2708$ & NMI=39.40, clusters=372 \\
        
        \hline
        
        2 & $c=372$ & NMI=45.00, clusters=36\\
        
        \hline
        
        3 & $c=36$ & NMI=51.17, clusters=14 \\
        
        \hline
        
        4 & $c=14$ & NMI=50.08, clusters=7\\
        
        \hline
        
        5 & $c=7$ & NMI=57.96, clusters=7 \\
        
        \bottomrule
    \end{tabular}
    \vspace{-4mm}
\end{table}

\vspace{-2mm}
\section{Conclusion}
\label{sec: conclusion}
This paper presents a novel unsupervised graph clustering framework, ~\model{}, which incorporates deep structural entropy. 
The proposed framework addresses the challenges of structural quantification and structural learning to enhance clustering performance.
We introduce a method for calculating structural entropy with soft assignment and design a Structural Learning Layer to optimize the original graph based on node features. 
Additionally, the Clustering Assignment Layer jointly learns node embeddings and a soft assignment matrix to derive node clusters through a new optimization approach that minimizes both SE loss and CE loss. 
Extensive experiments demonstrate the superiority and interoperability of ~\model{} while also showcasing its robustness in determining the number of clusters.
Our findings highlight the potential of structural information theory in graph structure learning and may open new avenues for research on trainable soft assignment structural entropy in the integration of features and structure.

\vspace{-2mm}
\begin{acks}
This work has been supported by NSFC through grants 62322202, 62441612, 62432006, 62202164, U21B2027 and U23A2038, Local Science and Technology Development Fund of Hebei Province Guided by the Central Government of China through grant 246Z0102G, the "Pioneer” and “Leading Goose” R\&D Program of Zhejiang" through grant 2025C02044, National Key Laboratory under grant 241-HF-D07-01, and CCF-DiDi GAIA collaborative Research Funds for Young Scholars through grant 202422, Yunnan Provincial Major Science and Technology Special Plan Projects (202402AG050007, 202303AP140008).
\end{acks}

\clearpage
\bibliographystyle{ACM-Reference-Format}
\bibliography{7_references}


\begin{thebibliography}{48}


\ifx \showCODEN    \undefined \def \showCODEN     #1{\unskip}     \fi
\ifx \showDOI      \undefined \def \showDOI       #1{#1}\fi
\ifx \showISBNx    \undefined \def \showISBNx     #1{\unskip}     \fi
\ifx \showISBNxiii \undefined \def \showISBNxiii  #1{\unskip}     \fi
\ifx \showISSN     \undefined \def \showISSN      #1{\unskip}     \fi
\ifx \showLCCN     \undefined \def \showLCCN      #1{\unskip}     \fi
\ifx \shownote     \undefined \def \shownote      #1{#1}          \fi
\ifx \showarticletitle \undefined \def \showarticletitle #1{#1}   \fi
\ifx \showURL      \undefined \def \showURL       {\relax}        \fi
\providecommand\bibfield[2]{#2}
\providecommand\bibinfo[2]{#2}
\providecommand\natexlab[1]{#1}
\providecommand\showeprint[2][]{arXiv:#2}

\bibitem[\protect\citeauthoryear{Bianchi, Grattarola, and Alippi}{Bianchi et~al\mbox{.}}{2020a}]%
        {bianchi2020spectral}
\bibfield{author}{\bibinfo{person}{Filippo~Maria Bianchi}, \bibinfo{person}{Daniele Grattarola}, {and} \bibinfo{person}{Cesare Alippi}.} \bibinfo{year}{2020}\natexlab{a}.
\newblock \showarticletitle{Spectral clustering with graph neural networks for graph pooling}. In \bibinfo{booktitle}{\emph{International conference on machine learning}}. PMLR, \bibinfo{pages}{874--883}.
\newblock


\bibitem[\protect\citeauthoryear{Bianchi, Grattarola, Livi, and Alippi}{Bianchi et~al\mbox{.}}{2020b}]%
        {bianchi2020hierarchical}
\bibfield{author}{\bibinfo{person}{Filippo~Maria Bianchi}, \bibinfo{person}{Daniele Grattarola}, \bibinfo{person}{Lorenzo Livi}, {and} \bibinfo{person}{Cesare Alippi}.} \bibinfo{year}{2020}\natexlab{b}.
\newblock \showarticletitle{Hierarchical representation learning in graph neural networks with node decimation pooling}.
\newblock \bibinfo{journal}{\emph{IEEE Transactions on Neural Networks and Learning Systems}} \bibinfo{volume}{33}, \bibinfo{number}{5} (\bibinfo{year}{2020}), \bibinfo{pages}{2195--2207}.
\newblock


\bibitem[\protect\citeauthoryear{Cao, Peng, Li, You, Hao, and Yu}{Cao et~al\mbox{.}}{2024a}]%
        {cao2024multi}
\bibfield{author}{\bibinfo{person}{Yuwei Cao}, \bibinfo{person}{Hao Peng}, \bibinfo{person}{Angsheng Li}, \bibinfo{person}{Chenyu You}, \bibinfo{person}{Zhifeng Hao}, {and} \bibinfo{person}{Philip~S Yu}.} \bibinfo{year}{2024}\natexlab{a}.
\newblock \showarticletitle{Multi-Relational Structural Entropy}. In \bibinfo{booktitle}{\emph{Proceedings of the Conference on Uncertainty in Artificial Intelligence}}. \bibinfo{pages}{1--15}.
\newblock


\bibitem[\protect\citeauthoryear{Cao, Peng, Yu, and Yu}{Cao et~al\mbox{.}}{2024b}]%
        {cao2024hierarchical}
\bibfield{author}{\bibinfo{person}{Yuwei Cao}, \bibinfo{person}{Hao Peng}, \bibinfo{person}{Zhengtao Yu}, {and} \bibinfo{person}{Philip~S Yu}.} \bibinfo{year}{2024}\natexlab{b}.
\newblock \showarticletitle{Hierarchical and Incremental Structural Entropy Minimization for Unsupervised Social Event Detection}. In \bibinfo{booktitle}{\emph{Proceedings of the AAAI conference on artificial intelligence}}. \bibinfo{pages}{1--13}.
\newblock


\bibitem[\protect\citeauthoryear{Chen, Perozzi, Hu, and Skiena}{Chen et~al\mbox{.}}{2018}]%
        {chen2018harp}
\bibfield{author}{\bibinfo{person}{Haochen Chen}, \bibinfo{person}{Bryan Perozzi}, \bibinfo{person}{Yifan Hu}, {and} \bibinfo{person}{Steven Skiena}.} \bibinfo{year}{2018}\natexlab{}.
\newblock \showarticletitle{Harp: Hierarchical representation learning for networks}. In \bibinfo{booktitle}{\emph{Proceedings of the AAAI conference on artificial intelligence}}, Vol.~\bibinfo{volume}{32}.
\newblock


\bibitem[\protect\citeauthoryear{Ding, Wang, Yang, and Liu}{Ding et~al\mbox{.}}{2023}]%
        {ding2023eliciting}
\bibfield{author}{\bibinfo{person}{Kaize Ding}, \bibinfo{person}{Yancheng Wang}, \bibinfo{person}{Yingzhen Yang}, {and} \bibinfo{person}{Huan Liu}.} \bibinfo{year}{2023}\natexlab{}.
\newblock \showarticletitle{Eliciting structural and semantic global knowledge in unsupervised graph contrastive learning}. In \bibinfo{booktitle}{\emph{Proceedings of the AAAI Conference on Artificial Intelligence}}, Vol.~\bibinfo{volume}{37}. \bibinfo{pages}{7378--7386}.
\newblock


\bibitem[\protect\citeauthoryear{Fatemi, El~Asri, and Kazemi}{Fatemi et~al\mbox{.}}{2021}]%
        {fatemi2021slaps}
\bibfield{author}{\bibinfo{person}{Bahare Fatemi}, \bibinfo{person}{Layla El~Asri}, {and} \bibinfo{person}{Seyed~Mehran Kazemi}.} \bibinfo{year}{2021}\natexlab{}.
\newblock \showarticletitle{Slaps: Self-supervision improves structure learning for graph neural networks}.
\newblock \bibinfo{journal}{\emph{Advances in Neural Information Processing Systems}}  \bibinfo{volume}{34} (\bibinfo{year}{2021}), \bibinfo{pages}{22667--22681}.
\newblock


\bibitem[\protect\citeauthoryear{Gong, Zhou, Tu, and Liu}{Gong et~al\mbox{.}}{2022}]%
        {gong2022attributed}
\bibfield{author}{\bibinfo{person}{Lei Gong}, \bibinfo{person}{Sihang Zhou}, \bibinfo{person}{Wenxuan Tu}, {and} \bibinfo{person}{Xinwang Liu}.} \bibinfo{year}{2022}\natexlab{}.
\newblock \showarticletitle{Attributed Graph Clustering with Dual Redundancy Reduction.}. In \bibinfo{booktitle}{\emph{IJCAI}}. \bibinfo{pages}{3015--3021}.
\newblock


\bibitem[\protect\citeauthoryear{Jin, Ma, Liu, Tang, Wang, and Tang}{Jin et~al\mbox{.}}{2020}]%
        {jin2020graph}
\bibfield{author}{\bibinfo{person}{Wei Jin}, \bibinfo{person}{Yao Ma}, \bibinfo{person}{Xiaorui Liu}, \bibinfo{person}{Xianfeng Tang}, \bibinfo{person}{Suhang Wang}, {and} \bibinfo{person}{Jiliang Tang}.} \bibinfo{year}{2020}\natexlab{}.
\newblock \showarticletitle{Graph structure learning for robust graph neural networks}. In \bibinfo{booktitle}{\emph{Proceedings of the 26th ACM SIGKDD international conference on knowledge discovery \& data mining}}. \bibinfo{pages}{66--74}.
\newblock


\bibitem[\protect\citeauthoryear{Kang, Peng, Cheng, Liu, Peng, Xu, and Tian}{Kang et~al\mbox{.}}{2021}]%
        {kang2021structured}
\bibfield{author}{\bibinfo{person}{Zhao Kang}, \bibinfo{person}{Chong Peng}, \bibinfo{person}{Qiang Cheng}, \bibinfo{person}{Xinwang Liu}, \bibinfo{person}{Xi Peng}, \bibinfo{person}{Zenglin Xu}, {and} \bibinfo{person}{Ling Tian}.} \bibinfo{year}{2021}\natexlab{}.
\newblock \showarticletitle{Structured graph learning for clustering and semi-supervised classification}.
\newblock \bibinfo{journal}{\emph{Pattern Recognition}}  \bibinfo{volume}{110} (\bibinfo{year}{2021}), \bibinfo{pages}{107627}.
\newblock


\bibitem[\protect\citeauthoryear{Kipf and Welling}{Kipf and Welling}{2016a}]%
        {kipf2016semi}
\bibfield{author}{\bibinfo{person}{Thomas~N Kipf} {and} \bibinfo{person}{Max Welling}.} \bibinfo{year}{2016}\natexlab{a}.
\newblock \showarticletitle{Semi-supervised classification with graph convolutional networks}.
\newblock \bibinfo{journal}{\emph{arXiv preprint arXiv:1609.02907}} (\bibinfo{year}{2016}).
\newblock


\bibitem[\protect\citeauthoryear{Kipf and Welling}{Kipf and Welling}{2016b}]%
        {kipf2016variational}
\bibfield{author}{\bibinfo{person}{Thomas~N Kipf} {and} \bibinfo{person}{Max Welling}.} \bibinfo{year}{2016}\natexlab{b}.
\newblock \showarticletitle{Variational graph auto-encoders}.
\newblock \bibinfo{journal}{\emph{arXiv preprint arXiv:1611.07308}} (\bibinfo{year}{2016}).
\newblock


\bibitem[\protect\citeauthoryear{Li and Pan}{Li and Pan}{2016}]%
        {li2016structural}
\bibfield{author}{\bibinfo{person}{Angsheng Li} {and} \bibinfo{person}{Yicheng Pan}.} \bibinfo{year}{2016}\natexlab{}.
\newblock \showarticletitle{Structural information and dynamical complexity of networks}.
\newblock \bibinfo{journal}{\emph{IEEE TIT}} \bibinfo{volume}{62}, \bibinfo{number}{6} (\bibinfo{year}{2016}), \bibinfo{pages}{3290--3339}.
\newblock


\bibitem[\protect\citeauthoryear{Li, Hu, Sun, Hu, Zhang, and Qu}{Li et~al\mbox{.}}{2020}]%
        {li2020deep}
\bibfield{author}{\bibinfo{person}{Xunkai Li}, \bibinfo{person}{Youpeng Hu}, \bibinfo{person}{Yaoqi Sun}, \bibinfo{person}{Ji Hu}, \bibinfo{person}{Jiyong Zhang}, {and} \bibinfo{person}{Meixia Qu}.} \bibinfo{year}{2020}\natexlab{}.
\newblock \showarticletitle{A deep graph structured clustering network}.
\newblock \bibinfo{journal}{\emph{IEEE Access}}  \bibinfo{volume}{8} (\bibinfo{year}{2020}), \bibinfo{pages}{161727--161738}.
\newblock


\bibitem[\protect\citeauthoryear{Lin, Coifman, Mishne, and Talmon}{Lin et~al\mbox{.}}{2023}]%
        {lin2023hyperbolic}
\bibfield{author}{\bibinfo{person}{Ya-Wei~Eileen Lin}, \bibinfo{person}{Ronald~R Coifman}, \bibinfo{person}{Gal Mishne}, {and} \bibinfo{person}{Ronen Talmon}.} \bibinfo{year}{2023}\natexlab{}.
\newblock \showarticletitle{Hyperbolic diffusion embedding and distance for hierarchical representation learning}. In \bibinfo{booktitle}{\emph{International Conference on Machine Learning}}. PMLR, \bibinfo{pages}{21003--21025}.
\newblock


\bibitem[\protect\citeauthoryear{Liu, Gong, Miao, Wang, and Li}{Liu et~al\mbox{.}}{2017}]%
        {liu2017structure}
\bibfield{author}{\bibinfo{person}{Jia Liu}, \bibinfo{person}{Maoguo Gong}, \bibinfo{person}{Qiguang Miao}, \bibinfo{person}{Xiaogang Wang}, {and} \bibinfo{person}{Hao Li}.} \bibinfo{year}{2017}\natexlab{}.
\newblock \showarticletitle{Structure learning for deep neural networks based on multiobjective optimization}.
\newblock \bibinfo{journal}{\emph{IEEE transactions on neural networks and learning systems}} \bibinfo{volume}{29}, \bibinfo{number}{6} (\bibinfo{year}{2017}), \bibinfo{pages}{2450--2463}.
\newblock


\bibitem[\protect\citeauthoryear{Liu, Liu, Zhang, Zhu, and Li}{Liu et~al\mbox{.}}{2019}]%
        {liu2019rem}
\bibfield{author}{\bibinfo{person}{Yiwei Liu}, \bibinfo{person}{Jiamou Liu}, \bibinfo{person}{Zijian Zhang}, \bibinfo{person}{Liehuang Zhu}, {and} \bibinfo{person}{Angsheng Li}.} \bibinfo{year}{2019}\natexlab{}.
\newblock \showarticletitle{REM: From structural entropy to community structure deception}.
\newblock \bibinfo{journal}{\emph{Proceedings of the Advances in Neural Information Processing Systems}}  \bibinfo{volume}{32} (\bibinfo{year}{2019}), \bibinfo{pages}{1--11}.
\newblock


\bibitem[\protect\citeauthoryear{Liu, Zheng, Zhang, Chen, Peng, and Pan}{Liu et~al\mbox{.}}{2022}]%
        {liu2022towards}
\bibfield{author}{\bibinfo{person}{Yixin Liu}, \bibinfo{person}{Yu Zheng}, \bibinfo{person}{Daokun Zhang}, \bibinfo{person}{Hongxu Chen}, \bibinfo{person}{Hao Peng}, {and} \bibinfo{person}{Shirui Pan}.} \bibinfo{year}{2022}\natexlab{}.
\newblock \showarticletitle{Towards unsupervised deep graph structure learning}. In \bibinfo{booktitle}{\emph{Proceedings of the ACM Web Conference 2022}}. \bibinfo{pages}{1392--1403}.
\newblock


\bibitem[\protect\citeauthoryear{Lyu, Zhang, and Zhang}{Lyu et~al\mbox{.}}{2017}]%
        {lyu2017enhancing}
\bibfield{author}{\bibinfo{person}{Tianshu Lyu}, \bibinfo{person}{Yuan Zhang}, {and} \bibinfo{person}{Yan Zhang}.} \bibinfo{year}{2017}\natexlab{}.
\newblock \showarticletitle{Enhancing the network embedding quality with structural similarity}. In \bibinfo{booktitle}{\emph{Proceedings of the 2017 ACM on Conference on Information and Knowledge Management}}. \bibinfo{pages}{147--156}.
\newblock


\bibitem[\protect\citeauthoryear{Mao, Wang, Goodison, and Sun}{Mao et~al\mbox{.}}{2015}]%
        {mao2015dimensionality}
\bibfield{author}{\bibinfo{person}{Qi Mao}, \bibinfo{person}{Li Wang}, \bibinfo{person}{Steve Goodison}, {and} \bibinfo{person}{Yijun Sun}.} \bibinfo{year}{2015}\natexlab{}.
\newblock \showarticletitle{Dimensionality reduction via graph structure learning}. In \bibinfo{booktitle}{\emph{Proceedings of the 21th ACM SIGKDD international conference on knowledge discovery and data mining}}. \bibinfo{pages}{765--774}.
\newblock


\bibitem[\protect\citeauthoryear{Mrabah, Bouguessa, Touati, and Ksantini}{Mrabah et~al\mbox{.}}{2022}]%
        {mrabah2022rethinking}
\bibfield{author}{\bibinfo{person}{Nairouz Mrabah}, \bibinfo{person}{Mohamed Bouguessa}, \bibinfo{person}{Mohamed~Fawzi Touati}, {and} \bibinfo{person}{Riadh Ksantini}.} \bibinfo{year}{2022}\natexlab{}.
\newblock \showarticletitle{Rethinking graph auto-encoder models for attributed graph clustering}.
\newblock \bibinfo{journal}{\emph{IEEE Transactions on Knowledge and Data Engineering}} \bibinfo{volume}{35}, \bibinfo{number}{9} (\bibinfo{year}{2022}), \bibinfo{pages}{9037--9053}.
\newblock


\bibitem[\protect\citeauthoryear{Newman}{Newman}{2006}]%
        {newman2006modularity}
\bibfield{author}{\bibinfo{person}{Mark~EJ Newman}.} \bibinfo{year}{2006}\natexlab{}.
\newblock \showarticletitle{Modularity and community structure in networks}.
\newblock \bibinfo{journal}{\emph{Proceedings of the national academy of sciences}} \bibinfo{volume}{103}, \bibinfo{number}{23} (\bibinfo{year}{2006}), \bibinfo{pages}{8577--8582}.
\newblock


\bibitem[\protect\citeauthoryear{Peng, Zhang, Huang, Hao, Li, Yu, and Yu}{Peng et~al\mbox{.}}{2024}]%
        {peng2024unsupervised}
\bibfield{author}{\bibinfo{person}{Hao Peng}, \bibinfo{person}{Jingyun Zhang}, \bibinfo{person}{Xiang Huang}, \bibinfo{person}{Zhifeng Hao}, \bibinfo{person}{Angsheng Li}, \bibinfo{person}{Zhengtao Yu}, {and} \bibinfo{person}{Philip~S Yu}.} \bibinfo{year}{2024}\natexlab{}.
\newblock \showarticletitle{Unsupervised Social Bot Detection via Structural Information Theory}.
\newblock \bibinfo{journal}{\emph{ACM Transactions on Information Systems}} \bibinfo{volume}{42}, \bibinfo{number}{6} (\bibinfo{year}{2024}), \bibinfo{pages}{42}.
\newblock


\bibitem[\protect\citeauthoryear{Shchur, Mumme, Bojchevski, and G{\"u}nnemann}{Shchur et~al\mbox{.}}{2018}]%
        {shchur2018pitfalls}
\bibfield{author}{\bibinfo{person}{Oleksandr Shchur}, \bibinfo{person}{Maximilian Mumme}, \bibinfo{person}{Aleksandar Bojchevski}, {and} \bibinfo{person}{Stephan G{\"u}nnemann}.} \bibinfo{year}{2018}\natexlab{}.
\newblock \showarticletitle{Pitfalls of graph neural network evaluation}.
\newblock \bibinfo{journal}{\emph{arXiv preprint arXiv:1811.05868}} (\bibinfo{year}{2018}).
\newblock


\bibitem[\protect\citeauthoryear{Tsitsulin, Palowitch, Perozzi, and M{\"u}ller}{Tsitsulin et~al\mbox{.}}{2023}]%
        {tsitsulin2023graph}
\bibfield{author}{\bibinfo{person}{Anton Tsitsulin}, \bibinfo{person}{John Palowitch}, \bibinfo{person}{Bryan Perozzi}, {and} \bibinfo{person}{Emmanuel M{\"u}ller}.} \bibinfo{year}{2023}\natexlab{}.
\newblock \showarticletitle{Graph clustering with graph neural networks}.
\newblock \bibinfo{journal}{\emph{Journal of Machine Learning Research}} \bibinfo{volume}{24}, \bibinfo{number}{127} (\bibinfo{year}{2023}), \bibinfo{pages}{1--21}.
\newblock


\bibitem[\protect\citeauthoryear{Veli{\v{c}}kovi{\'c}, Fedus, Hamilton, Li{\`o}, Bengio, and Hjelm}{Veli{\v{c}}kovi{\'c} et~al\mbox{.}}{2019}]%
        {velivckovic2018deep}
\bibfield{author}{\bibinfo{person}{Petar Veli{\v{c}}kovi{\'c}}, \bibinfo{person}{William Fedus}, \bibinfo{person}{William~L Hamilton}, \bibinfo{person}{Pietro Li{\`o}}, \bibinfo{person}{Yoshua Bengio}, {and} \bibinfo{person}{R~Devon Hjelm}.} \bibinfo{year}{2019}\natexlab{}.
\newblock \showarticletitle{Deep graph infomax}. In \bibinfo{booktitle}{\emph{International Conference on Learning Representations}}.
\newblock


\bibitem[\protect\citeauthoryear{Von~Luxburg}{Von~Luxburg}{2007}]%
        {von2007tutorial}
\bibfield{author}{\bibinfo{person}{Ulrike Von~Luxburg}.} \bibinfo{year}{2007}\natexlab{}.
\newblock \showarticletitle{A tutorial on spectral clustering}.
\newblock \bibinfo{journal}{\emph{Statistics and computing}}  \bibinfo{volume}{17} (\bibinfo{year}{2007}), \bibinfo{pages}{395--416}.
\newblock


\bibitem[\protect\citeauthoryear{Wang, Wang, Zhang, Yang, Zhao, and Liu}{Wang et~al\mbox{.}}{2023}]%
        {wang2023user}
\bibfield{author}{\bibinfo{person}{Yifei Wang}, \bibinfo{person}{Yupan Wang}, \bibinfo{person}{Zeyu Zhang}, \bibinfo{person}{Song Yang}, \bibinfo{person}{Kaiqi Zhao}, {and} \bibinfo{person}{Jiamou Liu}.} \bibinfo{year}{2023}\natexlab{}.
\newblock \showarticletitle{User: Unsupervised structural entropy-based robust graph neural network}.
\newblock \bibinfo{journal}{\emph{Proceedings of the AAAI Conference on Artificial Intelligence}}, \bibinfo{pages}{10235--10243}.
\newblock


\bibitem[\protect\citeauthoryear{Wei, Liang, Liu, and Wang}{Wei et~al\mbox{.}}{2022}]%
        {wei2022contrastive}
\bibfield{author}{\bibinfo{person}{Chunyu Wei}, \bibinfo{person}{Jian Liang}, \bibinfo{person}{Di Liu}, {and} \bibinfo{person}{Fei Wang}.} \bibinfo{year}{2022}\natexlab{}.
\newblock \showarticletitle{Contrastive graph structure learning via information bottleneck for recommendation}.
\newblock \bibinfo{journal}{\emph{Advances in Neural Information Processing Systems}}  \bibinfo{volume}{35} (\bibinfo{year}{2022}), \bibinfo{pages}{20407--20420}.
\newblock


\bibitem[\protect\citeauthoryear{White and Smyth}{White and Smyth}{2005}]%
        {white2005spectral}
\bibfield{author}{\bibinfo{person}{Scott White} {and} \bibinfo{person}{Padhraic Smyth}.} \bibinfo{year}{2005}\natexlab{}.
\newblock \showarticletitle{A spectral clustering approach to finding communities in graphs}. In \bibinfo{booktitle}{\emph{Proceedings of the 2005 SIAM international conference on data mining}}. SIAM, \bibinfo{pages}{274--285}.
\newblock


\bibitem[\protect\citeauthoryear{Wu, Chen, Shi, Li, and Xu}{Wu et~al\mbox{.}}{2023}]%
        {wu2023sega}
\bibfield{author}{\bibinfo{person}{Junran Wu}, \bibinfo{person}{Xueyuan Chen}, \bibinfo{person}{Bowen Shi}, \bibinfo{person}{Shangzhe Li}, {and} \bibinfo{person}{Ke Xu}.} \bibinfo{year}{2023}\natexlab{}.
\newblock \showarticletitle{SEGA: Structural Entropy Guided Anchor View for Graph Contrastive Learning}. In \bibinfo{booktitle}{\emph{Proceedings of the ICML}}. PMLR, \bibinfo{pages}{1--20}.
\newblock


\bibitem[\protect\citeauthoryear{Wu, Chen, Xu, and Li}{Wu et~al\mbox{.}}{2022a}]%
        {wu2022structural}
\bibfield{author}{\bibinfo{person}{Junran Wu}, \bibinfo{person}{Xueyuan Chen}, \bibinfo{person}{Ke Xu}, {and} \bibinfo{person}{Shangzhe Li}.} \bibinfo{year}{2022}\natexlab{a}.
\newblock \showarticletitle{Structural entropy guided graph hierarchical pooling}. In \bibinfo{booktitle}{\emph{Proceedings of the International Conference on Machine Learning}}. PMLR, \bibinfo{pages}{24017--24030}.
\newblock


\bibitem[\protect\citeauthoryear{Wu, Zhao, Li, Wipf, and Yan}{Wu et~al\mbox{.}}{2022c}]%
        {wu2022nodeformer}
\bibfield{author}{\bibinfo{person}{Qitian Wu}, \bibinfo{person}{Wentao Zhao}, \bibinfo{person}{Zenan Li}, \bibinfo{person}{David~P Wipf}, {and} \bibinfo{person}{Junchi Yan}.} \bibinfo{year}{2022}\natexlab{c}.
\newblock \showarticletitle{Nodeformer: A scalable graph structure learning transformer for node classification}.
\newblock \bibinfo{journal}{\emph{Advances in Neural Information Processing Systems}}  \bibinfo{volume}{35} (\bibinfo{year}{2022}), \bibinfo{pages}{27387--27401}.
\newblock


\bibitem[\protect\citeauthoryear{Wu, {\'E}lteto, Dasgupta, and Schulz}{Wu et~al\mbox{.}}{2022b}]%
        {wu2022learning}
\bibfield{author}{\bibinfo{person}{Shuchen Wu}, \bibinfo{person}{No{\'e}mi {\'E}lteto}, \bibinfo{person}{Ishita Dasgupta}, {and} \bibinfo{person}{Eric Schulz}.} \bibinfo{year}{2022}\natexlab{b}.
\newblock \showarticletitle{Learning Structure from the Ground up---Hierarchical Representation Learning by Chunking}.
\newblock \bibinfo{journal}{\emph{Advances in Neural Information Processing Systems}}  \bibinfo{volume}{35} (\bibinfo{year}{2022}), \bibinfo{pages}{36706--36721}.
\newblock


\bibitem[\protect\citeauthoryear{Yang, Tan, Liu, Liang, Wang, Zhou, Xia, Li, Liu, and Zhu}{Yang et~al\mbox{.}}{2023a}]%
        {yang2023convert}
\bibfield{author}{\bibinfo{person}{Xihong Yang}, \bibinfo{person}{Cheng Tan}, \bibinfo{person}{Yue Liu}, \bibinfo{person}{Ke Liang}, \bibinfo{person}{Siwei Wang}, \bibinfo{person}{Sihang Zhou}, \bibinfo{person}{Jun Xia}, \bibinfo{person}{Stan~Z Li}, \bibinfo{person}{Xinwang Liu}, {and} \bibinfo{person}{En Zhu}.} \bibinfo{year}{2023}\natexlab{a}.
\newblock \showarticletitle{Convert: Contrastive graph clustering with reliable augmentation}. In \bibinfo{booktitle}{\emph{Proceedings of the 31st ACM International Conference on Multimedia}}. \bibinfo{pages}{319--327}.
\newblock


\bibitem[\protect\citeauthoryear{Yang, Wu, He, Peng, Yang, Hao, and Liao}{Yang et~al\mbox{.}}{2024}]%
        {yang2024sebot}
\bibfield{author}{\bibinfo{person}{Yingguang Yang}, \bibinfo{person}{Qi Wu}, \bibinfo{person}{Buyun He}, \bibinfo{person}{Hao Peng}, \bibinfo{person}{Renyu Yang}, \bibinfo{person}{Zhifeng Hao}, {and} \bibinfo{person}{Yong Liao}.} \bibinfo{year}{2024}\natexlab{}.
\newblock \showarticletitle{SeBot: Structural Entropy Guided Multi-View Contrastive Learning for Social Bot Detection}. In \bibinfo{booktitle}{\emph{ACM SIGKDD Conference on Knowledge Discovery and Data Mining (KDD)}}.
\newblock


\bibitem[\protect\citeauthoryear{Yang, Cohen, and Salakhudinov}{Yang et~al\mbox{.}}{2016}]%
        {yang2016revisiting}
\bibfield{author}{\bibinfo{person}{Zhilin Yang}, \bibinfo{person}{William Cohen}, {and} \bibinfo{person}{Ruslan Salakhudinov}.} \bibinfo{year}{2016}\natexlab{}.
\newblock \showarticletitle{Revisiting semi-supervised learning with graph embeddings}. In \bibinfo{booktitle}{\emph{International conference on machine learning}}. PMLR, \bibinfo{pages}{40--48}.
\newblock


\bibitem[\protect\citeauthoryear{Yang, Zhang, Wu, Yang, Sheng, Peng, Li, Xue, and Su}{Yang et~al\mbox{.}}{2023b}]%
        {yang2023minimum}
\bibfield{author}{\bibinfo{person}{Zhenyu Yang}, \bibinfo{person}{Ge Zhang}, \bibinfo{person}{Jia Wu}, \bibinfo{person}{Jian Yang}, \bibinfo{person}{Quan~Z Sheng}, \bibinfo{person}{Hao Peng}, \bibinfo{person}{Angsheng Li}, \bibinfo{person}{Shan Xue}, {and} \bibinfo{person}{Jianlin Su}.} \bibinfo{year}{2023}\natexlab{b}.
\newblock \showarticletitle{Minimum entropy principle guided graph neural networks}. In \bibinfo{booktitle}{\emph{Proceedings of the Sixteenth ACM International Conference on Web Search and Data Mining}}. \bibinfo{pages}{114--122}.
\newblock


\bibitem[\protect\citeauthoryear{Ying, You, Morris, Ren, Hamilton, and Leskovec}{Ying et~al\mbox{.}}{2018}]%
        {ying2018hierarchical}
\bibfield{author}{\bibinfo{person}{Zhitao Ying}, \bibinfo{person}{Jiaxuan You}, \bibinfo{person}{Christopher Morris}, \bibinfo{person}{Xiang Ren}, \bibinfo{person}{Will Hamilton}, {and} \bibinfo{person}{Jure Leskovec}.} \bibinfo{year}{2018}\natexlab{}.
\newblock \showarticletitle{Hierarchical graph representation learning with differentiable pooling}.
\newblock \bibinfo{journal}{\emph{Advances in neural information processing systems}}  \bibinfo{volume}{31} (\bibinfo{year}{2018}).
\newblock


\bibitem[\protect\citeauthoryear{Yu, Zhang, Jiang, Wu, and Yang}{Yu et~al\mbox{.}}{2021}]%
        {yu2021graph}
\bibfield{author}{\bibinfo{person}{Donghan Yu}, \bibinfo{person}{Ruohong Zhang}, \bibinfo{person}{Zhengbao Jiang}, \bibinfo{person}{Yuexin Wu}, {and} \bibinfo{person}{Yiming Yang}.} \bibinfo{year}{2021}\natexlab{}.
\newblock \showarticletitle{Graph-revised convolutional network}. In \bibinfo{booktitle}{\emph{Machine Learning and Knowledge Discovery in Databases: European Conference, ECML PKDD 2020, Ghent, Belgium, September 14--18, 2020, Proceedings, Part III}}. Springer, \bibinfo{pages}{378--393}.
\newblock


\bibitem[\protect\citeauthoryear{Zeng, Peng, Li, Liu, Liu, Philip, and He}{Zeng et~al\mbox{.}}{2023b}]%
        {zeng2023unsupervised}
\bibfield{author}{\bibinfo{person}{Guangjie Zeng}, \bibinfo{person}{Hao Peng}, \bibinfo{person}{Angsheng Li}, \bibinfo{person}{Zhiwei Liu}, \bibinfo{person}{Chunyang Liu}, \bibinfo{person}{S~Yu Philip}, {and} \bibinfo{person}{Lifang He}.} \bibinfo{year}{2023}\natexlab{b}.
\newblock \showarticletitle{Unsupervised Skin Lesion Segmentation via Structural Entropy Minimization on Multi-Scale Superpixel Graphs}. In \bibinfo{booktitle}{\emph{Proceedings of the IEEE International Conference on Data Mining (ICDM)}}. \bibinfo{pages}{768--777}.
\newblock


\bibitem[\protect\citeauthoryear{Zeng, Peng, Li, Liu, He, and Yu}{Zeng et~al\mbox{.}}{2023a}]%
        {zeng2023hierarchical}
\bibfield{author}{\bibinfo{person}{Xianghua Zeng}, \bibinfo{person}{Hao Peng}, \bibinfo{person}{Angsheng Li}, \bibinfo{person}{Chunyang Liu}, \bibinfo{person}{Lifang He}, {and} \bibinfo{person}{Philip~S. Yu}.} \bibinfo{year}{2023}\natexlab{a}.
\newblock \showarticletitle{Hierarchical State Abstraction based on Structural Information Principles}. In \bibinfo{booktitle}{\emph{Proceedings of the Thirty-Second International Joint Conference on Artificial Intelligence, {IJCAI-23}}}, \bibfield{editor}{\bibinfo{person}{Edith Elkind}} (Ed.). \bibinfo{publisher}{International Joint Conferences on Artificial Intelligence Organization}, \bibinfo{pages}{4549--4557}.
\newblock


\bibitem[\protect\citeauthoryear{Zhang, Lei, Xiao, and Chen}{Zhang et~al\mbox{.}}{2022a}]%
        {zhang2022cross}
\bibfield{author}{\bibinfo{person}{Chengde Zhang}, \bibinfo{person}{Yu Lei}, \bibinfo{person}{Xia Xiao}, {and} \bibinfo{person}{Xinzhong Chen}.} \bibinfo{year}{2022}\natexlab{a}.
\newblock \showarticletitle{Cross-media video event mining based on attention graph structure learning}.
\newblock \bibinfo{journal}{\emph{Neurocomputing}}  \bibinfo{volume}{502} (\bibinfo{year}{2022}), \bibinfo{pages}{148--158}.
\newblock


\bibitem[\protect\citeauthoryear{Zhang, Li, Zhang, and Li}{Zhang et~al\mbox{.}}{2022b}]%
        {zhang2022embedding}
\bibfield{author}{\bibinfo{person}{Hongyuan Zhang}, \bibinfo{person}{Pei Li}, \bibinfo{person}{Rui Zhang}, {and} \bibinfo{person}{Xuelong Li}.} \bibinfo{year}{2022}\natexlab{b}.
\newblock \showarticletitle{Embedding graph auto-encoder for graph clustering}.
\newblock \bibinfo{journal}{\emph{IEEE Transactions on Neural Networks and Learning Systems}} \bibinfo{volume}{34}, \bibinfo{number}{11} (\bibinfo{year}{2022}), \bibinfo{pages}{9352--9362}.
\newblock


\bibitem[\protect\citeauthoryear{Zhang, Xiong, Zhang, Zhang, Jiao, and Zhu}{Zhang et~al\mbox{.}}{2020}]%
        {zhang2020commdgi}
\bibfield{author}{\bibinfo{person}{Tianqi Zhang}, \bibinfo{person}{Yun Xiong}, \bibinfo{person}{Jiawei Zhang}, \bibinfo{person}{Yao Zhang}, \bibinfo{person}{Yizhu Jiao}, {and} \bibinfo{person}{Yangyong Zhu}.} \bibinfo{year}{2020}\natexlab{}.
\newblock \showarticletitle{CommDGI: community detection oriented deep graph infomax}. In \bibinfo{booktitle}{\emph{Proceedings of the 29th ACM international conference on information \& knowledge management}}. \bibinfo{pages}{1843--1852}.
\newblock


\bibitem[\protect\citeauthoryear{Zhou, Zhou, Duan, Huang, Tan, and Yu}{Zhou et~al\mbox{.}}{2023}]%
        {zhou2023interest}
\bibfield{author}{\bibinfo{person}{Huachi Zhou}, \bibinfo{person}{Shuang Zhou}, \bibinfo{person}{Keyu Duan}, \bibinfo{person}{Xiao Huang}, \bibinfo{person}{Qiaoyu Tan}, {and} \bibinfo{person}{Zailiang Yu}.} \bibinfo{year}{2023}\natexlab{}.
\newblock \showarticletitle{Interest driven graph structure learning for session-based recommendation}. In \bibinfo{booktitle}{\emph{Pacific-Asia Conference on Knowledge Discovery and Data Mining}}. Springer, \bibinfo{pages}{284--296}.
\newblock


\bibitem[\protect\citeauthoryear{Zou, Peng, Huang, Yang, Li, Wu, Liu, and Yu}{Zou et~al\mbox{.}}{2023}]%
        {zou2023se}
\bibfield{author}{\bibinfo{person}{Dongcheng Zou}, \bibinfo{person}{Hao Peng}, \bibinfo{person}{Xiang Huang}, \bibinfo{person}{Renyu Yang}, \bibinfo{person}{Jianxin Li}, \bibinfo{person}{Jia Wu}, \bibinfo{person}{Chunyang Liu}, {and} \bibinfo{person}{Philip~S Yu}.} \bibinfo{year}{2023}\natexlab{}.
\newblock \showarticletitle{SE-GSL: A General and Effective Graph Structure Learning Framework through Structural Entropy Optimization}. In \bibinfo{booktitle}{\emph{Proceedings of the ACM Web Conference 2023}}. \bibinfo{pages}{499–510}.
\newblock


\bibitem[\protect\citeauthoryear{Zou, Wang, Li, Peng, Wang, Liu, Sheng, and Zhang}{Zou et~al\mbox{.}}{2024}]%
        {zou2024multispans}
\bibfield{author}{\bibinfo{person}{Dongcheng Zou}, \bibinfo{person}{Senzhang Wang}, \bibinfo{person}{Xuefeng Li}, \bibinfo{person}{Hao Peng}, \bibinfo{person}{Yuandong Wang}, \bibinfo{person}{Chunyang Liu}, \bibinfo{person}{Kehua Sheng}, {and} \bibinfo{person}{Bo Zhang}.} \bibinfo{year}{2024}\natexlab{}.
\newblock \showarticletitle{Multispans: A multi-range spatial-temporal transformer network for traffic forecast via structural entropy optimization}. In \bibinfo{booktitle}{\emph{Proceedings of the 17th ACM International Conference on Web Search and Data Mining}}. \bibinfo{pages}{1--10}.
\newblock


\end{thebibliography}

\clearpage
\appendix

\section{Notations} 
\label{ap: notations}
The comprehensive list of the primary symbols used throughout this paper is presented in Table~\ref{tab: notation}.
\begin{table}[h]
    \aboverulesep=0ex
    \belowrulesep=0ex
    \vspace{-2mm}
    \caption{Forms and interpretations of notations.} 
    \vspace{-4mm}
    \centering
    \renewcommand\arraystretch{1.1}
    \setlength{\tabcolsep}{1mm}
    \begin{adjustbox}{width=1.0\columnwidth,center}
    \begin{tabular}{r|l} 
        \toprule
        \textbf{Symbol} & \textbf{Definition} \\
        \hline
        
        $G=(V, \mathcal{E}, X)$ & The original graph.\\
        
        $A_g \in {\{0, 1\}}^{N \times N}$ & The adjacency matrix from the original graph.\\

        $A_f \in {\{0, 1\}}^{N \times N}$ & The adjacency matrix from the learned graph.\\
        
        $V$, $\mathcal{E}$ & The node/edge set.\\

        $X \in R^{N \times f}$ & The node feature matrix.\\

        $S \in {\{0, 1\}}^{N \times c}$ & The node assignment matrix. \\

        $S^k \in R^{N_k \times N_{k-1}}$ & The assignment matrix between layer $k$ and layer $k-1$.\\

        $C^k \in R^{N \times N_k}$ & The direct assignment matrix between nodes and layer $k$.\\
        
        $N$, $N_k$ & The number of nodes, The number of vertices of layer $k$.\\
         
        $M$, $c$ & The number of edges/clusters. \\

        $f$, $d$ & The dimension of feature/embedding. \\
        
        \hline
        
        $\mathcal{T}$; $\lambda$ & Encoding tree; The root vertice of the encoding tree. \\

        $\alpha$; $\alpha^-$ & Vertice on encoding tree. Parent vertice of vertice $\alpha$. \\

        $T_\alpha$ & A node subset corresponds to vertice $\alpha$. \\

        $g_\alpha$ & Number of cutting edges of nodes in vertice $\alpha$. \\

        $h$ & Height of the encoding tree.\\

        $vol(G); vol(\alpha)$ & Volume of Graph $G$; Volume of vertice $\alpha$.\\

        ${vol}^k[i]$ & Volume of vertex $i$ at layer $k$ with soft assignment.\\

        ${vol}^k_{in}[i]$ & Internal volume of vertex $i$ at layer $k$.\\

        $g^k_i$ & Cutting edges of vertex $i$ at layer $k$ with soft assignment.\\

        $H^\mathcal{T}(G)$ & The structural entropy.\\

        $H^\mathcal{T}_{sa}(G)$ & The soft assignment SE of graph $G$.\\

        $H_k(G)$ & The $k$-dimensional structure entropy.\\
        
        $H_{sa}(G;k)$ & The soft assignment structural entropy at layer k.\\
        
        $D \in R^N$ & The degree vector for all nodes. \\

        $W \in R^{N \times N}$ & The weight matrix. \\

        \hline

        $MLP(;\Theta_f)$ & The multilayer perceptron with parameter $\Theta_f$.\\
         
        $KNN(;K)$ & The k-nearest nerighbors algorithm with parameter $K$.\\
        
        ${GNN}_{emb}(;\Theta_1)$ & The embedding learner with parameter $\Theta_1$.\\
         
        ${GNN}_{ass}(;\Theta_2)$ & The soft assignment learner with parameter $\Theta_2$.\\
         
        $\Gamma$ & The attention matrix.\\

        $\mathcal{L}_{ce}$, $\mathcal{L}_{se}$ & The CE loss. The SE loss.\\

        $\lambda_{ce}$, $\lambda_{se}$ & The coefficients of CE loss and SE loss.\\

        $\beta_f$, $K$ & The weight of $A_f$. The number of neighbors in KNN.\\
        
        \bottomrule
    \end{tabular}
    \end{adjustbox}
    \label{tab: notation}
\end{table}

\vspace{-4mm}
\section{Algorithm} 
\label{ap: algo}
The algorithm of ~\model{} is summarized in Algorithm ~\ref{algo: DeSE}.

\vspace{-4mm}
\section{Baselines} 
\label{ap: baseline}
Detailed descriptions of 8 baselines compared to our work are introduced as follows:

$\bullet$
\textbf{DMoN}~\cite{tsitsulin2023graph} introduces a modularity measure of clustering quality to optimize cluster assignment in an end-to-end manner and proposes Deep Modularity Networks.

$\bullet$
\textbf{MinCut}~\cite{bianchi2020spectral} formulates a continuous relaxation of the normalized minCUT problem and trains a GNN to compute cluster assignments by optimizing this objective.

$\bullet$
\textbf{DGI}~\cite{velivckovic2018deep} is a versatile method for learning node representations in graph-structured data based on maximizing the mutual information between local patch representations and high-level graph summaries.

$\bullet$
\textbf{SUBLIME}~\cite{liu2022towards} is a structure bootstrapping contrastive Learning framework with the aid of self-supervised contrastive learning, where the learned graph topology is optimized by data itself.

$\bullet$
\textbf{EGAE}~\cite{zhang2022embedding} is a specific GAE-based model for graph clustering that is consistent with the theory of learning well-explainable representations.

$\bullet$
\textbf{CONVERT}~\cite{yang2023convert} is a contrastive graph clustering network with reliable augmentation and distills reliable semantic information by recovering the perturbed latent embeddings.

$\bullet$
\textbf{AGC-DRR}~\cite{gong2022attributed} is an attributed graph clustering framework with dual redundancy reduction to reduce the information redundancy in both input and latent feature space.

$\bullet$
\textbf{RDGAE}~\cite{mrabah2022rethinking} is a tailored GAE model that triggers a correction mechanism against Feature Drift by gradually transforming the reconstructed graph into a clustering-oriented one.
\vspace{-5mm}
\begin{algorithm}[h]
\LinesNumbered
    \KwIn{Original adjacency matrix: $A_g$; Feature matrix: $X$; Coefficient of CE and SE loss: $\lambda_{ce}$, $\lambda_{se}$; Weight of $A_f$: $\beta_f$; Number of neighbors: $K$; Number of clusters: $c$; Dimension of embedding: $d$; Number of layers: $h$.}

    \KwOut{Node assignment matrix: $S \in {\{0, 1\}}^{N \times c}$.}
    
    \tcp{SLL}
    Initialize $A_f$ with structure learning layer via Eq.~\ref{Eq: A_f_1};\\
    Update $A_f$ via Eq.~\ref{Eq: A_f_2};\\
    Calculate $W_0$ with $\beta_f$, $A_g$, and $A_f$ via Eq.~\ref{Eq: W};\\
    Initialize embedding $E$ with GNN and $W$;\\
    Initialize list $S_{list} \leftarrow \{\}$;\\

    \tcp{ASS}
    \For{k=1,2,..,h}{
        Calculate embedding $H$ with Embedding Learner ${GNN}_{emb}$ via Eq.~\ref{Eq: GNN_emb};\\
        Calculate matrix $S^k$ with Soft AssignmentLearner ${GNN}_{ass}$ via Eq.~\ref{Eq: GNN_ass};\\
        Update $A_g$ and $A_f$ via Eq.~\ref{Eq: update A};\\
        Calculate $W_k$ with $\beta_f$, $A_g$, and $A_f$ via Eq.~\ref{Eq: W};\\
        Store $S^k$ to $S_{list}$;\\
    }

    \tcp{Soft Assignment SE}
    Initialize $\mathcal{L}_{se} \leftarrow 0$;\\
    \For{k=1,2,..,h}{
        Calculate $C^k$ with $\{S^h, ..., S^{k+1}\}$ via Eq.~\ref{Eq: C};\\
        Calculate ${vol}^k$ via Eq.~\ref{Eq: vol};\\
        Calculate $g^k$ via Eq.~\ref{Eq: g};\\
        Calculate $H_{sa}(G;k)$ with $S^k$, $C^k$, ${vol}^k$, and $g^k$ via Eq.~\ref{Eq: H_soft_k};\\
        Update $\mathcal{L}_{se}  \leftarrow \mathcal{L}_{se}+H_{sa}(G;k)$;\\
    }
    
    Calculate $\mathcal{L}_{ce}$ with embedding $E$ via Eq.~\ref{Eq: loss_ce};\\
    Calculate final loss $\mathcal{L}$ with coefficients $\lambda_{ce}, \lambda_{se}$ via Eq.~\ref{Eq: loss};\\

    \tcp{hard assignment matrix}
    Initialize $S \leftarrow 0^{N \times c}$;\\
    Calculate the index $idx$ of the maximum value of $C^k$ along each row;\\
    \For{i=1,2,...,N}{
        Set $S[i,idx[i]]=1$;\\
    }

    Return the hard assignment matrix $S$.
    
    \caption{Algorithm of one epoch of ~\model{}.}
    \label{algo: DeSE}
\end{algorithm}

\section{Dataset}
\label{ap: dataset}
Table~\ref{tab: dataset} provides computed values like the average number of edges per node (Sparsity) and the number of isolated nodes (Iso.).
Detailed descriptions of four datasets are provided below:
$\bullet$
\textbf{Cora}~\cite{yang2016revisiting} is a citation network composed of research papers in the field of machine learning, with each paper linked to others that it cites. The nodes represent the papers, while the edges denote citation relationships between them. 
$\bullet$
\textbf{Citeseer}~\cite{yang2016revisiting} is a citation network similar to Cora, consisting of scientific papers. Each node represents a research paper, and the edges represent citation links between them. 
$\bullet$
\textbf{Computer}~\cite{shchur2018pitfalls} is part of the Amazon co-purchase graph, where nodes represent products in the "computers" category on Amazon, and edges indicate that two products are frequently bought together.
$\bullet$
\textbf{Photo}~\cite{shchur2018pitfalls} is a part of the Amazon co-purchase graph, but it focuses on products in the "photo" category (e.g., cameras, photography accessories). Similar to the Computer dataset, nodes represent products, and edges indicate frequent co-purchases.

\vspace{-3mm}
\section{Graph Clustering Performance} 
\label{ap: graph clustering performance}
\begin{figure}[t]
	\centering
	\subfigure[Cora.]{
		\includegraphics[width=0.48\linewidth]{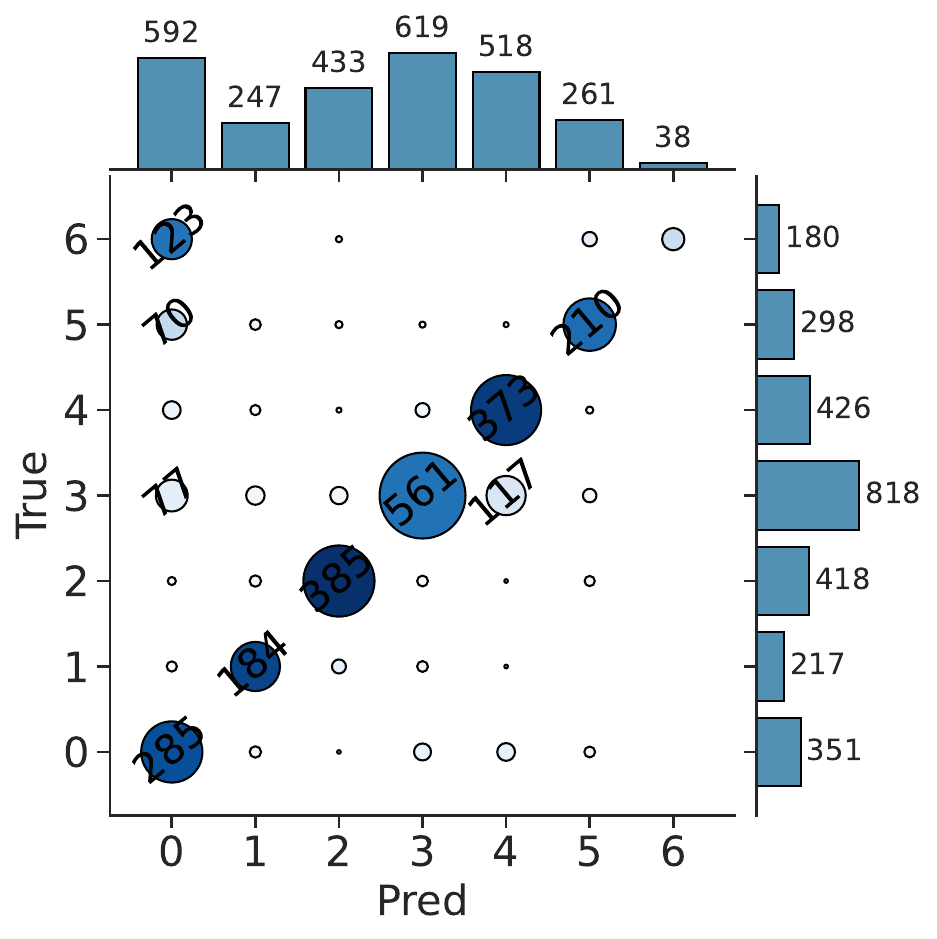}
		\label{fig: cluster_Cora}
	}\hfill
	\subfigure[Citeseer.]{
		\includegraphics[width=0.48\linewidth]{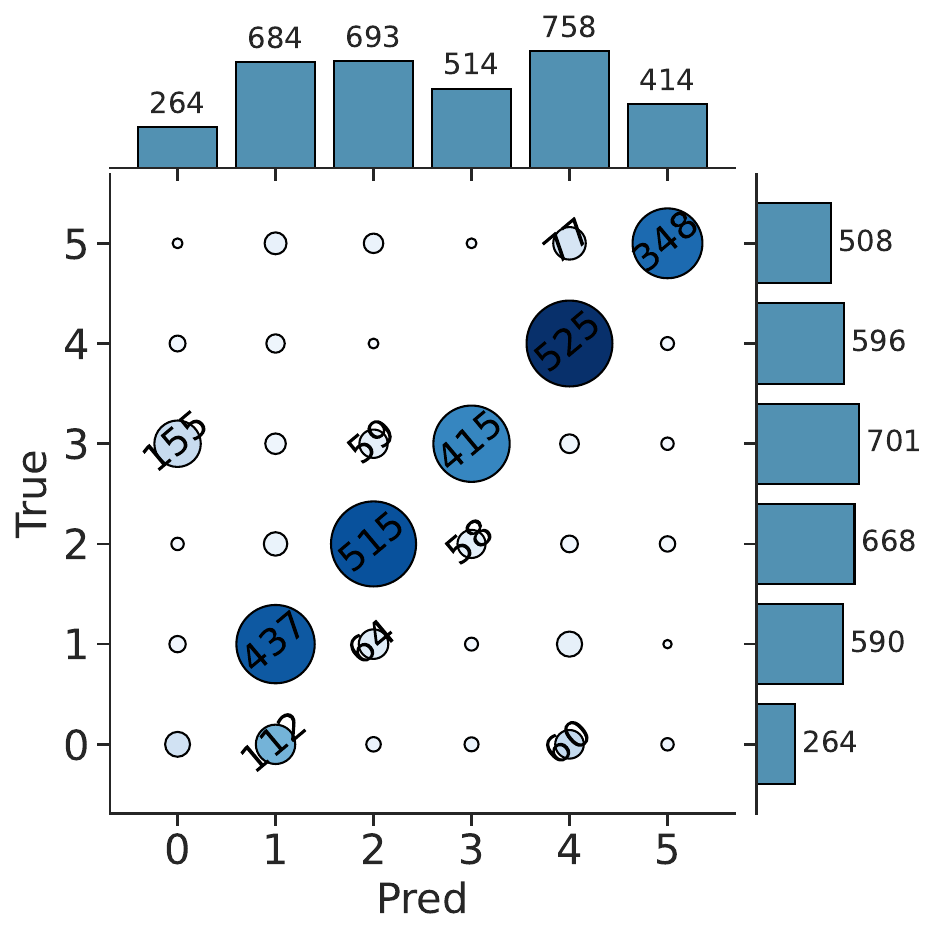}
		\label{fig: cluster_Citeseer}
	}\hfill
    
        \vspace{-4mm}
        \subfigure[Computer.]{
		\includegraphics[width=0.5\linewidth]{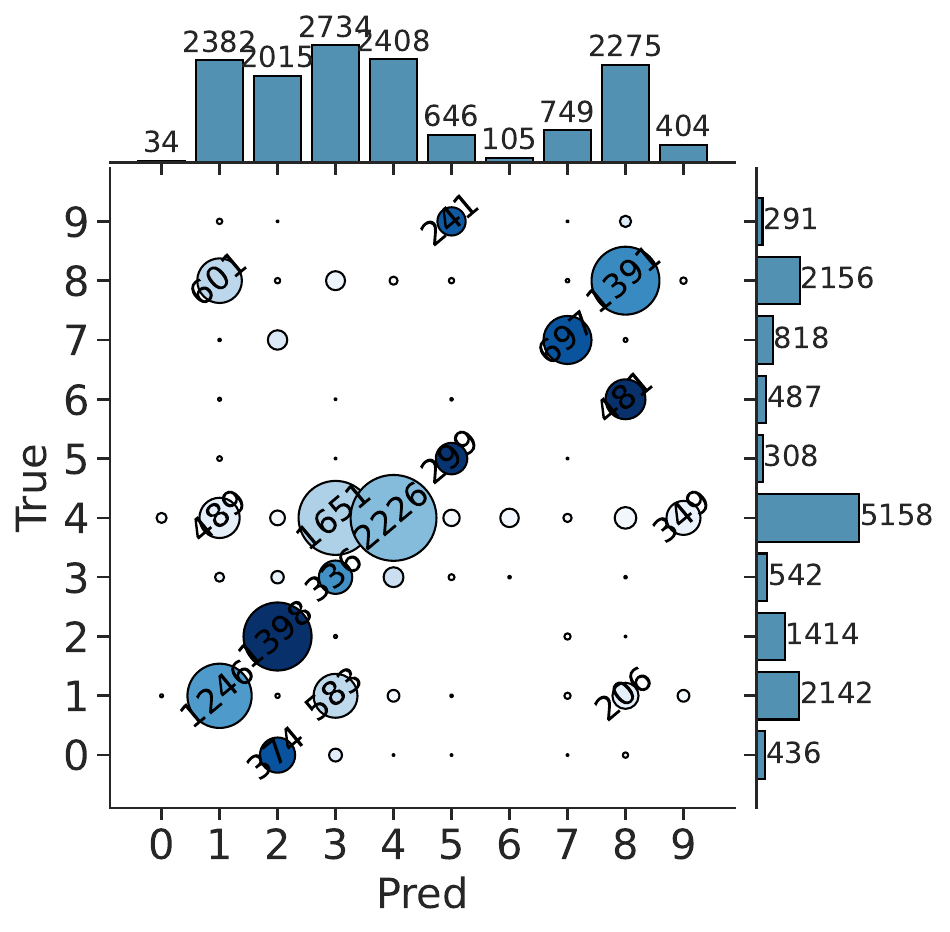}
		\label{fig: cluster_Computer}
	}
        \vspace{-6mm}
	\caption{Clusters of ~\model{} on Cora, Citeseer, and Computer datasets.}
	\label{fig: cluster_DeSE_all}
    \vspace{-6mm}
\end{figure}
The detailed correspondence between the nodes of the predicted clusters and the true clusters of~\model{} on Cora, Citeseer, and Computer datasets are presented in Figure~\ref{fig: cluster_DeSE_all}. 
We can observe that ~\model{} demonstrates concentrated and accurate predictions for larger clusters across the three datasets, while smaller clusters are often disregarded.
For example, in Cora, the predicted cluster 0 primarily contains most of the actual clusters 0 and 6, but since cluster 0 is larger in size, the predicted cluster is classified as cluster 0. 
Similarly, in Citeseer, the predicted cluster 0 mostly contains actual cluster 3, but because the predicted cluster 3 includes more of the actual cluster 3, it is not assigned to cluster 3. 
In contrast, there are more misclassifications in the Computer dataset, particularly with the actual cluster 4 being more widely dispersed in the predictions. 
We believe that the primary reason for these errors is the unclear boundaries between clusters. 
As seen in Figure~\ref{fig: cluster_DeSE_all}, the errors tend to appear collectively, indicating that smaller clusters are easily merged into a larger cluster or that a large cluster is split into smaller ones. 
Improving or fine-tuning cluster boundaries within ~\model{} is the next step for future research.\par

\vspace{-2mm}
\section{Robustness on Clutsers} 
\label{ap: robustness on clusters}
\vspace{-4mm}
\begin{figure}[h]
	\centering
	\begin{minipage}[b]{0.3\linewidth}
		\centering
		\includegraphics[width=\linewidth]{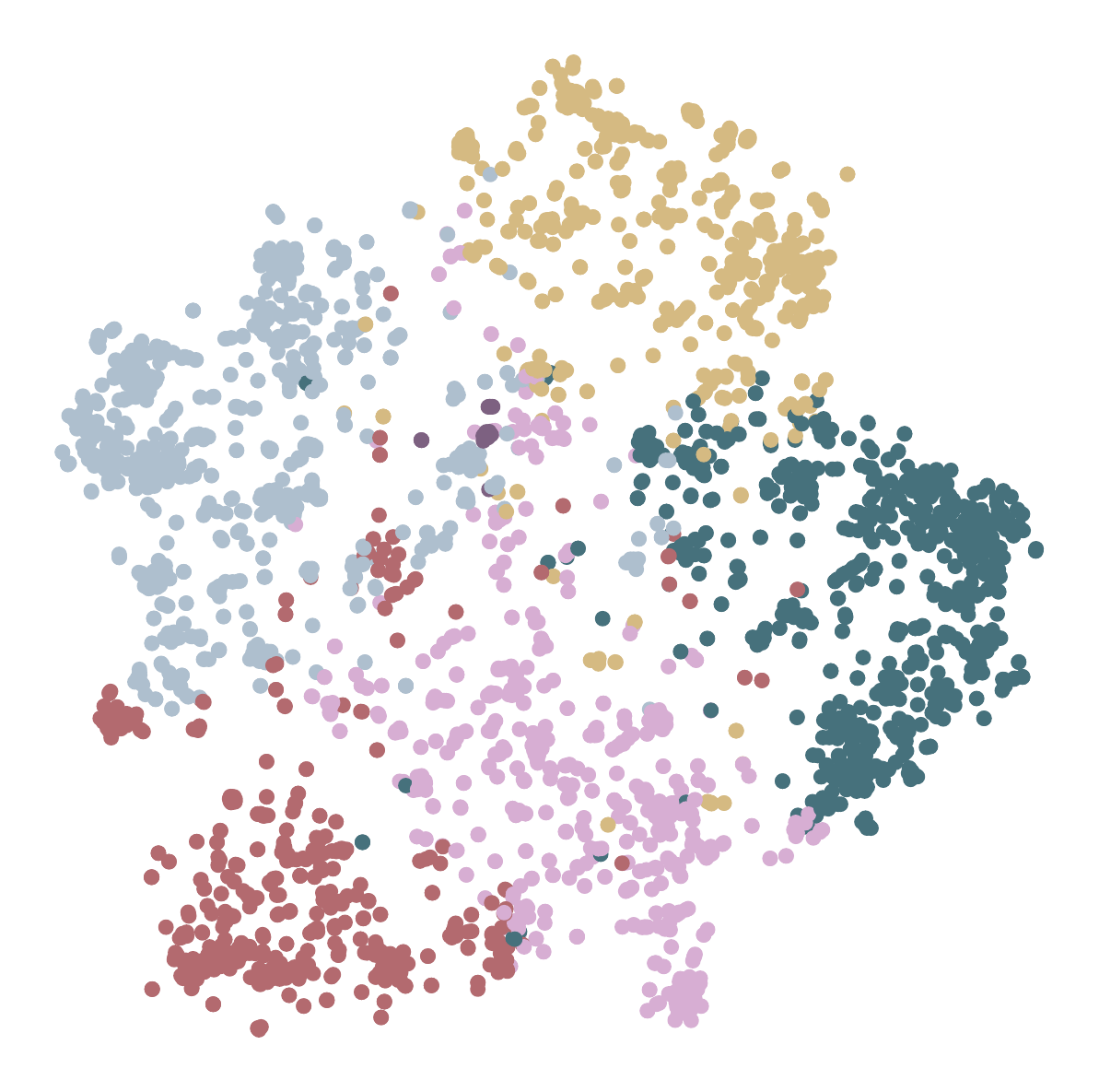}
        \vspace{-8mm}
		\caption*{\footnotesize\mdseries (a) cluster6 (NMI=42.16).}
	\end{minipage}\hfill
	\begin{minipage}[b]{0.3\linewidth}
		\centering
		\includegraphics[width=\linewidth]{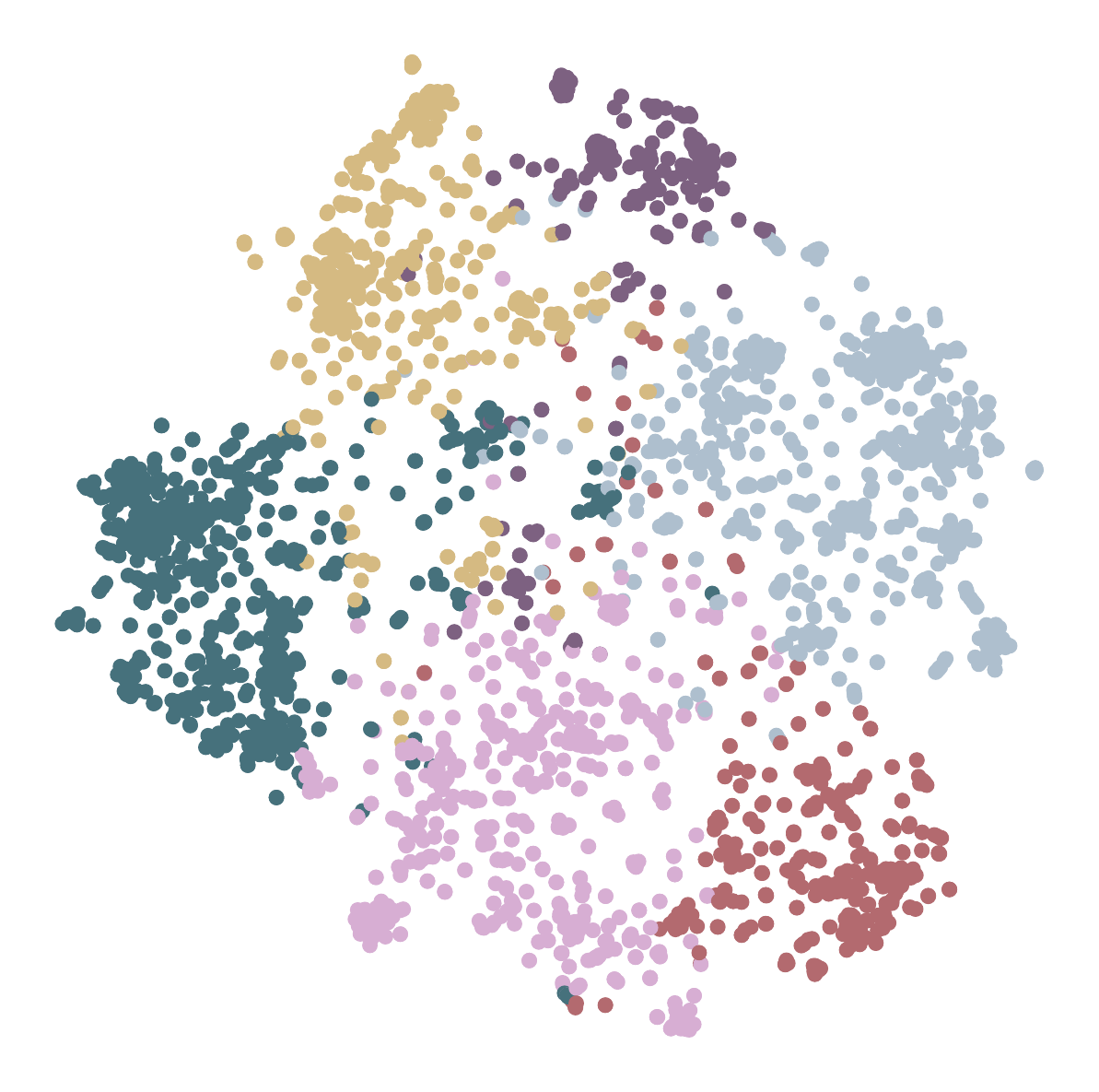}
        \vspace{-8mm}
		\caption*{\footnotesize\mdseries (c) cluster7 (NMI=44.34).}
	\end{minipage}\hfill
	\begin{minipage}[b]{0.3\linewidth}
		\centering
		\includegraphics[width=\linewidth]{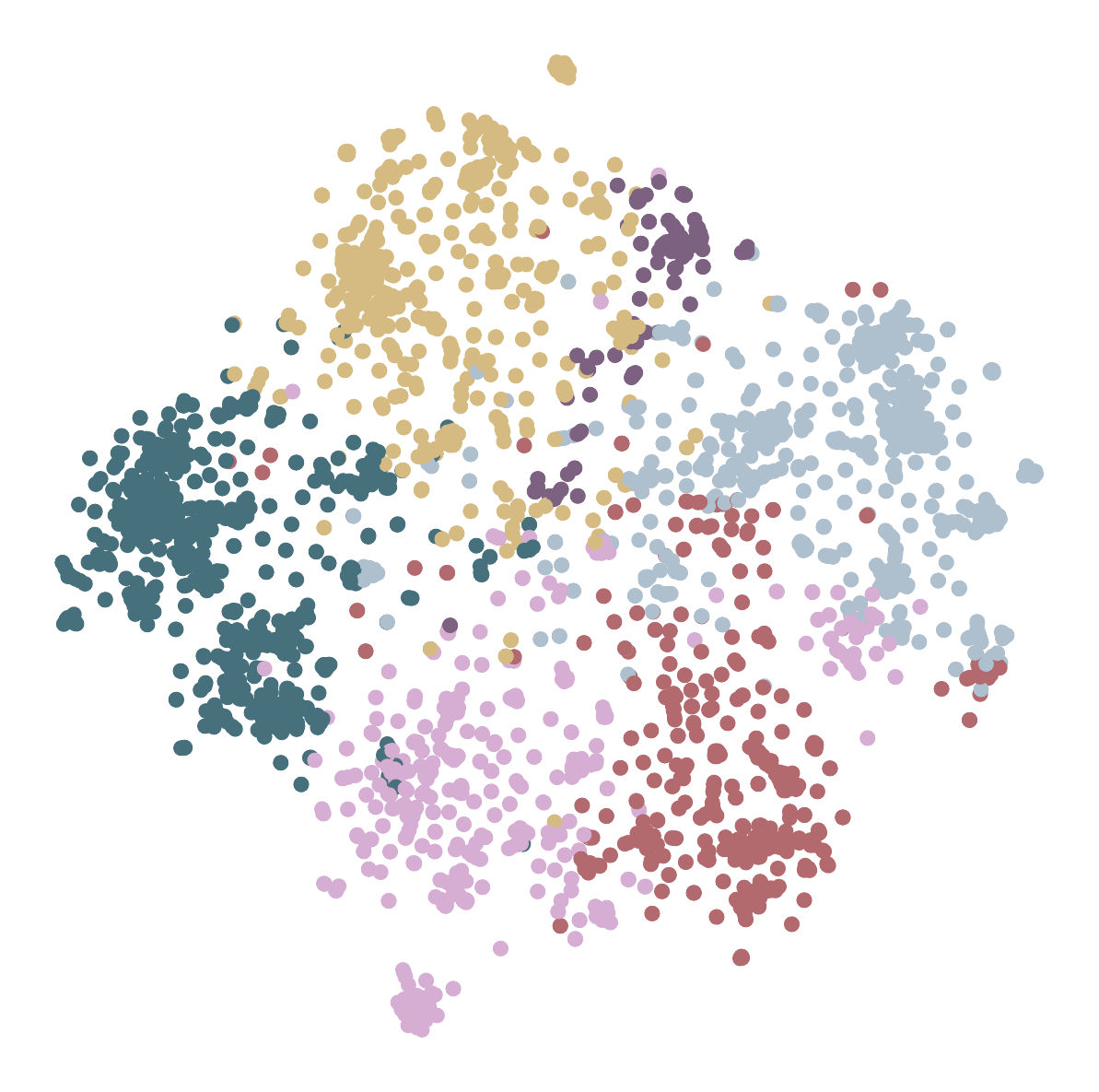}
        \vspace{-8mm}
		\caption*{\footnotesize\mdseries (e) cluster8 (NMI=41.98).}
	\end{minipage}\hfill
	\begin{minipage}[b]{0.3\linewidth}
		\centering
		\includegraphics[width=\linewidth]{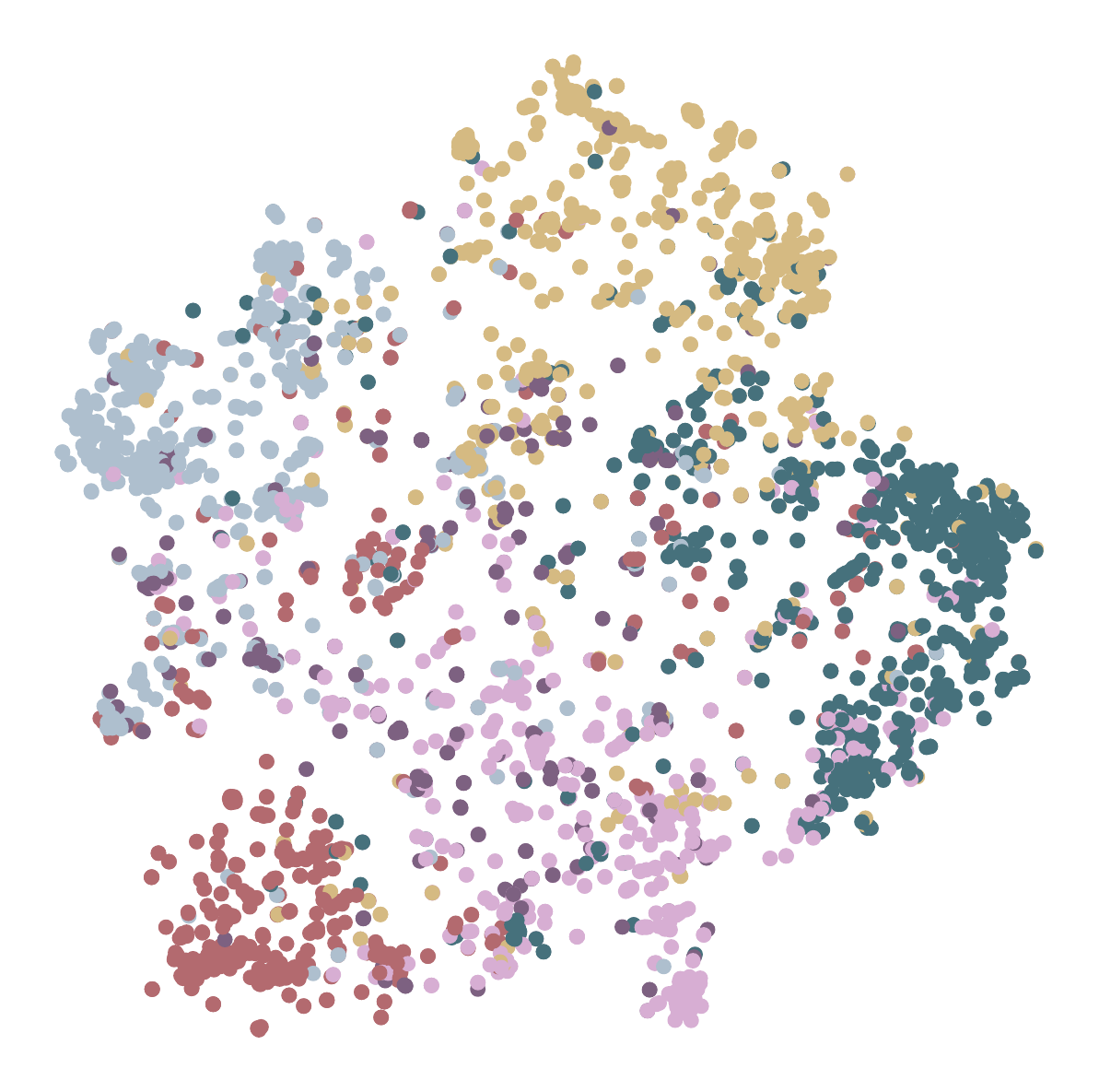}
        \vspace{-8mm}		
        \caption*{\footnotesize\mdseries (b) cluster6 true.}
	\end{minipage}\hfill
	\begin{minipage}[b]{0.3\linewidth}
		\centering
		\includegraphics[width=\linewidth]{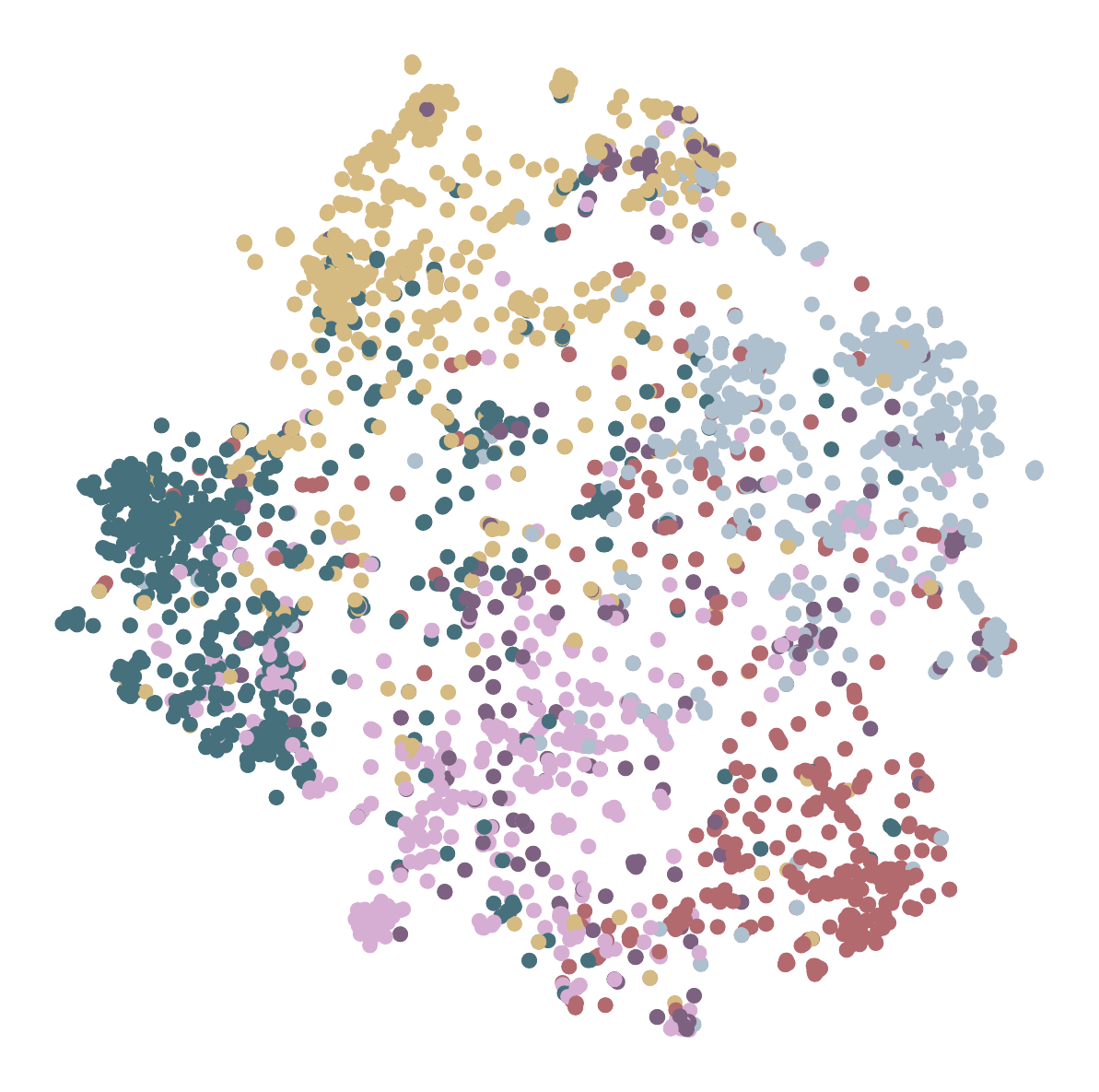}
        \vspace{-8mm}
		\caption*{\footnotesize\mdseries (d) cluster7 true.}
	\end{minipage}\hfill
	\begin{minipage}[b]{0.3\linewidth}
		\centering
		\includegraphics[width=\linewidth]{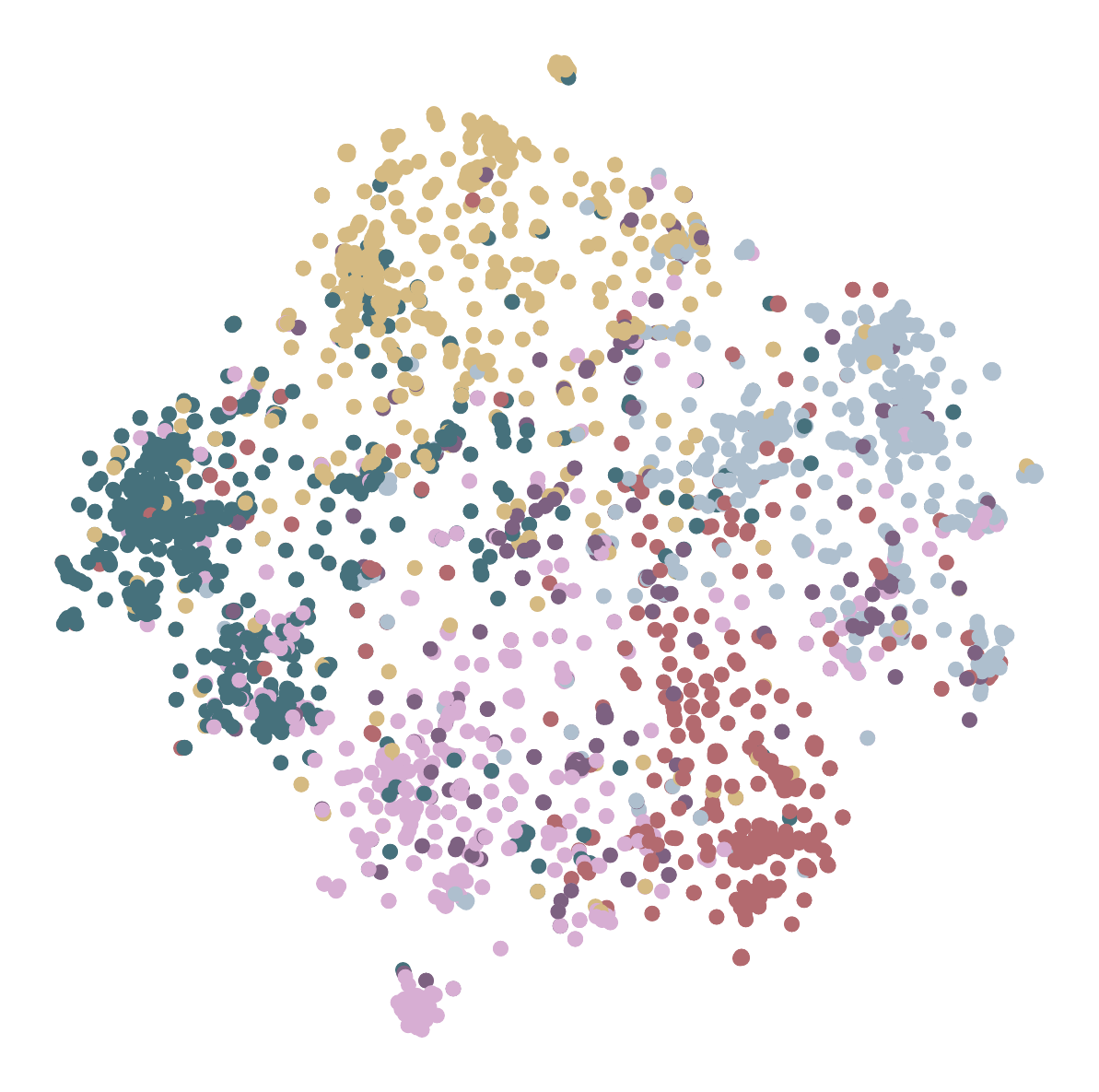}
        \vspace{-8mm}
		\caption*{\footnotesize\mdseries (f) cluster8 true.}
	\end{minipage}\hfill
	\vspace{-4mm}
	\caption{Robustness of cluster numbers on Citeseer.}
	\label{fig: num_cluster_citeseer}
        \vspace{-8mm}
\end{figure}
\begin{figure}[h]
	\centering
	\begin{minipage}[b]{0.3\linewidth}
		\centering
		\includegraphics[width=\linewidth]{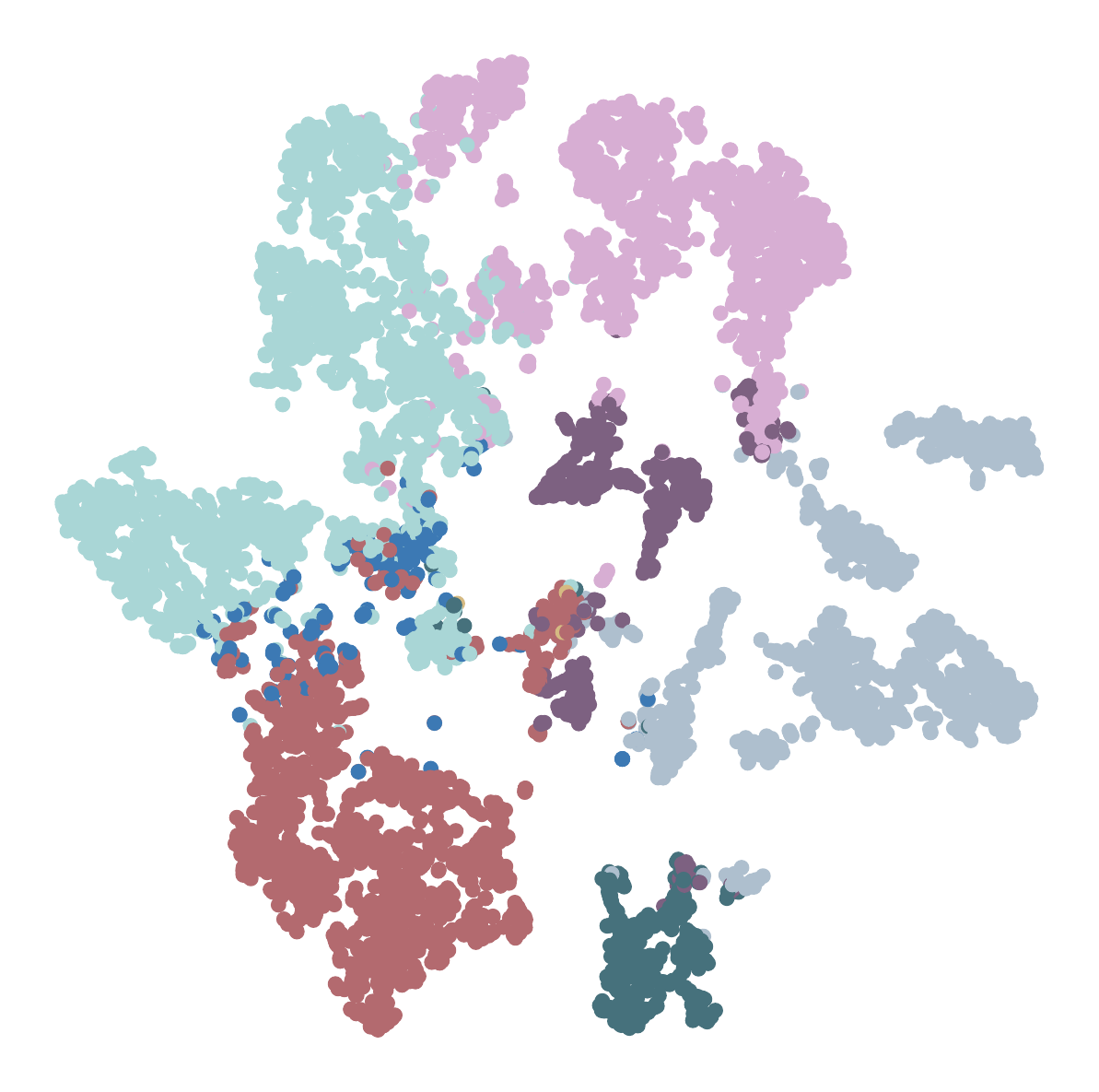}
        \vspace{-8mm}
		\caption*{\footnotesize\mdseries (a) cluster8 (NMI=63.04).}
	\end{minipage}\hfill
	\begin{minipage}[b]{0.3\linewidth}
		\centering
		\includegraphics[width=\linewidth]{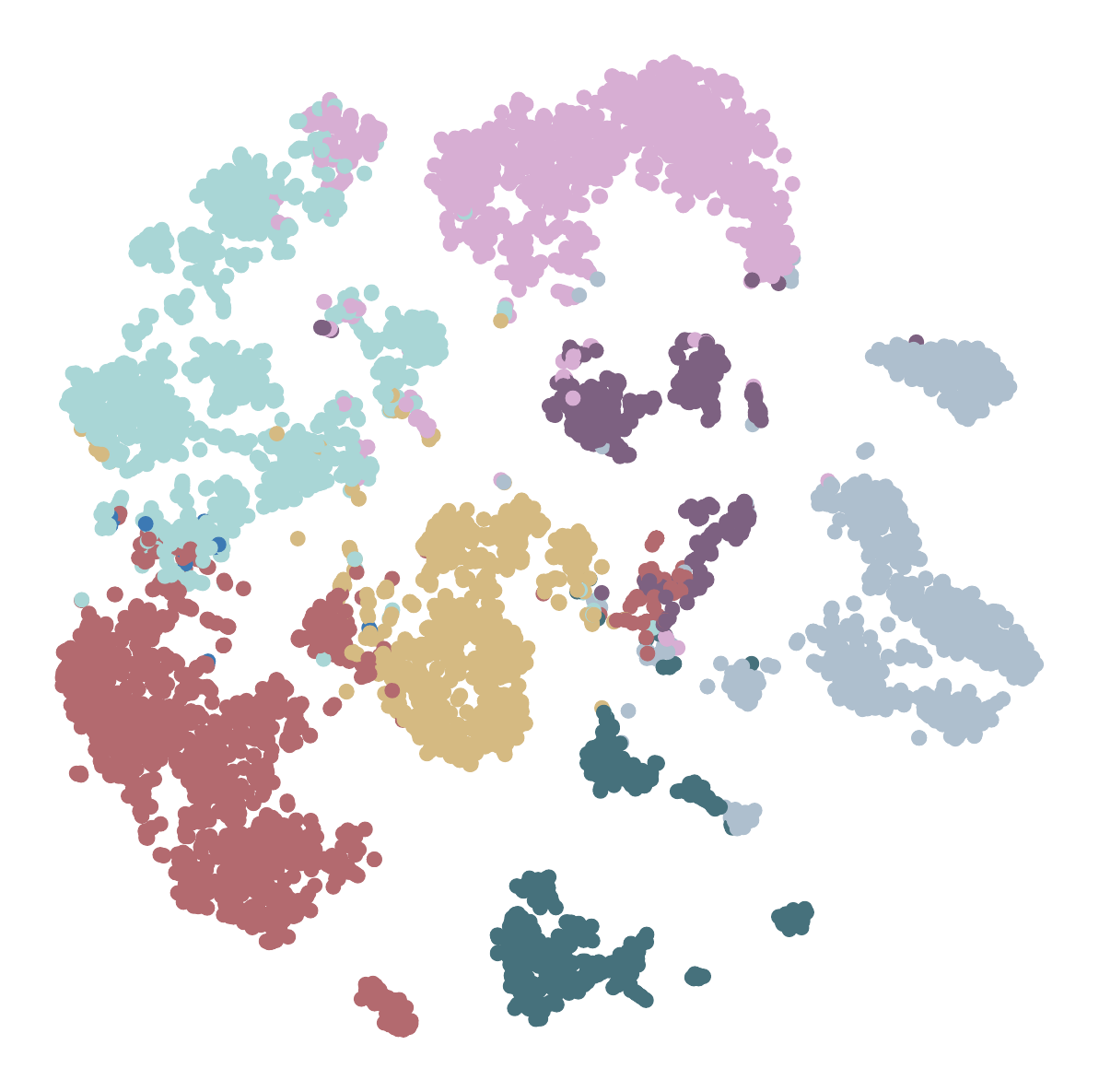}
        \vspace{-8mm}
		\caption*{\footnotesize\mdseries (c) cluster9 (NMI=70.13).}
	\end{minipage}\hfill
	\begin{minipage}[b]{0.3\linewidth}
		\centering
		\includegraphics[width=\linewidth]{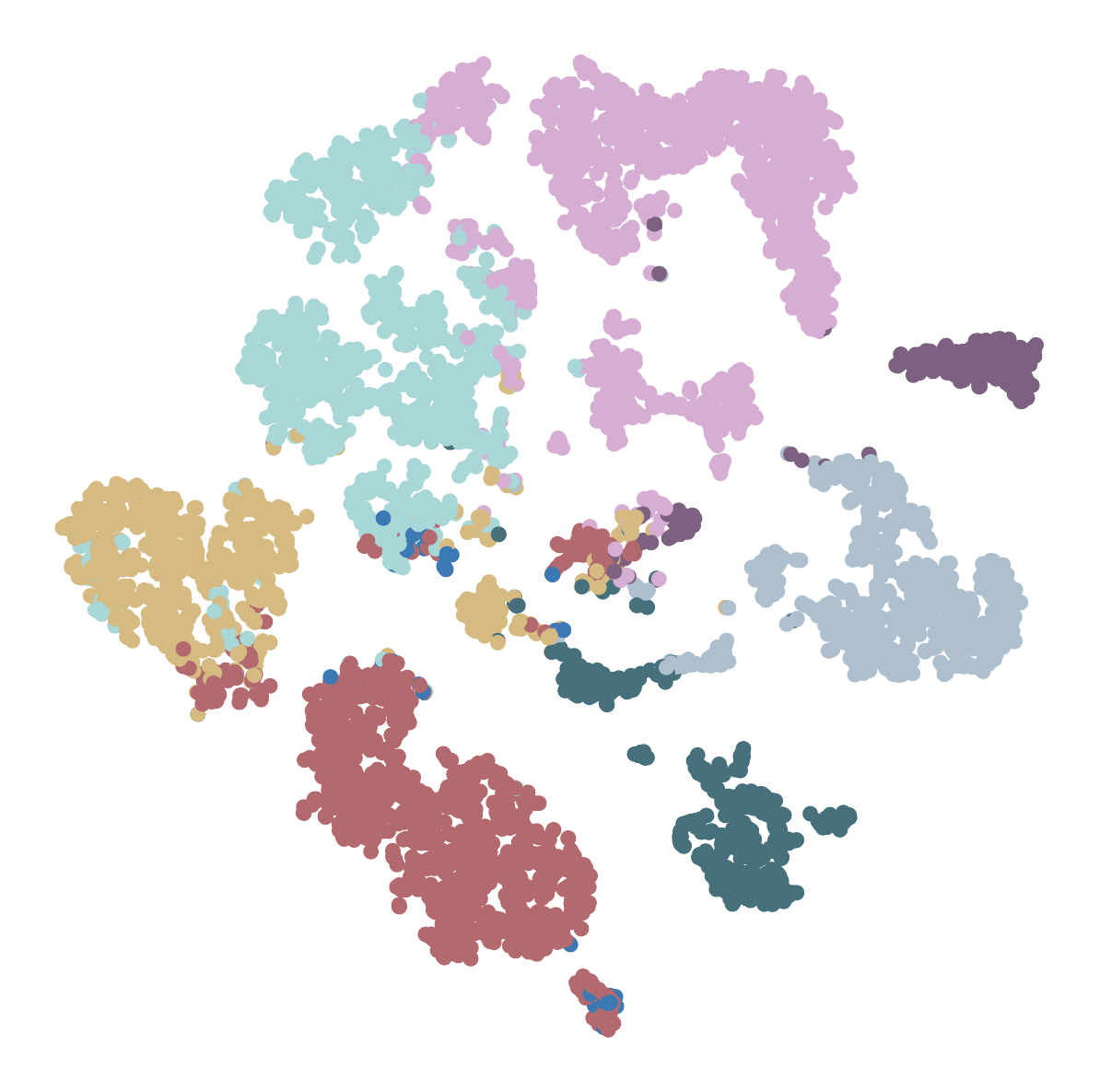}
        \vspace{-8mm}
		\caption*{\footnotesize\mdseries (e) cluster10 (NMI=68.78).}
	\end{minipage}\hfill
	\begin{minipage}[b]{0.3\linewidth}
		\centering
		\includegraphics[width=\linewidth]{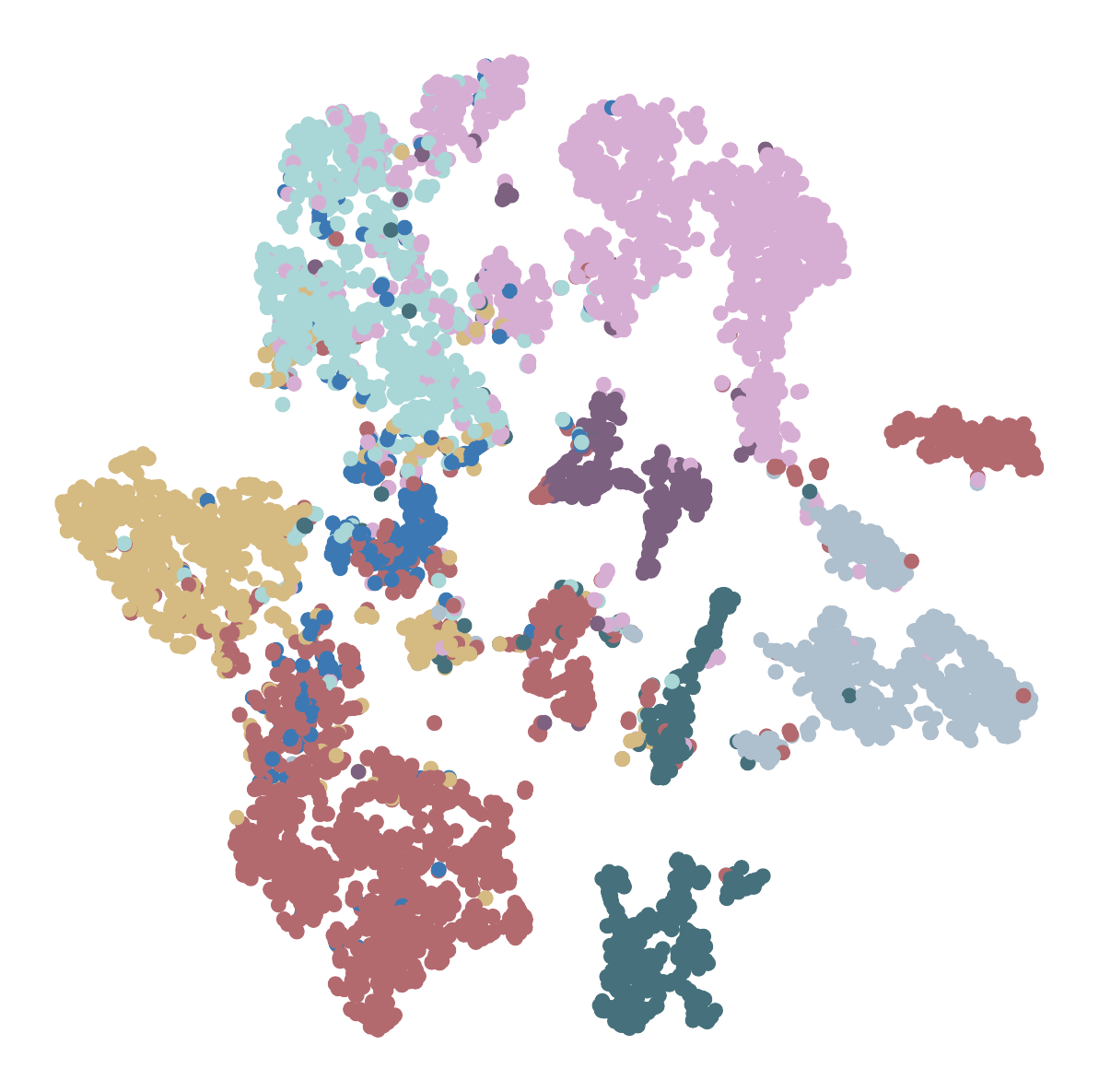}
        \vspace{-8mm}		
        \caption*{\footnotesize\mdseries (b) cluster8 true.}
	\end{minipage}\hfill
	\begin{minipage}[b]{0.3\linewidth}
		\centering
		\includegraphics[width=\linewidth]{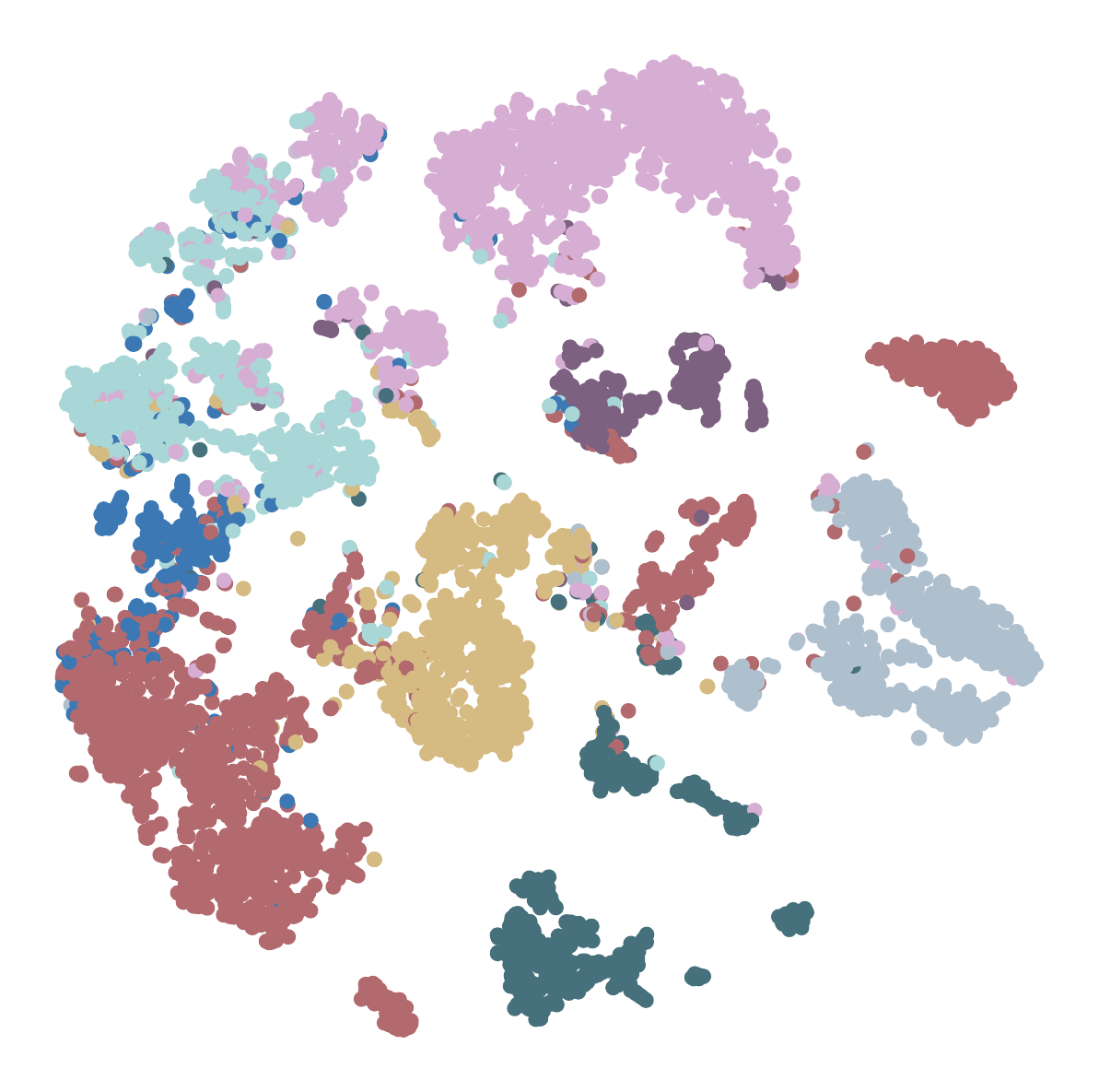}
        \vspace{-8mm}
		\caption*{\footnotesize\mdseries (d) cluster9 true.}
	\end{minipage}\hfill
	\begin{minipage}[b]{0.3\linewidth}
		\centering
		\includegraphics[width=\linewidth]{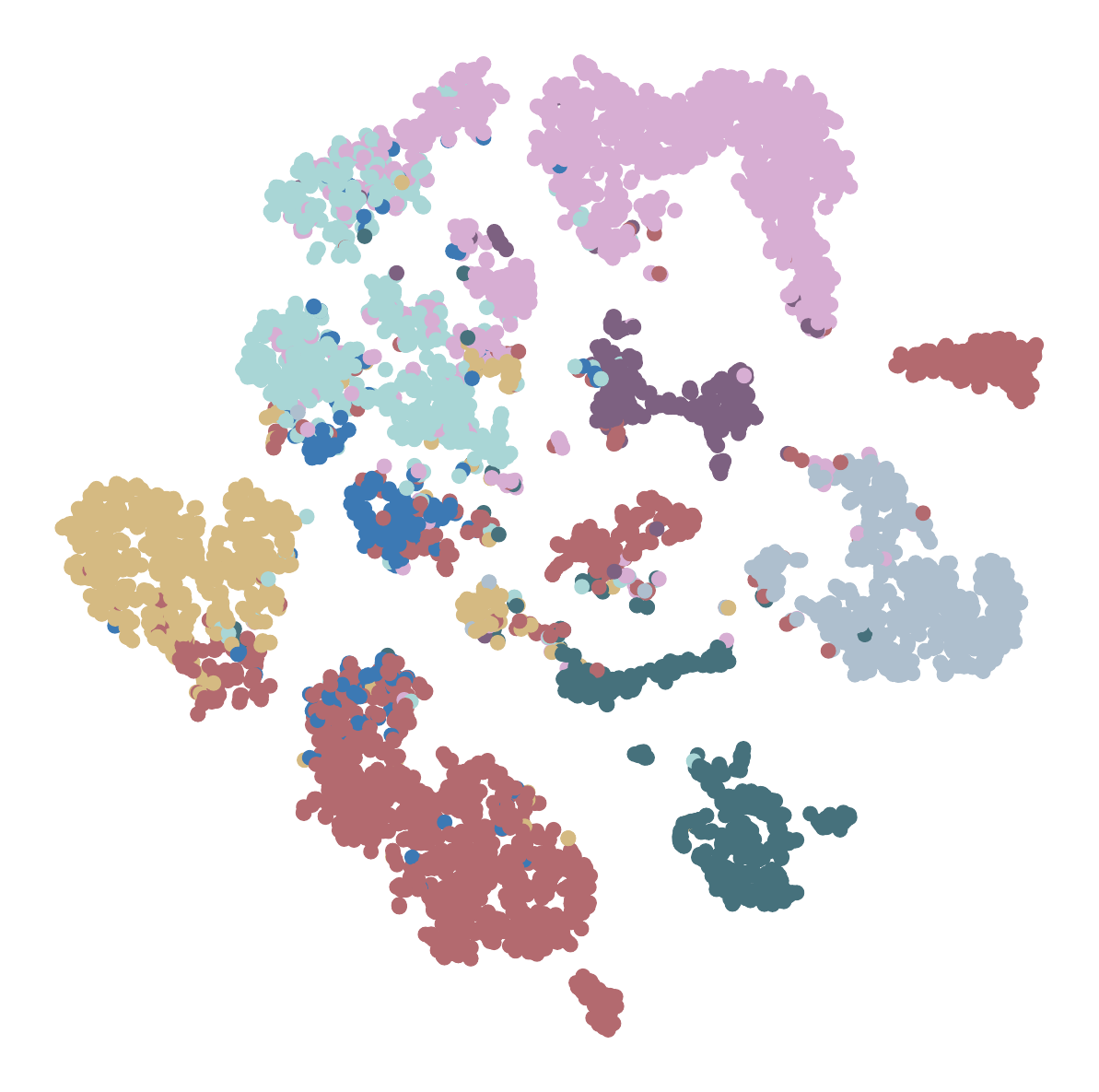}
        \vspace{-8mm}
		\caption*{\footnotesize\mdseries (f) cluster10 true.}
	\end{minipage}\hfill
	\vspace{-4mm}
	\caption{Robustness of cluster numbers On Photo.}
	\label{fig: num_cluster_photo}
        \vspace{-8mm}
\end{figure}
\begin{figure}[h]
	\centering
	\begin{minipage}[b]{0.3\linewidth}
		\centering
		\includegraphics[width=\linewidth]{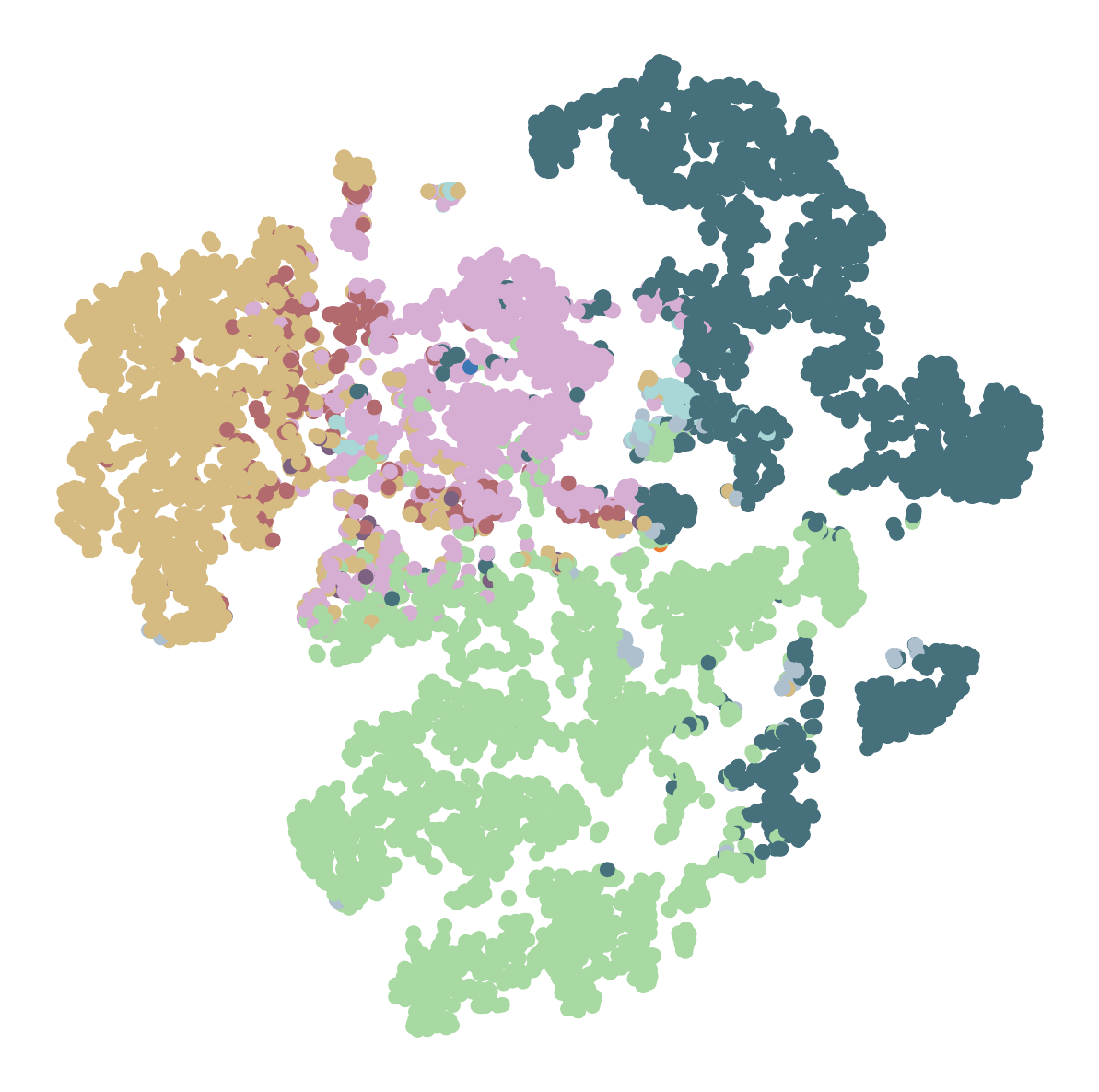}
        \vspace{-8mm}
		\caption*{\footnotesize\mdseries (a) cluster10 (NMI=47.63).}
	\end{minipage}\hfill
	\begin{minipage}[b]{0.3\linewidth}
		\centering
		\includegraphics[width=\linewidth]{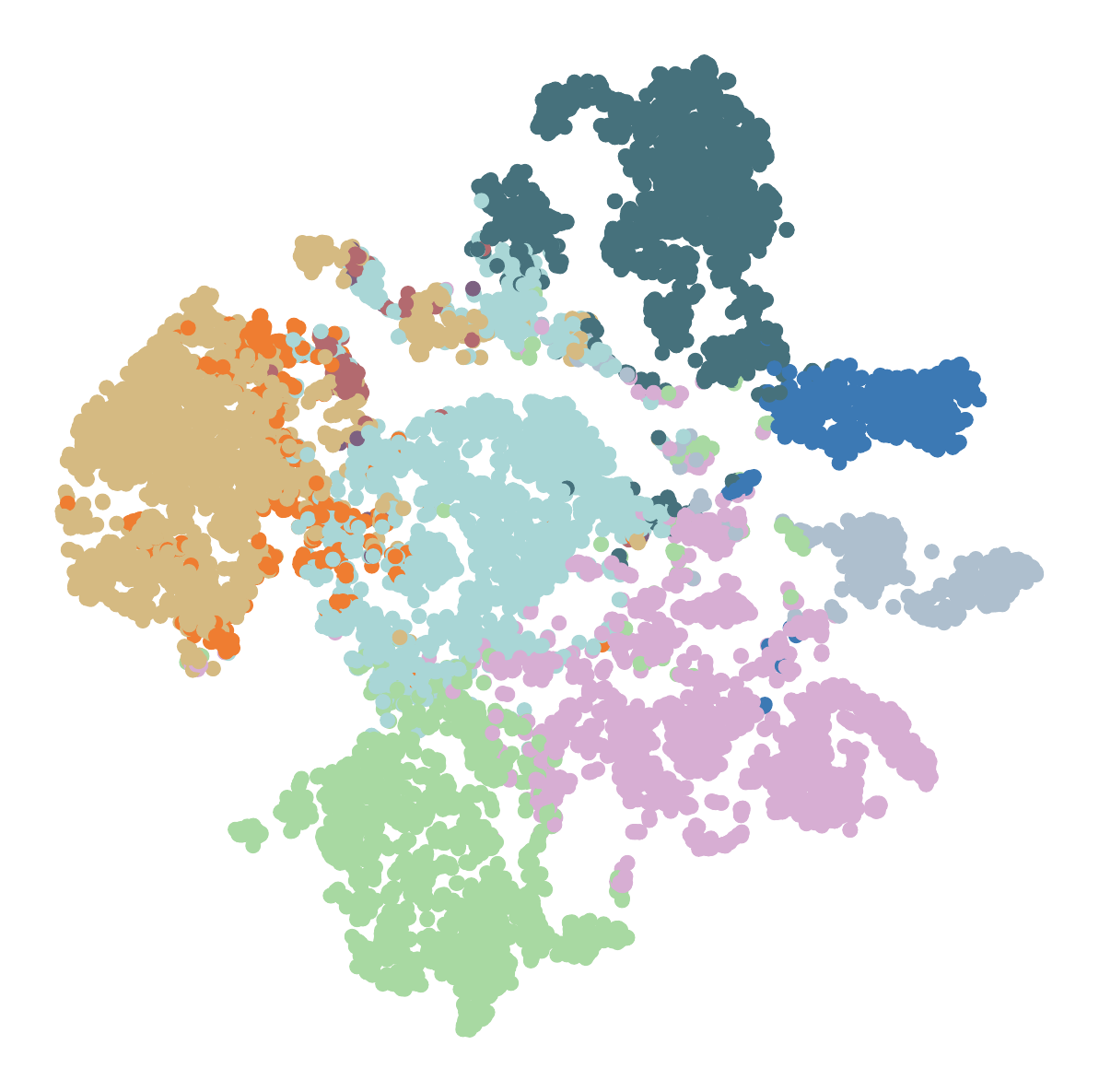}
        \vspace{-8mm}
		\caption*{\footnotesize\mdseries (c) cluster11 (NMI=51.22).}
	\end{minipage}\hfill
	\begin{minipage}[b]{0.3\linewidth}
		\centering
		\includegraphics[width=\linewidth]{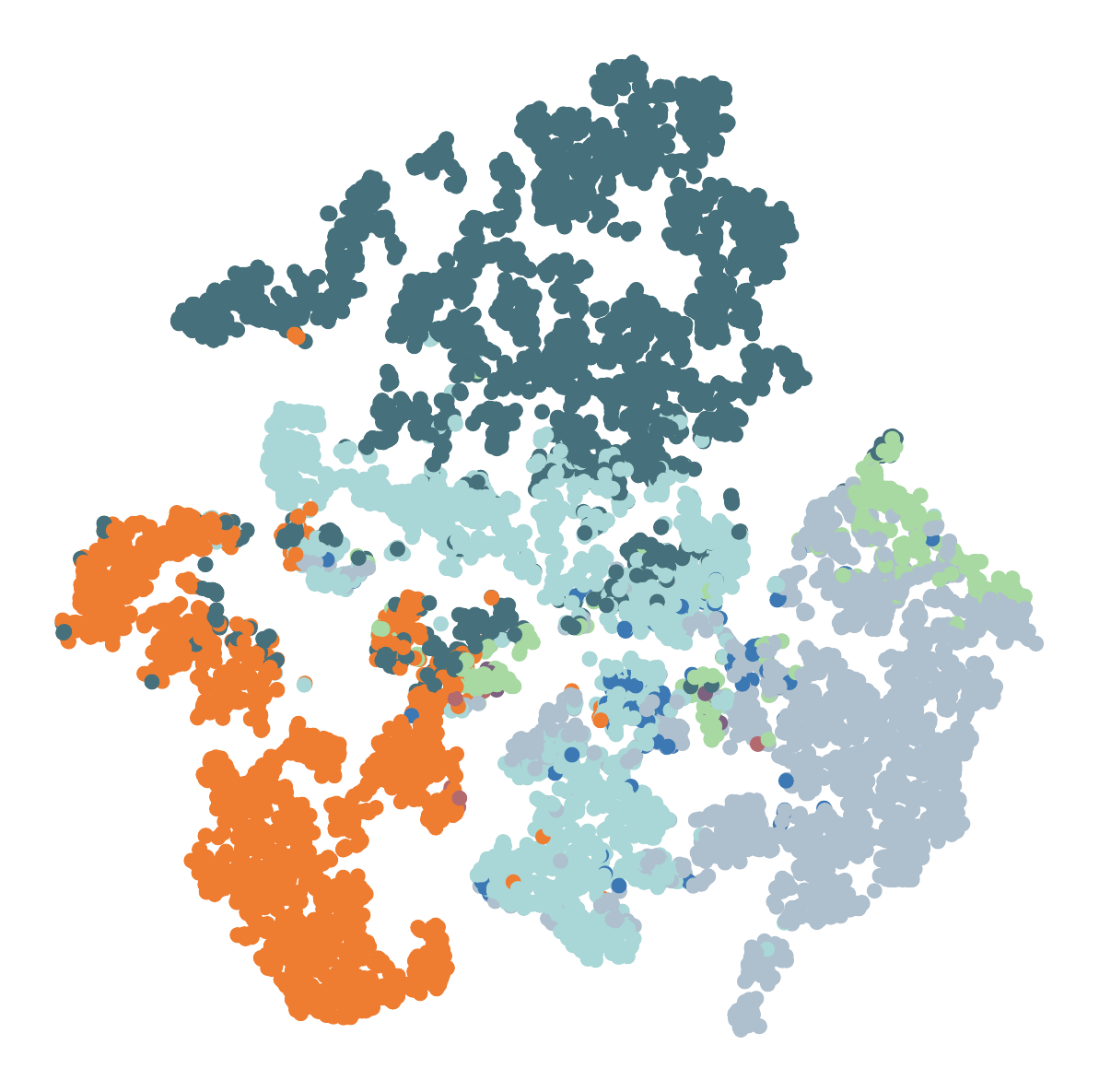}
        \vspace{-8mm}
		\caption*{\footnotesize\mdseries (e) cluster12 (NMI=49.20).}
	\end{minipage}\hfill
	\begin{minipage}[b]{0.3\linewidth}
		\centering
		\includegraphics[width=\linewidth]{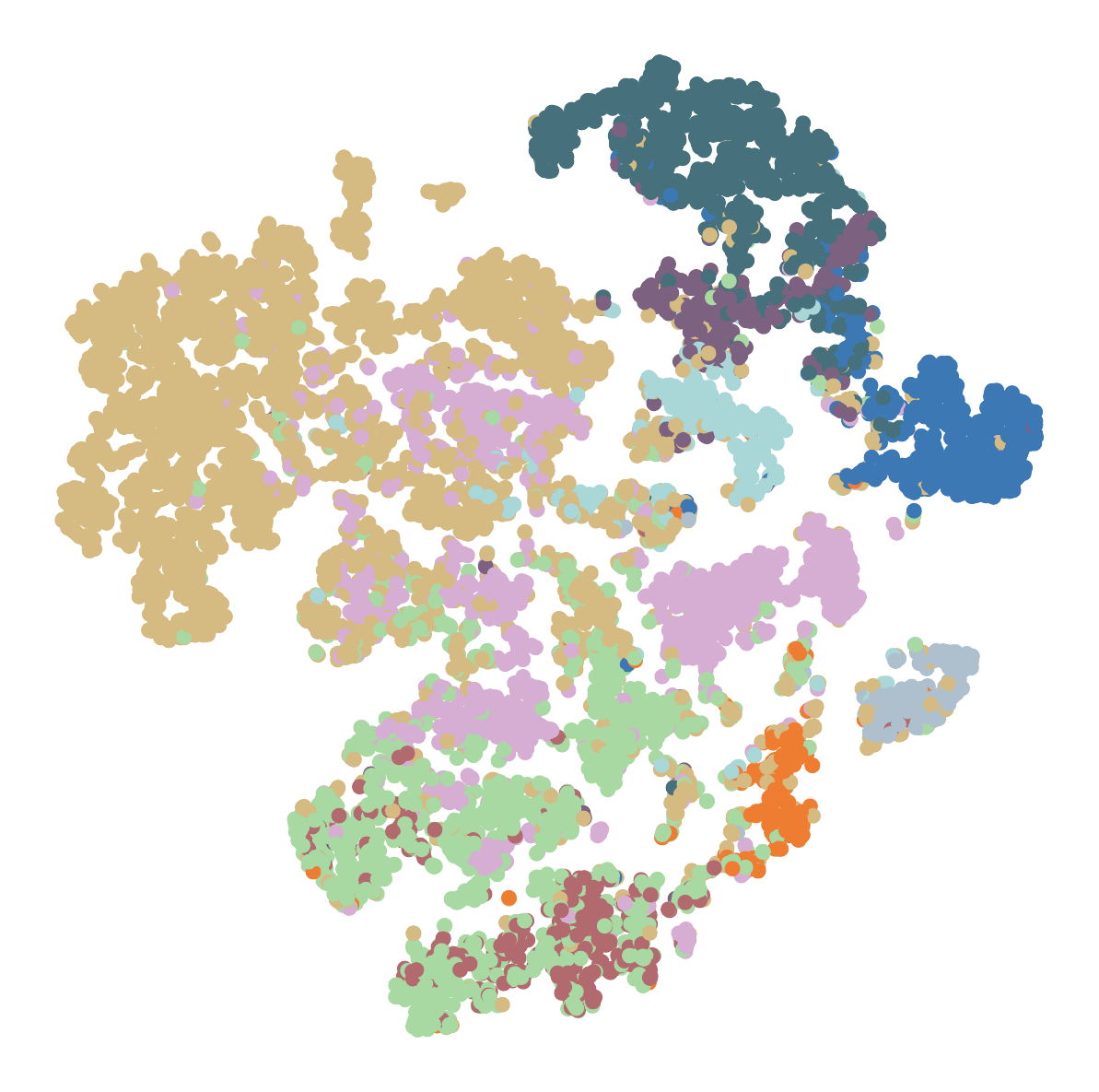}
        \vspace{-8mm}		
        \caption*{\footnotesize\mdseries (b) cluster10 true.}
	\end{minipage}\hfill
	\begin{minipage}[b]{0.3\linewidth}
		\centering
		\includegraphics[width=\linewidth]{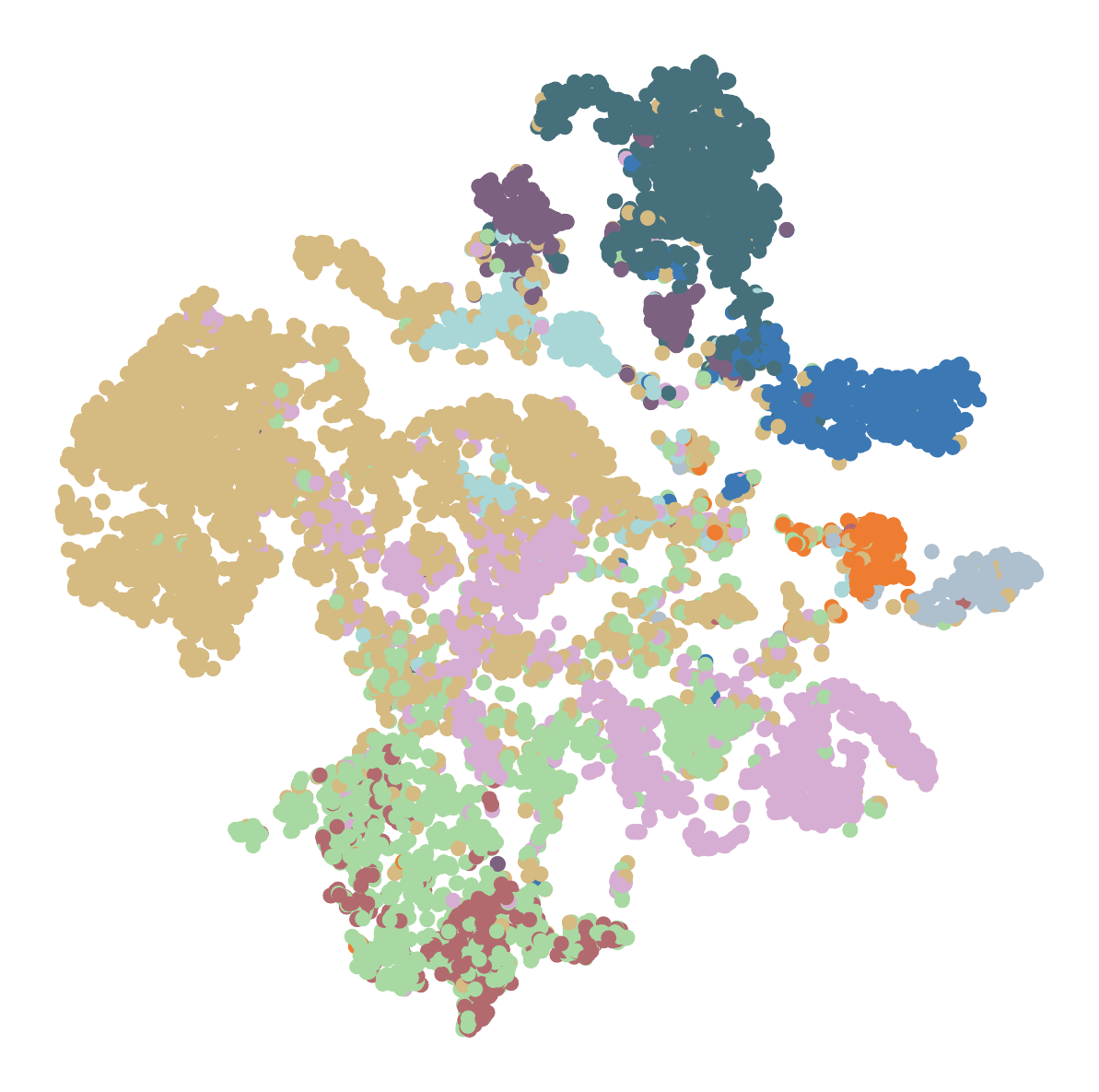}
        \vspace{-8mm}
		\caption*{\footnotesize\mdseries (d) cluster11 true.}
	\end{minipage}\hfill
	\begin{minipage}[b]{0.3\linewidth}
		\centering
		\includegraphics[width=\linewidth]{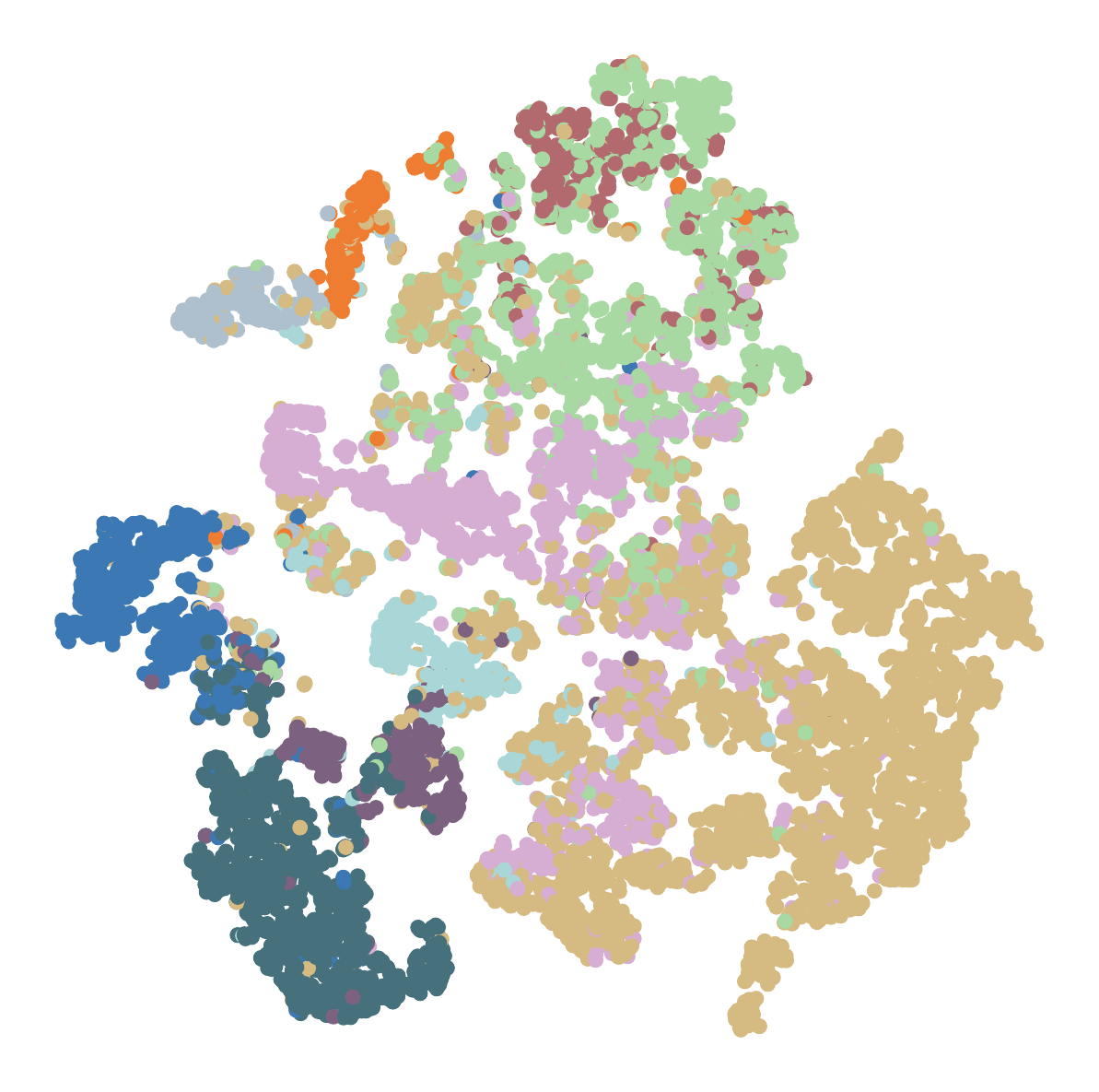}
        \vspace{-8mm}
		\caption*{\footnotesize\mdseries (f) cluster12 true.}
	\end{minipage}\hfill
        \vspace{-4mm}
	\caption{Robustness of cluster numbers On Computer.}
	\label{fig: num_cluster_computer}
        \vspace{-4mm}
\end{figure}

The visual results of our model~\model{} on Citeseer, Photo, and Computer datasets about the robustness of clusters are presented in Figure~\ref{fig: num_cluster_citeseer}, Figure~\ref{fig: num_cluster_photo}, and Figure~\ref{fig: num_cluster_computer}. 
We also observe that, regardless of the number of clusters set (as long as it is greater than or equal to the original number), the model consistently forms the same number of clusters adaptively. 
While there are some differences in NMI, this is due to the lack of fine-tuning of parameters for different cluster numbers.
A key direction for future work is to improve the model to reduce the impact of hyperparameters on clustering accuracy across varying numbers of clusters. \par

\section{Sensitivity Analysis}
\label{ap: sensitivity analysis}
Table~\ref{tab: hyper_loss_acc} presents the ACC results of~\model{} with $\lambda_{se}$ in the range of $\{0.01, 0.05, 0.2, 0.5\}$ and $\lambda_{ce}$ in the range of $\{0.1, 0.5, 1, 5\}$ as a supplement to Section~\ref{sec: hyperparameter} "Coefficients of SE loss and CE loss $\lambda_{se}$ and $\lambda_{ce}$".
It can be observed that, for ACC, the optimal parameter selection follows a similar trend to NMI in Table~\ref{tab: hyper_loss}. 
Specifically, the SE loss tends to favor smaller coefficients, while the CE loss prefers larger coefficients. 
Despite its smaller value, the SE loss plays a significant role in improving clustering accuracy.\par

\section{Time and Memory Analysis} 
\label{ap: time analysis}
We set the epoch to 600 and conduct 10 experiments on ~\model{}. 
Table~\ref{tab: time} records the average runtime across the four datasets.
It can be observed that the ~\model{}'s runtime increases with the number of nodes and edges in the dataset, especially for the Computer dataset, which has higher "Sparsity".
However, overall, the runtime remains within an acceptable range.
Future efficiency improvements may be achievable through enhancements in the selection of K-nearest neighbors in large-scale graphs and the computation of soft-assignment structural entropy.
In addition, Table~\ref{tab: memory} shows that the hyperparameter memory usage of~\model{} is also a major advantage.\par

\begin{table}[h]
    \centering
    \caption{Sensitivity of hyperparameters $\lambda_{se}$ and $\lambda_{ce}$ with ACC.}
    \label{tab: hyper_loss_acc}
    \vspace{-4mm}
    \renewcommand\arraystretch{1.0}
    \setlength{\tabcolsep}{0.8mm}
    \begin{tabular}{c|cccc|cccc}
        \toprule
        \multirow{2}*{Variation} & \multicolumn{4}{c|}{\multirow{1}*{\textbf{SE loss $\lambda_{se}$}}} & \multicolumn{4}{c}{\multirow{1}*{\textbf{CE loss $\lambda_{ce}$}}} \\
        
        \cline{2-9}

        & 0.01 & 0.05 & 0.2 & \multicolumn{1}{c|}{0.5}   
        & 0.1 & 0.5 & 1 & 5 \\

        \hline
        
        Cora 
        & \textbf{75.22} & 74.45 & 72.90 & 72.08
        & 72.30 & 74.59 & 74.63 & \textbf{75.22}\\
        
        Citeseer 
        & \textbf{68.86} & 52.30 & 51.52 & 50.86
        & 67.96 & 68.86 & 68.89 & \textbf{69.04} \\
        
        Computer 
        & 43.20 & - & \textbf{55.87} & 50.49
        & 54.46 & \textbf{55.87} & 44.18 & 43.70 \\

        Photo 
        & \textbf{80.55} & 80.08 & 71.16 & 66.77
        & 66.76 & 71.71 & 72.18 & \textbf{80.55} \\
        
        \bottomrule
    \end{tabular}
    \vspace{-6mm}
\end{table}
\begin{table}[b]
    \centering
    \caption{Average time cost for~\model{} and baselines (Sec).}
    \vspace{-4mm}
    \label{tab: time}
    \renewcommand\arraystretch{1.0}
    \setlength{\tabcolsep}{3mm}
    \begin{tabular}{c|cccc}
        \toprule
        Method & Cora & Citeseer & Computer & Photo\\
        \hline
        \rowcolor{gray!20}
        \model{} & 65.87 & 81.17 & 2037.16 & 684.54\\
        
        \hline
        
        DMoN & 65.07 & 77.80 & 293.47 & 179.90 \\
        MinCut & 69.56 & 79.18 & 243.15 & 146.48 \\
        DGI & 178.24 & 250.88 & 1144.88 & 581.18 \\
        SUBLIME & 104.80 & 137.21 & 477.92 & 215.87 \\
        EGAE & 106.17 & 67.52 &  506.03 & 184.83 \\
        CONVERT & 91.85 & 146.44 & 281.85 & 140.51 \\
        AGC-DRR & 234.44 & 533.16 & 8096.46 & 2885.40 \\
        RDGAE & 56.03 & 107.47 & 1284.25 & 580.86 \\
        \bottomrule
    \end{tabular}
    \vspace{-2mm}
\end{table}
\begin{table}[b]
    \centering
    \caption{Memory usage analysis for~\model{} and baselines (MB).}
    \vspace{-4mm}
    \label{tab: memory}
    \renewcommand\arraystretch{1.0}
    \setlength{\tabcolsep}{3mm}
    \begin{tabular}{c|cccc}
        \toprule
        Method & Cora & Citeseer & Computer & Photo\\
        \hline
        
        \rowcolor{gray!20}
        \model{} & 0.09 & 0.92 & 0.21 & 0.43 \\
        
        \hline
        
        DMoN & 0.96 & 2.07 & 0.63 & 0.62 \\
        MinCut & 0.96 & 2.07 & 0.63 & 0.62 \\
        DGI & 5.81 & 10.24 & 4.51 & 4.47 \\
        SUBLIME & 2.96 & 7.39 & 1.66 & 1.61 \\
        EGAE & 1.52 & 3.74 & 0.87 & 0.85 \\
        CONVERT & 14.69 & 51.58 & 5.75 & 5.75 \\
        AGC-DRR & 6.11 & 6.11 & 6.11 & 6.11 \\
        RDGAE & 0.18 & 0.45 & 0.10 & 0.09 \\
        \bottomrule
    \end{tabular}
\end{table}

\end{document}